\documentclass[runningheads]{llncs}

\usepackage{eccv}

\usepackage{eccvabbrv}

\usepackage{graphicx}
\usepackage{booktabs}
\usepackage{bm}
\usepackage{multirow}
\usepackage[table]{xcolor} 
\usepackage{siunitx} 
\usepackage{algorithm}
\usepackage{algpseudocode}
\usepackage{adjustbox}
\usepackage{float}

\usepackage[accsupp]{axessibility}  % Improves PDF readability for those with disabilities.

\usepackage{hyperref}

\usepackage{orcidlink}

\begin{document}

\title{DP-BOA: Dirichlet-Process Birth-or-Assign for On-the-Fly Category Discovery} 

\titlerunning{DP-BOA}

\author{Peiyan Gu\inst{1}\orcidlink{0009-0002-5084-0350} \and
Zixin Teng\inst{1}\orcidlink{0009-0009-2699-6743} \and
Xuming He\inst{1,2}\thanks{Corresponding author.}\orcidlink{0000-0003-2150-1237}}

\authorrunning{P.~Gu et al.}

\institute{ShanghaiTech University, Shanghai, China \and
Shanghai Engineering Research Center of Intelligent Vision and Imaging 
\email{peiyangu@outlook.com, \{tengzx2025,hexm\}@shanghaitech.edu.cn}}

\maketitle

\begin{abstract}

  % On-the-fly category discovery requires deciding for each incoming test sample whether to assign it to an existing category or spawn a new one. Prevailing radius and hashing heuristics rely on a single global threshold—implicitly assuming spherical, equally scaled categories—while ignoring how decision confidence should adapt as category size grows; they also lack a principled mechanism to expand capacity as novel categories continually emerge. To address these, we propose DP-BOA, a probabilistic posterior-predictive approach based on an online Dirichlet-process Gaussian mixture (DP-GMM) with a Normal–Inverse–Wishart (NIW) prior. During training, we learn a shared prior over category Gaussians from labeled training data and estimate NIW posterior parameters for each known class. At test time, we compute the posterior predictive probability of the sample belonging to each existing category, and compare it against a principled birth probability derived from the DP prior. The sample is assigned or a new category is born based on these probabilities. After this, category statistics are updated online, requiring only constant memory per category. The method supports anisotropic category geometry, evidence-adaptive boundaries, and expandable category capacity. Across OCD benchmarks, DP-BOA consistently surpasses strong baselines and recent state-of-the-art, setting a new SOTA for novel-class discovery while maintaining high known-class accuracy.
  On-the-fly category discovery requires deciding for each incoming test sample whether to assign it to an existing category or spawn a new one. Existing methods typically implement this decision through matching-based heuristics, such as radius- or hash-based rules. While effective in practice, these methods usually treat category birth implicitly as a fallback when no existing category matches confidently, rather than as an explicit alternative supported by its own statistical evidence. To address this, we propose DP-BOA, a posterior-predictive decision framework based on an online Dirichlet-process Gaussian mixture model with a Normal–Inverse–Wishart prior. During training, we use labeled data to calibrate a shared NIW prior over category Gaussians and warm-start the known-category posteriors. At test time, for each incoming sample, DP-BOA compares the posterior predictive evidence for assignment to existing categories against the evidence for spawning a new category induced by the DP prior, and then updates category statistics online after the decision. The method captures anisotropic category geometry and naturally adapts decision confidence as evidence accumulates. Across standard OCD benchmarks, DP-BOA consistently outperforms strong baselines and delivers particularly strong novel-class discovery performance while maintaining competitive known-class accuracy. The project page is available at \href{https://algpy.github.io/DP-BOA-pages/}{\textcolor{NavyBlue}{DP-BOA}}.
  \keywords{On-the-Fly Category Discovery\and Generalized Category Discovery\and Bayesian Nonparametrics}
  
\end{abstract}

\vspace{-2mm}
\section{Introduction}
\label{sec:intro}
\vspace{-1mm}

Deep learning has achieved remarkable progress in visual recognition \cite{krizhevsky2012imagenet,he2016deep,dosovitskiy2020image}, but most models still assume a closed world, i.e., all test categories are seen during training. This assumption rarely holds in dynamic real-world environments. Open-world settings such as Novel Category Discovery (NCD) \cite{han2019learning} and Generalized Category Discovery (GCD) \cite{vaze2022generalized} partially address this issue by using labeled known classes to organize and discover novel ones in an unlabeled set. However, they typically assume offline access to the full unlabeled dataset, which is impractical in many applications (e.g., autonomous driving, robotics, live content moderation) where data arrive continuously and decisions must be made immediately.
To address this gap, On-the-fly Category Discovery (OCD) \cite{du2023fly} considers a strict streaming setting: samples arrive one by one, and the system must instantly decide whether each sample belongs to an existing category or should start a new one. This is particularly challenging because early mistakes can propagate and degrade later decisions.

A key challenge in OCD is the \emph{assign-versus-birth} decision. It involves two coupled questions: how to define ``new'', and how to represent ``old'' under a stream. For the birth, the system needs a reliable criterion for deciding when a sample should start a new category, and labeled known categories are the main available source of prior guidance for this decision. For the assignment, category representations should evolve with the stream: they should be updated with newly assigned samples, remain uncertain when evidence is scarce, and become more confident as support accumulates. These suggest that an effective OCD method should combine an explicit birth criterion with online category representations whose uncertainty adapts to the amount of observed evidence.

Most existing OCD methods make online decisions via distance- or similarity-based matching rules~\cite{du2023fly,zheng2024prototypical,liu2025generate,banerjeelanguage}: a sample is assigned to the nearest prototype if sufficiently close, otherwise it spawns a new one.
While this design is simple and efficient, it treats category birth implicitly as a fallback from failed assignment, rather than as an explicit alternative supported by its own evidence.
It also brings two limitations.
First, such matching rules often induce prototype-centered regions of fixed shape that ignore direction-dependent covariance (e.g., Hamming balls\cite{du2023fly,zheng2024prototypical}, Euclidean balls\cite{liu2025generate}, or angular caps under cosine similarity\cite{banerjeelanguage}), which can be misaligned with heteroscedastic and anisotropic feature distributions (see Appendix~M for empirical geometry statistics). %%%% Such as directly write Appendix~K not good， but isolatedly compile have to do so. \ref{app:appendix_geometry} will show "??"
Second, threshold control in these methods is typically fixed or heuristic and does not explicitly model how predictive uncertainty should contract as category evidence accumulates.
As a result, it becomes harder to distinguish newly formed categories from stable ones, increasing the risk of error propagation after early false births.

To address these limitations, we propose \textbf{DP-BOA}, a probabilistic OCD framework that formulates assign-versus-birth as a Bayesian evidence comparison. For each incoming test sample, DP-BOA compares the prior-weighted predictive evidence for assignment to existing categories against that for spawning a new category, so that birth is no longer treated merely as rejection by existing categories but is explicitly evaluated within the same decision framework. We instantiate this idea with an online Dirichlet Process Gaussian Mixture Model (DP-GMM) \cite{rasmussen2000infinite,wang2011online} and a Normal--Inverse--Wishart (NIW) prior \cite{bishop2006pattern}. Statistics from labeled known classes are used to initialize known-category posteriors and to estimate a shared prior over category Gaussians. Full-covariance category modeling captures anisotropic geometry, posterior-predictive updates allow predictive uncertainty to contract as evidence accumulates, and the DP prior enables open-ended category growth.

We validate these design choices through streaming experiments on standard OCD benchmarks, consistently outperforming baselines. Ablations further show that support-set prior calibration, the size- and concentration-dependent DP category prior, full-covariance modeling, and evidence-adaptive posterior updates each contribute meaningfully to the overall gains.

In summary, our contributions are:
\begin{itemize}
    \item We reformulate OCD as a Bayesian evidence comparison between assigning a sample to an existing category and spawning a new one, making birth an explicit alternative rather than an implicit fallback from failed assignment.
    \item We instantiate this formulation with an online DP-GMM and an NIW prior, yielding an online posterior-predictive algorithm that leverages known-class statistics for prior initialization, captures anisotropic category geometry, and adapts predictive uncertainty as evidence accumulates.
    \item We demonstrate strong performance on standard OCD benchmarks, particularly for novel classes.
\end{itemize}
\vspace{-1.5em}
\section{Related Work}
\label{sec:related_works}
\vspace{-1mm}

\subsection{Generalized / Novel Category Discovery}
Novel Class Discovery (NCD) \cite{han2019learning} aims to cluster novel classes in an unlabeled set by leveraging a labeled known-class set. Early work focused on pairwise similarity prediction for clustering \cite{hsu2018learning,hsu2018multi}, while subsequent methods improved similarity estimation \cite{han2021autonovel,zhao2021novel}, feature learning \cite{zhong2021neighborhood,zhong2021openmix,wang2024self,liu2024novel}, and clustering objectives \cite{fini2021unified,zhang2023novel,xu2024dual}. Generalized Category Discovery (GCD) \cite{vaze2022generalized} extends NCD to unlabeled data containing both known and novel classes. Early GCD methods established strong pool-based baselines with contrastive representation learning and semi-supervised clustering \cite{vaze2022generalized,zhang2023promptcal,wen2023parametric,vaze2023no}. More recent work has explored several distinct directions, including clustering-oriented representation learning via mean-shift updates \cite{choi2024contrastive}, efficient adaptation with spatial prompt tuning \cite{wang2024sptnet}, robustness under domain shifts \cite{wang2024hilo}, debiased parametric learning with distribution guidance \cite{liu2025debgcd}, and hierarchy-aware geometry \cite{liu2025hyperbolic}. 
Beyond centralized visual GCD, recent studies have also extended the problem to decentralized and multimodal settings. Fed-GCD~\cite{pu2024federated} studies GCD under federated learning, where data are distributed across clients with heterogeneous label spaces, while language- or multimodal-assisted methods exploit vision-language models, LLM feedback, or multimodal alignment to improve category discovery~\cite{ouldnoughi2023clip,an2024generalized,su2025multimodal}.
Unlike these offline, pool-based methods, we target a single-pass streaming setting. 
\vspace{-0.5em}
\subsection{On-the-Fly Category Discovery}
On-the-Fly Category Discovery (OCD) studies a single-pass test stream in which each incoming sample must be immediately assigned to an existing category or trigger a new one \cite{du2023fly}. SMILE \cite{du2023fly} introduces the task and uses instance-level hash codes with a Hamming-radius rule for online birth-or-assign decisions. PHE \cite{zheng2024prototypical} improves this line with prototype-guided hash encoding to better preserve category structure in Hamming space. More recent variants incorporate auxiliary priors: DiffGRE \cite{liu2025generate} leverages synthesized novel samples through a generate--refine--encode pipeline, while Sync \cite{banerjeelanguage} augments OCD with language-assisted representations and lightweight active querying. In contrast, we focus on the standard OCD setting without such auxiliary priors.
\vspace{-0.5em}
\subsection{Bayesian Non-parametrics for Mixtures}
Our approach builds on Dirichlet-process (DP) mixture models. Blackwell and MacQueen \cite{blackwell1973ferguson} showed that marginalizing a DP yields the Chinese Restaurant Process, whose concentration parameter governs new-component creation, and Sethuraman’s stick-breaking construction provides a constructive representation \cite{sethuraman1994constructive}. Rasmussen’s Infinite Gaussian Mixture Model \cite{rasmussen2000infinite} popularized DP-GMMs with Normal--Inverse--Wishart (NIW) priors and closed-form Student-$t$ predictive densities under conjugacy. Subsequent work studied more scalable or streaming inference for non-parametric mixtures and clustering \cite{wang2011online,dinari2022sampling,schaeffer2022streaming}, while adjacent non-parametric deep clustering and open-world recognition methods explored related ideas in learned embedding spaces \cite{ronen2022deepdpm,willes2022bayesian}. Unlike these methods, which primarily address unsupervised density modeling or clustering, OCD requires single-pass online decisions with known class support-set supervision.

\vspace{-2mm}
\section{Method}
\label{sec:method}

\vspace{-1mm}
\subsection{Problem Formulation and Preliminaries}
\label{sec:formulation}

We follow the On-the-Fly Category Discovery (OCD) setting \cite{du2023fly}. Let
$\mathcal{D}_S = \{(\mathbf{x}_i, y_i)\}_{i=1}^M \subset \mathcal{X} \times \mathcal{Y}_S$
denote the labeled support set used for training, where $\mathcal{Y}_S$ is the set of known classes. For evaluation, the test stream is
$\mathcal{D}_Q = \{(\mathbf{x}_j^q, y_j^q)\}_{j=1}^N \subset \mathcal{X} \times \mathcal{Y}_Q$,
whose label space satisfies $\mathcal{Y}_Q = \mathcal{Y}_S \cup \mathcal{Y}_N$ with
$\mathcal{Y}_S \cap \mathcal{Y}_N = \varnothing$, where $\mathcal{Y}_N$ denotes the novel classes. Following the standard OCD protocol \cite{du2023fly,zheng2024prototypical}, only $\mathcal{D}_S$ is available during training; at test time, query samples are revealed sequentially, and their ground-truth labels are used only for evaluation. For each arriving sample $\mathbf{x}_t$, the model must decide whether to assign it to one of the existing categories or to declare the birth of a new one.

% \paragraph{Feature Space.}
% Our method operates in a learned feature space. We use a feature encoder $f_\theta$ to map each sample to a $d$-dimensional representation $\mathbf{z}_t = f_\theta(\mathbf{x}_t)$. To keep the feature learning stage simple and standard, the encoder is trained offline on $\mathcal{D}_S$ using only the conventional supervised cross-entropy loss, without additional objective design, and is then frozen during online inference. All subsequent birth-or-assign decisions are made in this feature space. Unlike prior OCD methods, we keep feature learning simple and concentrate our contribution on the online probabilistic decision head. Training details are provided in Sec.\ref{sec:setup}.
\vspace{-0.5em}
\paragraph{Feature Space.}
Our method operates in a feature space learned from the labeled support set.
Specifically, a feature encoder $f_\theta$ maps each sample $\mathbf{x}_t$ to a $d$-dimensional representation $\mathbf{z}_t=f_\theta(\mathbf{x}_t)$.
We train $f_\theta$ offline on the labeled support set $D_S$ using the standard supervised cross-entropy loss, and freeze it during online inference.
We deliberately adopt this simple and generic feature-learning setup to keep the focus of the paper on the birth-or-assign decision mechanism, rather than on task-specific representation objectives.
All probabilistic decisions are then performed in this fixed feature space.
Details are provided in Sec. \ref{sec:setup}.

\begin{figure*}[t]
    \centering
    \includegraphics[width=\linewidth]{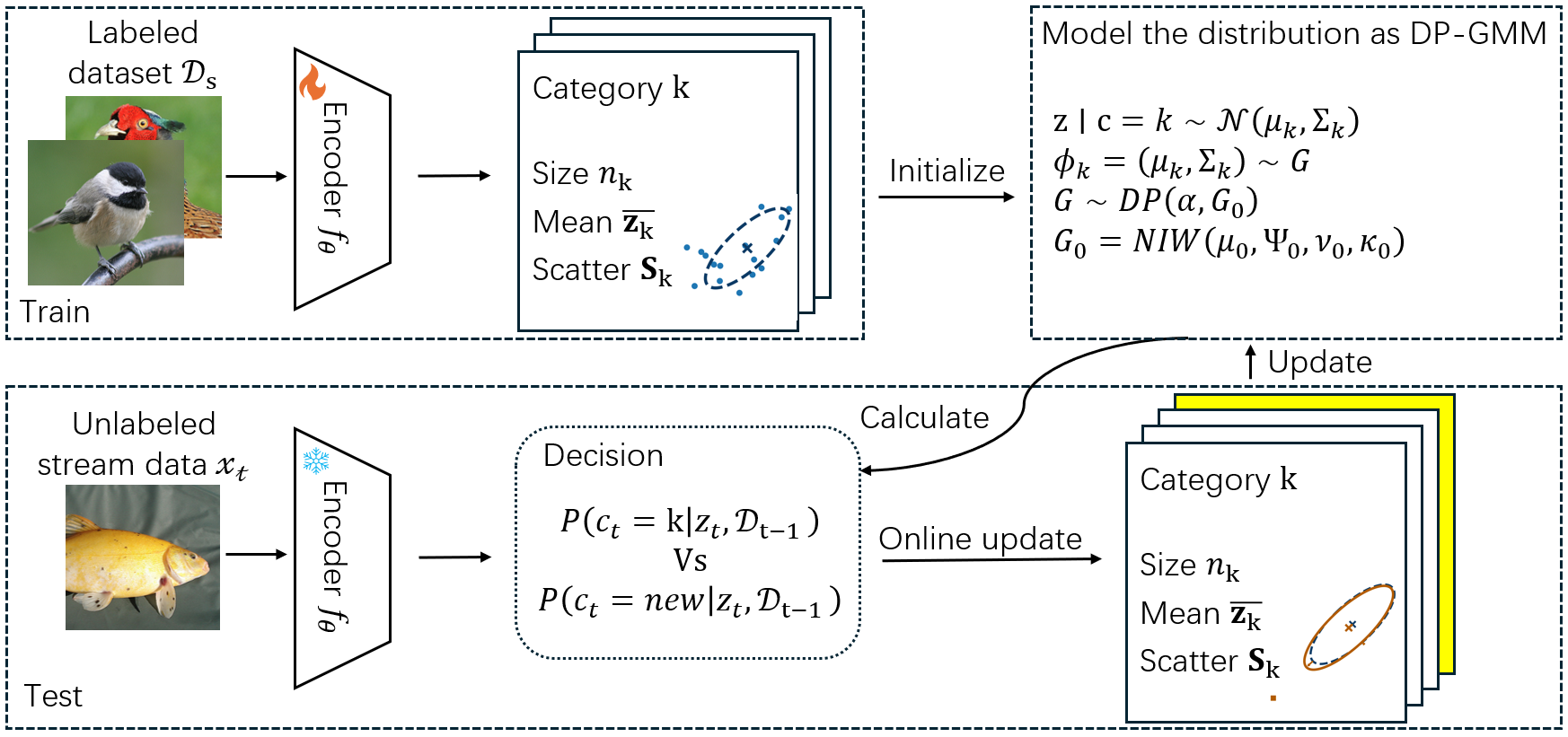}
    \label{fig:short-a}
    \vspace{-1.5em}
    \caption{\textbf{Overview of our proposed DP-BOA framework.}
        Our method consists of two phases.
        \textbf{(Top) Offline Initialization:} We first train a feature encoder $f_\theta$ on the labeled set $\mathcal{D}_S$. We then use the extracted features to compute the sufficient statistics ($n_k, \bar{\mathbf{z}}_k, \mathbf{S}_k$) for all $K_S$ known classes. These statistics are used to initialize our DP-GMM model (i.e., calibrate the global NIW hyperparameters from support-set statistics (\cref{sec:initialization})).
        \textbf{(Bottom) Online Inference:} At test time, a new sample $\mathbf{x}_t$ is passed through the frozen $f_\theta$. Our model performs a posterior-predictive birth-or-assign decision (\cref{eq:final_rule}), comparing the posterior probability of assigning $\mathbf{z}_t$ to an existing category ($P(c_t=k|...)$) versus birthing a new category ($P(c_t=\text{new}|...)$).
        Based on this decision, the statistics of the corresponding category are updated online.
    }
    \label{fig:framework}
    \vspace{-1.5em}
\end{figure*}

\vspace{-1em}
\subsection{A Probabilistic Framework for OCD}
\label{sec:framework}

We cast the core \emph{assign-versus-birth} decision in OCD as a Bayesian evidence comparison. At time $t$, let $K_{t-1}$ be the number of existing categories after processing the first $t-1$ query samples, and let $\mathcal{D}_{t-1}$ denote the current online state. For exposition, we write this state as a history of past observations, although in implementation it is represented only through per-category sufficient statistics. For an arriving feature vector $\mathbf{z}_t$, we choose a decision $c_t \in \{1,\dots,K_{t-1}, \text{new}\}$ by maximizing the posterior:
\begin{equation}
    \label{eq:map_objective}
    \hat{c}_t = \arg\max_{k \in \{1,\dots,K_{t-1}, \text{new}\}} P(c_t=k \mid \mathbf{z}_t, \mathcal{D}_{t-1}) .
\end{equation}

By Bayes' rule,
$
P(c_t=k \mid \mathbf{z}_t, \mathcal{D}_{t-1})
\propto
P(c_t=k \mid \mathcal{D}_{t-1}) \,
p(\mathbf{z}_t \mid c_t=k, \mathcal{D}_{t-1}) .
$
%To obtain a tractable online rule, we make two natural simplifications. 
To obtain a tractable online rule, we make two practical modeling approximations. For an existing category $k$, once $c_t=k$ is assumed, the predictive density depends only on that category's own history, giving $p(\mathbf{z}_t \mid c_t=k, \mathcal{D}_{t-1}) = p(\mathbf{z}_t \mid \mathcal{D}_{t-1}^k)$. For a new category, we use the prior predictive induced by a shared global prior estimated offline, so $p(\mathbf{z}_t \mid c_t=\text{new}, \mathcal{D}_{t-1}) = p(\mathbf{z}_t \mid c_t=\text{new})$. This yields 
%the decision rule
\begin{equation}
    \label{eq:core_formula}
    \begin{split}
        \hat{c}_t
        = & \arg\max \Big\{
        \max_{k \in \{1,\dots,K_{t-1}\}}
        \big[
            P(c_t=k \mid \mathcal{D}_{t-1}) \,
            p(\mathbf{z}_t \mid \mathcal{D}_{t-1}^k)
        \big], \\
        & \hspace{3.3em}
        \big[
            P(c_t=\text{new} \mid \mathcal{D}_{t-1}) \,
            p(\mathbf{z}_t \mid c_t=\text{new})
        \big]
        \Big\} .
    \end{split}
\end{equation}

The remaining task is to specify the two probabilistic components in \cref{eq:core_formula}: the category priors $P(c_t \mid \mathcal{D}_{t-1})$ and the predictive densities $p(\mathbf{z}_t \mid \cdot)$. The following section details our specifications for each.

\vspace{-0.5em}
\subsection{Probabilistic Modeling via Adapted DP-GMM}
\label{sec:model_spec}

As shown in \cref{fig:framework}, we instantiate the framework in \cref{eq:core_formula} with a Dirichlet--Process Gaussian Mixture Model (DP-GMM) \cite{rasmussen2000infinite} and a Normal--Inverse--Wishart (NIW) prior. In contrast to standard DP-GMM usage, OCD provides a labeled support set $\mathcal{D}_S$ and requires single-pass online birth-or-assign decisions without revisiting past samples. We therefore adapt DP-GMM to this setting by using $\mathcal{D}_S$ to estimate a shared prior and initialize known categories. We briefly introduce DP-GMM in our method below.

\vspace{-0.5em}

\paragraph{Predictive Densities (Gaussian + NIW).}
\label{sec:GMMNIW}
To define the continuous density terms in \cref{eq:core_formula} and model anisotropic category geometry, we represent each category $k$ by a full-covariance Gaussian
$\mathcal{N}(\mathbf{z} \mid \bm{\mu}_k, \mathbf{\Sigma}_k)$.
We adopt a Bayesian formulation and place a conjugate Normal--Inverse--Wishart (NIW) prior over the Gaussian parameters $(\bm{\mu}, \mathbf{\Sigma})$, parameterized by global hyperparameters
$\bm{\theta}_0 = (\bm{\mu}_0, \mathbf{\Psi}_0, \nu_0, \kappa_0)$.
Here, $\bm{\mu}_0$ is the prior mean, $\mathbf{\Psi}_0$ is the scale matrix for covariance, $\nu_0$ is the degrees-of-freedom parameter, and $\kappa_0$ controls the strength of the prior on the mean.
A key property of this conjugate prior is that, after observing data, the posterior over $(\bm{\mu}, \mathbf{\Sigma})$ remains NIW.
For an existing category $k$ with data $\mathcal{D}_{t-1}^k$, the posterior NIW hyperparameters are
$\bm{\theta}_k = (\bm{\mu}_k, \mathbf{\Psi}_k, \nu_k, \kappa_k)$.
These can be computed in closed form from the prior $\bm{\theta}_0$ and the sufficient statistics of category $k$: the count $n_k = |\mathcal{D}_{t-1}^k|$, the sample mean $\bar{\mathbf{z}}_k$, and the scatter matrix
$\mathbf{S}_k = \sum_{\mathbf{z} \in \mathcal{D}_{t-1}^k} (\mathbf{z} - \bar{\mathbf{z}}_k)(\mathbf{z} - \bar{\mathbf{z}}_k)^\top$.
The exact NIW update equations are given in Appendix C.
%% app:appendix_niw NIW Update Equations
Integrating out $(\bm{\mu}, \mathbf{\Sigma})$ yields closed-form multivariate Student-$t$ predictive densities $t_d$:
\begin{equation}
    p(\mathbf{z}_t \mid \text{new}) = t_d\!\Bigl(\mathbf{z}_t \,\Bigm|\, \bm{\mu}_0,\,
    \frac{\kappa_0 + 1}{\kappa_0(\nu_0 - d + 1)} \mathbf{\Psi}_0,\,
    \nu_0 - d + 1\Bigr)
    \label{eq:prior_pred}
\end{equation}
\begin{equation}
    p(\mathbf{z}_t \mid \mathcal{D}_{t-1}^k) = t_d\!\Bigl(\mathbf{z}_t \,\Bigm|\, \bm{\mu}_k,\,
    \frac{\kappa_k + 1}{\kappa_k(\nu_k - d + 1)} \mathbf{\Psi}_k,\,
    \nu_k - d + 1\Bigr)
    \label{eq:post_pred}
\end{equation}

Because the posterior hyperparameters are updated with accumulated category evidence, the predictive density for category $k$ becomes progressively sharper as $n_k$ increases. This allows category uncertainty to contract naturally with support, rather than being handled only indirectly through heuristic matching rules.
Appendix D provides the derivation and an illustration of how the predictive density evolves as evidence accumulates.
%% app:appendix_student_t
\vspace{-0.5em}

\paragraph{Category Priors (Dirichlet Process).}
To define the discrete prior probability terms in \cref{eq:core_formula}, we assume a Dirichlet Process (DP) prior \cite{rasmussen2000infinite}, which non-parametrically allows for dynamically expandable category growth.

A standard result from the marginalization of the DP (the Blackwell MacQueen Polya Urn scheme \cite{ferguson1973bayesian,blackwell1973ferguson}) provides the exact prior probabilities for the decision $c_t$ (see Appendix E for details):%% app:appendix_dp_rule
\begin{align}
    P(c_t=k | \mathcal{D}_{t-1}) &= \frac{n_k}{\alpha + N_{t-1}} \label{eq:prior_assign} \\
    P(c_t=\text{new} | \mathcal{D}_{t-1}) &= \frac{\alpha}{\alpha + N_{t-1}} \label{eq:prior_birth}
\end{align}
where $n_k$ is the number of samples in category $k$, $N_{t-1} = \sum_k n_k$ is the total number of samples processed so far, and $\alpha > 0$ is the DP's concentration parameter, acting as a principled ``innovation rate'' for new categories. 

\vspace{-1em}
\subsection{Initialization from Known Categories}
\label{sec:initialization}

%Having specified the predictive likelihoods, we next describe how to initialize the model from the labeled support set $\mathcal{D}_S$.
Unlike standard DP-GMM formulations, which are fully unsupervised, OCD provides labels for a subset of categories through the support set $\mathcal{D}_S$.
We therefore use $\mathcal{D}_S$ to calibrate an EB-style NIW prior and to initialize posterior NIW states for known categories.
This initialization plays two roles:
(1) it uses $\mathcal{D}_S$ to calibrate a robust global prior $\bm{\theta}_0$ (for the ``Birth''), and
(2) it initializes the known categories with their posterior NIW statistics (for the ``Assign'').

\vspace{-1em}
\paragraph{Estimating global priors.}
Our global prior is defined by $\bm{\theta}_0 = (\bm{\mu}_0, \mathbf{\Psi}_0, \nu_0, \kappa_0)$ and the DP concentration $\alpha$.
We use the $K_S=|\mathcal{Y}_S|$ known classes in $\mathcal{D}_S$ to estimate them. For each known category $k$, let $\mathcal{D}_k$ denote the set of support-set features belonging to that category. First, we define the necessary base statistics:
\vspace{-1.5em}
\begin{equation*}
% \resizebox{1.0\linewidth}{!}{
\begin{minipage}{\linewidth}
\begin{align*}
    \quad&\bar{\mathbf{z}}_k = \frac{1}{n_k}\sum_{\mathbf{z} \in \mathcal{D}_k} \mathbf{z}  \quad \text{(category mean)} \\
    \quad&\mathbf{S}_k = \sum_{\mathbf{z} \in \mathcal{D}_k} (\mathbf{z} - \bar{\mathbf{z}}_k)(\mathbf{z} - \bar{\mathbf{z}}_k)^\top  \quad \text{(category scatter)} \\
    \quad&\bar{\mathbf{z}} = \frac{1}{M}\sum_{k=1}^{K_S} n_k \bar{\mathbf{z}}_k  \quad \text{(global mean)} \\
    \quad&\mathbf{\Sigma}_{\text{within}} = \frac{1}{M-K_S}\sum_{k=1}^{K_S} \mathbf{S}_k  \quad \text{(pooled within-category covariance)} \\
    \quad&\mathbf{\Sigma}_{\text{means}} = \frac{1}{K_S-1}\sum_{k=1}^{K_S} (\bar{\mathbf{z}}_k - \bar{\mathbf{z}})(\bar{\mathbf{z}}_k - \bar{\mathbf{z}})^\top  \quad \text{(covariance of category means)} \\
    \quad&\overline{n^{-1}} = \frac{1}{K_S}\sum_{k=1}^{K_S} \frac{1}{n_k} \quad \text{(average inverse category size)}.
\end{align*}
\end{minipage}
% }
\end{equation*}

We use these statistics to estimate the hyperparameters via an empirical-Bayes method \cite{murphy2012machine}(see Appendix F for derivations):%% app:appendix_eb_niw Estimation of NIW Hyperparameters
\begin{itemize}
    \item \textbf{Prior mean ($\bm{\mu}_0$):}
    We set the prior mean to the global mean:
    \begin{equation}
        \label{eq:estimate_nu0}
        \bm{\mu}_0 = \bar{\mathbf{z}}
    \end{equation}
    
    \item \textbf{Covariance scale ($\mathbf{\Psi}_0$):}
    We choose the expected value of the prior covariance $\mathbb{E}[\mathbf{\Sigma}]$ to match the pooled within-category covariance:
    \begin{equation}
        \label{eq:estimate_psi0}
        \mathbb{E}[\mathbf{\Sigma}] = \mathbf{\Sigma}_{\text{within}}
        \quad\Rightarrow\quad
        \mathbf{\Psi}_0 = (\nu_0 - d - 1)\,\mathbf{\Sigma}_{\text{within}}
    \end{equation}
    
    \item \textbf{Mean strength ($\kappa_0$):}
    We estimate $\kappa_0$ by matching the trace of the observed covariance of category means $\mathbf{\Sigma}_{\text{means}}$
    to its theoretical expectation under the NIW prior:
    \begin{equation}
        \label{eq:kappa_trace_match}
        \mathbf{\Sigma}_{\text{means}} \;\approx\;
        \mathbf{\Sigma}_{\text{within}}\left(\overline{n^{-1}} + \frac{1}{\kappa_0}\right)
    \end{equation}
    We use the trace approximation to solve this equation.

    \item \textbf{Tunable priors ($\nu_0, \alpha$):}
    While our EB-style calibration sets most NIW parameters $(\bm{\mu}_{0},\kappa_{0},\mathbf{\Psi}_{0})$ from $\mathcal D_{S}$, two scalars must still be specified: the NIW degrees of freedom $\nu_{0}$ and the DP concentration $\alpha$.
    
    We first focus on $\nu_0$, which we reparameterize as $n_{0}=\nu_{0}-d-1$ for interpretability.
    This $n_0$ primarily controls the prior predictive density $p(\mathbf{z}\mid\text{new})$ for the ``Birth'' hypothesis.
    Given $\mathbf{\Psi}_0=n_0\,\mathbf{\Sigma}_{\text{within}}$, the prior-predictive scale $\mathbf{V}_0$ in \cref{eq:prior_pred} simplifies to:
    % \begin{equation}
    % \label{eq:prior_scale_effect}
    % \resizebox{0.8\linewidth}{!}{$
    % \mathbf{V}_0 \;=\; \frac{\kappa_0+1}{\kappa_0(\nu_0-d+1)}\,\mathbf{\Psi}_0 
    % \;=\; \frac{\kappa_0+1}{\kappa_0}\cdot \frac{n_0}{n_0+2}\,\mathbf{\Sigma}_{\text{within}}$}
    % \end{equation}
    
    \vspace{-0.75em}
    \begin{equation}
    \label{eq:prior_scale_effect}
    \mathbf{V}_0 \;=\; \frac{\kappa_0+1}{\kappa_0(\nu_0-d+1)}\,\mathbf{\Psi}_0 
    \;=\; \frac{\kappa_0+1}{\kappa_0}\cdot \frac{n_0}{n_0+2}\,\mathbf{\Sigma}_{\text{within}}
    \end{equation}
    \vspace{-0.75em}
    
    and the predictive degrees of freedom in \cref{eq:prior_pred} become $\nu_0'=\nu_0 -d + 1 =n_0+2$.
    Thus, $n_0$ jointly tunes the ``peakiness'' (via $\mathbf{V}_0$) and ``tail thickness'' (via $\nu_0'$) of the evidence required to birth a new category.
    Since a single fixed constant (e.g., $n_0=10$) fails to generalize across datasets with different scales (see \cref{tab:sensitivity-n0}), we propose a data-driven heuristic:
    
    \vspace{-0.75em}
	\begin{equation}
	\label{eq:n0_rule}
	n_0 \;=\; \min\!\left(\tfrac{1}{2}\,\bar{n},\; n_{\mathrm{cap}}\right),
	\qquad
	\bar{n}=\frac{1}{K_S}\sum_{k=1}^{K_S} n_k ,
	\end{equation}
    \vspace{-0.75em}
    
    where $n_{\mathrm{cap}}$ is a fixed hyperparameter that limits the maximum prior strength.
    This rule is guided by two principles:
    (1) tying $n_0$ to $\bar{n}$ interprets it as a \emph{pseudo-sample size}, calibrating the prior's strength to the scale of the observed data; and
    (2) the cap at $n_{\mathrm{cap}}$ prevents an over-confident prior on datasets with large $\bar{n}$.
    This is particularly important in OCD, where an overly strong prior derived from $\mathcal{D}_S$ could suppress the discovery of novel categories with different covariance structures.
    Empirically, \cref{eq:n0_rule} yields consistently strong performance and obviates per-dataset tuning.

    For the DP concentration $\alpha$, we find that model performance is highly robust to its value.
    We therefore set a single constant across all datasets; detailed sensitivity analyses of $\nu_0$ and $\alpha$ are provided in \cref{sec:sens}.
\end{itemize}
\vspace{-1.5em}

\paragraph{Initializing known-category posteriors.}
Finally, we initialize the first $K_S = |\mathcal{Y}_S|$ known categories.
Instead of starting them as empty components, we use their sufficient statistics $(n_k, \bar{\mathbf{z}}_k, \mathbf{S}_k)$ computed from $\mathcal{D}_S$ to derive their posterior NIW parameters $(\bm{\mu}_k, \mathbf{\Psi}_k, \nu_k, \kappa_k)$ using the update rules in Appendix C.
%% app:appendix_niw
These $K_S$ initialized categories form the initial set for the online assignment process.

This ``warm start'' ensures not only that the known ``Assign'' categories are modeled with high confidence from the outset, but also that the ``Birth'' hypothesis is governed by a meaningful, data-driven standard (the estimated $\bm{\theta}_0$).

\vspace{-1em}
\subsection{Online Inference and Update}
\label{sec:online_inference}

With the $K_S$ known categories and the global prior initialized, the model begins processing the query stream $\mathcal{D}_Q$. For each arriving sample $\mathbf{z}_t$, which could belong to either an existing category or a novel one, the model must perform the two-stage process: (1) Inference and (2) Update.
\vspace{-1em}

\paragraph{Decision Rule.}
First, the model executes the Bayesian evidence comparison (\cref{eq:core_formula}) to determine if $\mathbf{z}_t$ belongs to an existing category (either one of the $K_S$ known ones or a previously discovered novel one) or represents a new, unseen category. We substitute the components from \cref{sec:model_spec}, yielding the decision $\hat{c}_t$:
\setlength{\abovedisplayskip}{6pt}
\setlength{\belowdisplayskip}{3pt}
% \begin{equation}
% \label{eq:final_rule}
% \resizebox{0.88\linewidth}{!}{$
% \begin{aligned}
% &\arg\max \Big\{ \big[ \alpha \cdot t_d(\mathbf{z}_t | \bm{\mu}_0,
% \frac{\kappa_0+1}{\kappa_0(\nu_0-d+1)}\mathbf{\Psi}_0, \nu_0-d+1) \big],\\
% &\max_{k \in \{1...K_{t-1}\}} n_k t_d(\mathbf{z}_t | \bm{\mu}_k,
% \frac{\kappa_k+1}{\kappa_k(\nu_k-d+1)}\mathbf{\Psi}_k, \nu_k-d+1 )\big]
% \Big\}
% \end{aligned}
% $}
% \end{equation}

\vspace{-0.5em}
\begin{equation}
\label{eq:final_rule}
\begin{aligned}
&\arg\max \Big\{ \big[ \alpha \cdot t_d(\mathbf{z}_t | \bm{\mu}_0,
\frac{\kappa_0+1}{\kappa_0(\nu_0-d+1)}\mathbf{\Psi}_0, \nu_0-d+1) \big],\\
&\max_{k \in \{1...K_{t-1}\}} n_k t_d(\mathbf{z}_t | \bm{\mu}_k,
\frac{\kappa_k+1}{\kappa_k(\nu_k-d+1)}\mathbf{\Psi}_k, \nu_k-d+1 )\big]
\Big\}
\end{aligned}
\end{equation}

\vspace{-1em}

\paragraph{Online Update.}
Second, based on the decision $\hat{c}_t$, the model updates its state online. This update step is critical for the OCD setting, as it allows the model to learn from and refine its understanding of novel categories as they emerge. Due to the conjugacy of the NIW prior, we only need to update the relevant statistics for the chosen category.

\setlength{\abovedisplayskip}{6pt}
\setlength{\belowdisplayskip}{6pt}

\textbf{If Assign ($\hat{c}_t = k$):} We update the sufficient statistics $(n_k, \bar{\mathbf{z}}_k, \mathbf{S}_k)$ of the chosen category $k$ using standard online update rules \cite{welford1962note} (see Appendix G):
%% app:appendix_online_stats
\vspace{-1.5em}
\begin{align}
    n_k^{\text{new}} &= n_k^{\text{old}} + 1 \\
    \bar{\mathbf{z}}_k^{\text{new}} &= \bar{\mathbf{z}}_k^{\text{old}} + \frac{1}{n_k^{\text{new}}}(\mathbf{z}_t - \bar{\mathbf{z}}_k^{\text{old}}) \\
    \mathbf{S}_k^{\text{new}} &= \mathbf{S}_k^{\text{old}} + (\mathbf{z}_t - \bar{\mathbf{z}}_k^{\text{old}})(\mathbf{z}_t - \bar{\mathbf{z}}_k^{\text{new}})^\top
\end{align}

\textbf{If Birth ($\hat{c}_t = \text{new}$):} We increment the category count $K_t = K_{t-1} + 1$, and initialize this new category $k' = K_t$ with its own sufficient statistics:
\begin{equation}
    n_{k'} = 1,\quad  \bar{\mathbf{z}}_{k'} = \mathbf{z}_t,\quad \mathbf{S}_{k'} = \mathbf{0}
\end{equation}

These updated statistics $(n_{\hat{c}_t}, \bar{\mathbf{z}}_{\hat{c}_t}, \mathbf{S}_{\hat{c}_t})$ are then used to compute the category’s posterior parameters $(\bm{\mu}_{\hat{c}_t}, \mathbf{\Psi}_{\hat{c}_t}, \nu_{\hat{c}_t}, \kappa_{\hat{c}_t})$ via the NIW update rules in Appendix C, which in turn define the category’s predictive density $p(\mathbf{z}_{t+1} | \mathcal{D}_{\hat{c}_t})$ in subsequent decisions, thereby fulfilling the ``evidence-adaptive'' requirement. %% app:appendix_niw
\vspace{-0.5em}

\paragraph{Complexity.}
\label{sec:complexity}
% Since DP-BOA replaces lightweight Hamming/radius checks with full-covariance Student-$t$ predictive densities and DP-style updates, we quantify the overhead.
% Let $d$ be the feature dimension and $K$ the number of active categories.
% Because the backbone cost is the same as in prior OCD methods, we focus on the online head.
% For each incoming feature, evaluating full-covariance Student-$t$ predictive densities for all existing categories plus the ``new'' hypothesis costs $O(K d^2)$ time, and updating the sufficient statistics and NIW posterior for the selected category has worst-case cost $O(d^3)$.
% Thus, the per-sample complexity of the head is $O(d^3 + K d^2)$ time with $O(K d^2)$ memory.
% In practice, we cache per-category quantities such as covariance inverses and log-determinants, so only the affected category is recomputed after each update.
% Compared to hashing-based OCD methods~\cite{du2023fly,zheng2024prototypical}, which require only $O(KL)$ operations with $L\!\ll\!d$, DP-BOA explicitly trades extra space and computation for calibrated probabilistic decisions and full-covariance geometry.
% As shown in \cref{sec:runtime}, this yields per-sample latency in the tens of milliseconds on a single GPU, indicating that the method remains practical.
% Appendix I further discusses this trade-off and introduces DP-BOA-L, a lightweight variant designed to reduce the overhead.
Since DP-BOA replaces lightweight Hamming/radius checks with full-covariance Student-$t$ predictive densities and DP-style updates, its online head incurs additional overhead. Let $d$ be the feature dimension and $K$ the number of active categories. Because the backbone cost is the same as in prior OCD methods, we focus on the online head. For each incoming feature, evaluating full-covariance Student-$t$ predictive densities for all existing categories plus the ``new'' costs $O(K d^2)$ time, and updating the sufficient statistics and NIW posterior for the selected category has worst-case cost $O(d^3)$. Thus, DP-BOA has per-sample complexity $O(d^3 + K d^2)$ time with $O(K d^2)$ memory. In practice, we cache covariance inverses and log-determinants, so only the affected category is recomputed after each update. 
To improve scalability, we further consider DP-BOA-L, a lightweight variant that replaces full covariance with a rank-$r$ approximation ($r \ll d$) maintained by Frequent Directions~\cite{liberty2013simple}:
\begin{equation}
\Sigma_k \approx \sigma_k^2 I + U_k \operatorname{diag}(\lambda_k) U_k^\top,\quad U_k\in\mathbb{R}^{d\times r},\ r\ll d .
\end{equation}
Here, the columns of $U_k$ are the $r$ approximate dominant covariance directions, $\lambda_k$ contains the corresponding approximate eigenvalues, and $\sigma_k^2$ is the scalar residual variance used to summarize the covariance outside the low-rank subspace.
With fixed small $r$, its per-sample complexity becomes $O(K d r + d r^2)$ time with $O(K d r)$ memory, reducing the dependence on representation dimensionality from quadratic to linear. 
This preserves the same probabilistic birth-or-assign principle while bringing DP-BOA-L to the same linear order in both representation dimensionality and the number of categories as recent OCD heads~\cite{zheng2024prototypical,liu2025generate,banerjeelanguage}.
DP-BOA-L thus serves as a scalability-oriented approximation to full DP-BOA: it preserves the same online birth-or-assign decision rule, while the low-rank covariance representation substantially reduces latency and memory at the cost of only a small accuracy drop.
Appendix~K provides the implementation and detailed complexity analysis of DP-BOA-L.
%% app:appendix_runtime

%Empirically, DP-BOA-L substantially reduces latency and memory with only a small accuracy drop, showing that the overhead of full-covariance DP-BOA can be effectively alleviated by a simple low-rank approximation.

\section{Experiments}
\label{sec:experiments}

\subsection{Setup}
\label{sec:setup}

\paragraph{Datasets.} Following \cite{du2023fly,zheng2024prototypical}, we conduct experiments on multiple datasets, including two generic datasets—CIFAR100 \cite{krizhevsky2009learning} and ImageNet100 \cite{deng2009imagenet}—as well as eight fine-grained datasets: CUB \cite{wah2011caltech}, Stanford Cars \cite{krause20133d}, Herbarium19 \cite{tan2019herbarium}, Oxford-IIIT Pet~\cite{parkhi2012cats}, and four super-categories from iNaturalist~\cite{van2018inaturalist}, including Fungi, Arachnida, Animalia, and Mollusca. Following OCD \cite{du2023fly}, the categories of each dataset are split into subsets of known and novel classes. Specifically, 50\% of the samples from the known classes are used to form the labeled set $\mathcal{D}_S$ for training, while the remainder forms the query stream $\mathcal{D}_Q$ for on-the-fly testing. The details of the datasets are provided in Appendix A.
\vspace{-0.5em}

\paragraph{Evaluation Protocol.}
We follow the protocol~\cite{vaze2022generalized,zheng2024prototypical} and evaluate using clustering accuracy on unlabeled datasets. 
This metric is calculated by finding an optimal one-to-one mapping $\mathcal{P}$ between predicted cluster indices $\hat{y}_i$ and ground-truth labels $y_i$ with the Hungarian algorithm~\cite{kuhn1955hungarian}:
$
\text{ACC} = \frac{1}{N} \sum_{i=1}^N 1\left\{y_i = \mathcal{P}\left(\hat{y}_i\right)\right\}
$.
We report model performance separately for known, novel, and all classes.
\vspace{-0.5em}

\paragraph{Implementation Details.} We implement our method in PyTorch and run all experiments on NVIDIA TITAN RTX GPUs.
Following OCD \cite{du2023fly,zheng2024prototypical,liu2025generate}, we use a DINO-pretrained ViT-B/16 backbone~\cite{caron2021emerging} and fine-tune only the last transformer block for 100 epochs.
We train with SGD \cite{amari1993backpropagation} (momentum $0.9$) using a cosine learning-rate schedule from $1.0$ down to $10^{-4}$ and a batch size of $256$.
Unlike prior OCD methods~\cite{du2023fly,zheng2024prototypical,liu2025generate,banerjeelanguage}, we do not rely on sophisticated representation learning.
We simply train the encoder with standard supervised cross-entropy on the labeled support set and keep it frozen during online inference, so that any performance gain mainly comes from our probabilistic birth-or-assign head.
After training, we fix the backbone as the encoder $f_\theta$ with feature dimension $d = 768$, and set $\alpha = 10^{-9}$ and $n_{\mathrm{cap}}=50$.
All results are averaged over 3 runs with different seeds.
%In addition, the appendix provides extended analyses of efficiency, robustness, cluster growth dynamics, and hyperparameter sensitivity.

\vspace{-1em}
\subsection{Comparison with the State of the Art}

\label{sec:sota}
\begin{table*}[t]
\caption{Comparison with State-of-the-Art methods on the first five datasets (Part I). We report accuracy for All, Known, and Novel classes. \textbf{Bold} indicates the best performance. Rows highlighted in gray (marked with $^\dagger$) use external knowledge.}
\vspace{-0.5em}
\label{tab:main_results_1}
\centering
\resizebox{1.0\linewidth}{!}{%
\sisetup{table-format=2.1, round-mode=places, round-precision=1} 
\begin{tabular}{l SSS SSS SSS SSS SSS}
\toprule
\multirow{2}{*}{Method} & \multicolumn{3}{c}{Animalia} & \multicolumn{3}{c}{Arachnida} & \multicolumn{3}{c}{CIFAR100} & \multicolumn{3}{c}{CUB} & \multicolumn{3}{c}{Fungi} \\
\cmidrule(lr){2-4} \cmidrule(lr){5-7} \cmidrule(lr){8-10} \cmidrule(lr){11-13} \cmidrule(lr){14-16}
& {All} & {Known} & {Novel} & {All} & {Known} & {Novel} & {All} & {Known} & {Novel} & {All} & {Known} & {Novel} & {All} & {Known} & {Novel} \\
\midrule
SLC\cite{hartigan1975clustering} & 32.4 & 61.9 & 19.3 & 25.4 & 44.6 & 11.4 & 44.4 & 59.0 & 15.1 & 28.6 & 44.0 & 20.9 & 27.7 & 60.0 & 13.4 \\
RankStat\cite{han2021autonovel} & 31.4 & 54.9 & 21.6 & 26.6 & 51.0 & 10.0 & 35.0 & 44.0 & 17.0 & 21.2 & 26.9 & 18.4 & 23.8 & 50.5 & 12.0 \\
WTA\cite{jia2021joint} & 33.4 & 59.8 & 22.4 & 28.1 & 55.5 & 10.9 & 40.8 & 52.9 & 16.7 & 21.9 & 26.9 & 19.4 & 27.5 & 65.5 & 12.0 \\
SMILE\cite{du2023fly} & 35.9 & 49.4 & 30.3 & 29.9 & 57.9 & 12.2 & 51.6 & 61.5 & 31.7 & 32.2 & 50.9 & 22.9 & 29.3 & 64.6 & 13.6 \\
PHE\cite{zheng2024prototypical} & 40.3 & 55.7 & 31.8 & 37.0 & 75.7 & 12.6 & 57.4 & 72.1 & 27.9 & 36.4 & 55.8 & 27.0 & 31.4 & \textbf{67.9} & 15.2 \\
\midrule
\rowcolor{gray!8}
DiffGRE$^\dagger$\cite{liu2025generate} & 43.5 & 63.2 & 35.3 & 47.7 & \textbf{76.6} & 29.4 & {-} & {-} & {-} & 42.5 & 54.4 & 36.5 & {-} & {-} & {-} \\
\rowcolor{gray!8}
Sync$^\dagger$\cite{banerjeelanguage} & {-} & {-} & {-} & {-} & {-} & {-} & 56.1 & 68.4 & 31.5 & 45.3 & 54.3 & 40.9 & {-} & {-} & {-} \\
\midrule
DP-BOA & \textbf{50.7} & \textbf{67.6} & \textbf{43.7} & \textbf{50.6} & 53.4 & \textbf{49.4} & \textbf{60.7} & \textbf{75.0} &\textbf{32.0} & \textbf{53.4} & \textbf{57.2} & \textbf{51.6} & \textbf{51.8} & 65.0 & \textbf{46.0} \\
\bottomrule
\end{tabular}
}
\vspace{-0.5em}
\end{table*}

\begin{table*}[t]
\caption{Comparison with SOTA methods on the remaining five datasets (Part II).}
\vspace{-0.5em}
\label{tab:main_results_2}
\centering
\resizebox{1.0\linewidth}{!}{%
\sisetup{table-format=2.1, round-mode=places, round-precision=1} 
\begin{tabular}{l SSS SSS SSS SSS SSS}
\toprule
\multirow{2}{*}{Method} & \multicolumn{3}{c}{Herbarium19} & \multicolumn{3}{c}{ImageNet100} & \multicolumn{3}{c}{Mollusca} & \multicolumn{3}{c}{OxfordPets} & \multicolumn{3}{c}{StanfordCars} \\
\cmidrule(lr){2-4} \cmidrule(lr){5-7} \cmidrule(lr){8-10} \cmidrule(lr){11-13} \cmidrule(lr){14-16}
& {All} & {Known} & {Novel} & {All} & {Known} & {Novel} & {All} & {Known} & {Novel} & {All} & {Known} & {Novel} & {All} & {Known} & {Novel} \\
\midrule
SLC\cite{hartigan1975clustering} & 14.9 & 27.4 & 8.1 & 32.9 & \textbf{86.5} & 5.2 & 31.1 & 59.8 & 15.0 & 35.5 & 41.3 & 33.1 & 14.0 & 23.0 & 9.7 \\
RankStat\cite{han2021autonovel} & 13.8 & 20.6 & 10.2 & 31.1 & 73.3 & 9.8 & 29.3 & 55.2 & 15.5 & 33.2 & 42.3 & 28.4 & 14.8 & 19.9 & 12.3 \\
WTA\cite{jia2021joint} & 14.6 & 21.2 & 11.1 & 30.8 & 72.9 & 19.4 & 30.3 & 55.4 & 17.0 & 35.2 & 46.3 & 29.3 & 17.1 & 24.4 & 13.6 \\
SMILE\cite{du2023fly} & 22.9 & 39.3 & 14.1 & 33.8 & 74.2 & 13.4 & 33.3 & 44.5 & 27.2 & 41.2 & 42.1 & 40.7 & 26.2 & 46.6 & 16.2 \\
PHE\cite{zheng2024prototypical} & 22.6 & 40.5 & 12.9 & 34.0 & 80.2 & 10.9 & 39.9 & \textbf{65.0} & 26.5 & 48.3 & 53.8 & 45.4 & 31.3 & \textbf{61.9} & 16.8 \\
\midrule
\rowcolor{gray!8}
DiffGRE$^\dagger$\cite{liu2025generate} & {-} & {-} & {-} & {-} & {-} & {-} & 42.6 & 62.0 & 32.3 & 49.6 & 50.1 & 49.3 & 27.7 & 48.1 & 17.8 \\
\rowcolor{gray!8}
Sync$^\dagger$\cite{banerjeelanguage} & {-} & {-} & {-} & \textbf{44.0} & 86.2 & \textbf{22.8} & {-} & {-} & {-} & \textbf{61.6} & \textbf{69.5} & \textbf{57.5} & 24.6 & 34.8 & 19.5 \\
\midrule
DP-BOA & \textbf{30.2} & \textbf{46.4} & \textbf{21.5} & 33.8 & 75.8 & 12.7 & \textbf{48.9} & 53.9 & \textbf{46.3} & 59.0 & 63.6 & 56.6 & \textbf{32.4} & 58.8 & \textbf{19.7} \\
\bottomrule
\end{tabular}
}
\vspace{-1.5em}
\end{table*}

\paragraph{Compared Methods.}
In \cref{tab:main_results_1,tab:main_results_2}, we compare DP-BOA against representative methods on ten benchmarks: three from adjacent settings that are standard OCD baselines (SLC~\cite{hartigan1975clustering}, RankStat~\cite{han2021autonovel}, WTA~\cite{jia2021joint}) and four OCD-specific methods (SMILE~\cite{du2023fly}, PHE~\cite{zheng2024prototypical}, DiffGRE$^\dagger$~\cite{liu2025generate}, Sync$^\dagger$~\cite{banerjeelanguage}). Methods marked with $^\dagger$ use \emph{external knowledge} (e.g., diffusion models or CLIP+text), making the comparison conservative for DP-BOA. For DiffGRE, we report its strongest variant —PHE + DiffGRE with online clustering inference—which achieves the highest reported average. ``--'' denotes metrics not reported in the original works.
% and unavailable for re-evaluation due to missing code.

\vspace{-0.5em}
\paragraph{Results.}
Without using any external knowledge, DP-BOA attains the best \emph{All} score on \textbf{8/10} datasets. It is particularly strong on novel classes: on CUB, DP-BOA reaches \textbf{51.6\%} Novel accuracy, a \textbf{+10.7} point gain over the next-best method (Sync), and on Mollusca it achieves \textbf{46.3\%} vs \textbf{32.3\%} for DiffGRE (\textbf{+14.0}). This indicates that our probabilistic framework is highly effective for the core ``on-the-fly discovery'' task. 
Sync obtains higher \emph{All} accuracy on ImageNet100 and OxfordPets, but relies on CLIP and text. Among methods that do not use external knowledge, DP-BOA (33.8\%) is competitive with PHE (34.0\%) on ImageNet100 and clearly outperforms PHE on OxfordPets (59.0\% vs 48.3\%). Overall, DP-BOA sets a new state-of-the-art among OCD methods that operate purely on the visual stream, offering stronger and more balanced novel-category discovery without external knowledge bases. 
% Further investigations into scalability (including large-K extrapolation and a low-rank variant), as well as robustness under multi-domain distribution shifts, are detailed in the appendix.

\vspace{-1em}
\subsection{Ablation and Analysis}

\begin{table}[t]
\caption{\textbf{Ablation on probabilistic design choices in DP-BOA.}
Each row disables one component while keeping others fixed:
\emph{w/o $\mu_0$}: zero prior mean (no data–driven centering);
\emph{w/o $\Psi_0$}: identity prior covariance (no scale matching);
\emph{w/o $\kappa_0$}: set $\kappa_0{=}1$ (no EB estimation of $\kappa_0$);
\emph{w/o DP prior}: remove the DP category prior terms and score hypotheses using predictive densities only;
\emph{w/o adapt}: freeze NIW posteriors at test time (no evidence-adaptive updating);
\emph{Spherical}: use an isotropic covariance instead of a full matrix.
Best in \textbf{bold}.
}
\vspace{-0.5em}
\label{tab:ablation_bayes}
\centering
\begingroup
\footnotesize
\setlength{\tabcolsep}{4.0pt}
\resizebox{0.9\linewidth}{!}{%
\begin{tabular}{lccccccccc}
\toprule
\multirow{2}{*}{Method} &
\multicolumn{3}{c}{Animalia} &
\multicolumn{3}{c}{CUB} &
\multicolumn{3}{c}{OxfordPets} \\
\cmidrule(lr){2-4} \cmidrule(lr){5-7} \cmidrule(lr){8-10}
& All & Known & Novel & All & Known & Novel & All & Known & Novel \\
\midrule
w/o $\mu_0$            & 50.1 & 65.4 & \textbf{43.7} & 52.9 & 51.1 & \textbf{53.8} & 58.1 & 63.0 & 55.5 \\
w/o $\Psi_0$           & 47.9 & 70.6 & 38.4 & 40.8 & 39.8 & 41.3 & 56.0 & 52.0 & \textbf{58.2} \\
w/o $\kappa_0$         & 45.2 & 59.1 & 39.4 & 52.2 & 51.7 & 52.4 & 57.5 & 60.1 & 56.2 \\
w/o DP prior           & 48.0 & \textbf{71.3} & 38.4 & 52.3 & \textbf{57.3} & 49.8 & 57.1 & 62.7 & 54.2 \\
w/o adapt              & 40.2 & 63.7 & 30.5 & 35.9 & 53.6 & 27.0 & 54.7 & 55.8 & 54.2 \\
Spherical              & 41.7 & 61.7 & 33.5 & 40.4 & 49.0 & 36.1 & 47.1 & 45.5 & 47.9 \\
\midrule
\textbf{DP-BOA} & \textbf{50.7} & 67.6 & \textbf{43.7} &
\textbf{53.4} & 57.2 & 51.6 &
\textbf{59.0} & \textbf{63.6} & 56.6 \\
\bottomrule
\end{tabular}%
}
\endgroup
\vspace{-1.5em}
\end{table}

\paragraph{Ablation on probabilistic design choices.}
We ablate our probabilistic design by disabling one component at a time while keeping all others fixed (\cref{tab:ablation_bayes}).

\textbf{Prior initialization.}
%The first three rows remove individual empirical-Bayes (EB) components of the birth prior.
The first three rows remove individual support-set calibration components of the birth prior.
Overall, the full calibrated initialization (DP-BOA) attains the best \emph{All} accuracy on all three datasets.
Removing the data-driven prior mean $\mu_0$ has only a mild effect, but discarding the learned covariance scale $\Psi_0$ or mean strength $\kappa_0$ is clearly harmful.
In particular, \emph{w/o $\Psi_0$} produces the largest drops (e.g., CUB \emph{All} 53.4→40.8), and \emph{w/o $\kappa_0$} consistently reduces \emph{All} by 5.5/1.2/1.5 points, indicating that EB scale matching and mean shrinkage are key to a well-calibrated prior.

\textbf{Dirichlet–process prior.}
Removing the DP prior in \cref{eq:prior_birth,eq:prior_assign} (\emph{w/o DP prior}) and scoring categories purely by predictive density—treating all existing clusters as equally likely a priori—degrades performance on all datasets.
Novel accuracy drops from 43.7 to 38.4 on Animalia and from 56.6 to 54.2 on OxfordPets, suggesting that the DP size- and concentration-dependent terms help stabilize the birth rate and avoid over-fragmenting categories.

\textbf{Evidence-adaptive updates.}
Freezing NIW posteriors at test time (\emph{w/o adapt}) is particularly damaging for novel-class discovery.
Novel accuracy drops from 43.7→30.5 on Animalia and 51.6→27.0 on CUB, while Known accuracy remains relatively high, showing that continuously updating cluster posteriors with incoming evidence is crucial.

\textbf{Covariance geometry.}
Finally, we compare our full-covariance model with the \emph{Spherical} variant that enforces isotropy for both prior and posteriors.
Concretely, we (i) match the prior scale to the full model by setting
$\Psi_0^{\text{sph}} = (\nu_0 - d - 1)\,\bar\sigma^2 \mathbf I$, where
$\bar\sigma^2 = \tfrac{1}{d}\mathrm{tr}(\Sigma_{\text{within}})$,
and (ii) after each NIW update, project the posterior to spherical by
$\Psi_k^{\text{sph}} \leftarrow \big(\tfrac{1}{d}\mathrm{tr}(\Psi_k)\big)\mathbf I$.
Even under this calibrated construction, Spherical lags far behind Full:
\emph{All} drops by 9.0/13.0/11.9 points and \emph{Novel} by 10.2/15.5/8.7 on Animalia/CUB/OxfordPets.
This highlights that modeling anisotropic category shapes is essential for accurate posterior predictives and reliable birth-or-assign decisions.

% \vspace{-0.5em}
% \paragraph{Sensitivity to $n_0$.}
% \label{sec:sens}
% The reparameterization $n_0=\nu_0-d-1$ controls the tail heaviness of the Student-$t$ prior predictive: \emph{smaller} $n_0$ yields heavier tails (favoring \emph{birth}), whereas \emph{larger} $n_0$ tightens the predictive (favoring \emph{assign}).
% As shown in \cref{tab:sensitivity-n0}, our simple rule $n_0=\min(\bar n/2,\,n_{\mathrm{cap}})$, denoted as ``$\bar n/2^{*}$'', is consistently near-optimal—\underline{second-best} and sometimes best (e.g., \textit{Known} on OxfordPets)—while staying within $0.7$ (Animalia All), $2.0$ (CUB All), and $1.8$ (OxfordPets All) points of the peak.
% This suggests that a data-scaled $n_0$ provides a robust birth/assign calibration.
% Although more sophisticated dataset- or stream-adaptive schedules for $n_0$ may exist, we adopt this rule for its simplicity, stability, and strong second-best performance.

\vspace{-0.5em}
\paragraph{Sensitivity to $n_0$ and $\alpha$.}
\label{sec:sens}
The reparameterization $n_0=\nu_0-d-1$ controls the tail heaviness of the Student-$t$ prior predictive, while the DP concentration $\alpha$ sets the prior odds of birthing a new category.
Very small $n_0$ makes the birth predictive diffuse, while large $n_0$ concentrates it around the global prior; similarly, too small $\alpha$ suppresses novel-class creation, whereas too large $\alpha$ can over-fragment the stream.
In \cref{tab:sensitivity-n0}, the data-scaled rule $n_0=\min(\bar n/2,\,n_{\mathrm{cap}})$ remains close to the best \textit{All} accuracy, within $0.7$, $2.0$, and $1.8$ points on Animalia, CUB, and OxfordPets.
\Cref{tab:sensitivity-alpha} shows similar stability for $\alpha$: the default $\log\alpha=-9$ is within $0.5$, $0.1$, and $2.4$ \textit{All} points of the best value on the same datasets.
Although more sophisticated dataset- or stream-adaptive schedules may exist, we adopt these rules for their simplicity, stability, and near-optimal performance.

% \begin{table}[t]
% \centering
% \caption{\textbf{Sensitivity to $n_0$}.
% \textbf{Bold} = best per column, \underline{underlined} = second-best.
% Row $\bar n/2^{*}$ is our default rule $n_0=\min(\bar n/2,\,n_{\mathrm{cap}})$.}
% \label{tab:sensitivity-n0}
% \setlength{\tabcolsep}{4pt}
% \resizebox{0.9\linewidth}{!}{
% \begin{tabular}{lccccccccc}
% \toprule
% \multirow{2}{*}{$n_0$} &
% \multicolumn{3}{c}{Animalia ($\bar n/2{=}19$)} &
% \multicolumn{3}{c}{CUB ($\bar n/2{=}7$)} &
% \multicolumn{3}{c}{OxfordPets ($\bar n/2{=}24$)} \\
% \cmidrule(lr){2-4}\cmidrule(lr){5-7}\cmidrule(lr){8-10}
% & All & Known & Novel & All & Known & Novel & All & Known & Novel \\
% \midrule
% 5   & 40.7 & 60.7 & 32.5 & \textbf{55.4} & \textbf{60.0} & \textbf{53.2} & 58.5 & 62.0 & 56.6 \\
% 10  & 46.5 & 63.3 & 39.5 & 50.7 & 52.3 & 49.9 & 57.8 & 55.6 & 58.9 \\
% 15  & 48.0 & \textbf{69.7} & 39.0 & 47.6 & 47.4 & 46.1 & 58.4 & 55.4 & \underline{60.0} \\
% 20  & 49.9 & 66.6 & 42.9 & 46.2 & 43.0 & 48.8 & 58.7 & 61.2 & 57.5 \\
% 25  & \textbf{51.4} & 63.5 & \textbf{46.4} & 42.6 & 41.7 & 43.0 & \textbf{60.8} & 60.9 & \textbf{61.2} \\
% 50  & 46.7 & 58.9 & 41.6 & 39.9 & 38.3 & 40.7 & 54.4 & \underline{63.2} & 49.8 \\
% \midrule
% $\bar n/2^{*}$ & \underline{50.7} & \underline{67.6} & \underline{43.7} &
% \underline{53.4} & \underline{57.2} & \underline{51.6} &
% \underline{59.0} & \textbf{63.6} & 56.6 \\
% \bottomrule
% \end{tabular}}
% \vspace{-1.5em}
% \end{table}
\begin{table}[t]
\centering
\caption{\textbf{Sensitivity to $n_0$}.
\textbf{Bold} = best per column, \underline{underlined} = second-best.
Row $\bar n/2^{*}$ is our default rule $n_0=\min(\bar n/2,\,n_{\mathrm{cap}})$.}
\vspace{-0.5em}
\label{tab:sensitivity-n0}
\begingroup
\footnotesize
\setlength{\tabcolsep}{4pt}
\begin{adjustbox}{max width=0.9\linewidth, scale=0.92}
\begin{tabular}{lccccccccc}
\toprule
\multirow{2}{*}{$n_0$} &
\multicolumn{3}{c}{Animalia ($\bar n/2{=}19$)} &
\multicolumn{3}{c}{CUB ($\bar n/2{=}7$)} &
\multicolumn{3}{c}{OxfordPets ($\bar n/2{=}24$)} \\
\cmidrule(lr){2-4}\cmidrule(lr){5-7}\cmidrule(lr){8-10}
& All & Known & Novel & All & Known & Novel & All & Known & Novel \\
\midrule
5   & 40.7 & 60.7 & 32.5 & \textbf{55.4} & \textbf{60.0} & \textbf{53.2} & 58.5 & 62.0 & 56.6 \\
10  & 46.5 & 63.3 & 39.5 & 50.7 & 52.3 & 49.9 & 57.8 & 55.6 & 58.9 \\
15  & 48.0 & \textbf{69.7} & 39.0 & 47.6 & 47.4 & 46.1 & 58.4 & 55.4 & \underline{60.0} \\
20  & 49.9 & 66.6 & 42.9 & 46.2 & 43.0 & 48.8 & 58.7 & 61.2 & 57.5 \\
25  & \textbf{51.4} & 63.5 & \textbf{46.4} & 42.6 & 41.7 & 43.0 & \textbf{60.8} & 60.9 & \textbf{61.2} \\
50  & 46.7 & 58.9 & 41.6 & 39.9 & 38.3 & 40.7 & 54.4 & \underline{63.2} & 49.8 \\
\midrule
$\bar n/2^{*}$ & \underline{50.7} & \underline{67.6} & \underline{43.7} &
\underline{53.4} & \underline{57.2} & \underline{51.6} &
\underline{59.0} & \textbf{63.6} & 56.6 \\
\bottomrule
\end{tabular}
\end{adjustbox}
\endgroup
\vspace{-0.5em}
\end{table}

\begin{table}[t]
\centering
\caption{\textbf{Sensitivity to $\alpha$}. \textbf{Bold} = best per column, \underline{underlined} = second-best. The row marked ${}^*$ is our default.}
\vspace{-0.5em}
\label{tab:sensitivity-alpha}
\begingroup
\footnotesize
\setlength{\tabcolsep}{4pt}
\begin{adjustbox}{max width=0.9\linewidth, scale=0.92}
\begin{tabular}{lccccccccc}
\toprule
\multirow{2}{*}{$\log \alpha$} &
\multicolumn{3}{c}{Animalia} &
\multicolumn{3}{c}{CUB} &
\multicolumn{3}{c}{OxfordPets} \\
\cmidrule(lr){2-4}\cmidrule(lr){5-7}\cmidrule(lr){8-10}
& All & Known & Novel & All & Known & Novel & All & Known & Novel \\
\midrule
$-6$  & 48.8 & 65.4 & 41.9 & 52.4 & 53.8 & 51.7 & \underline{61.0} & 60.9 & \textbf{61.1} \\
$-7$  & 49.5 & 67.1 & 42.3 & 53.0 & 53.3 & \textbf{52.9} & \textbf{61.4} & 62.9 & \underline{60.6} \\
$-8$  & 50.2 & \textbf{68.5} & 42.7 & 52.8 & 55.0 & 51.6 & 59.7 & 63.2 & 57.9 \\
$-9^*$ & \underline{50.7} & \underline{67.6} & 43.7 & \underline{53.4} & \textbf{57.2} & 51.6 & 59.0 & 63.6 & 56.6 \\
$-10$ & \textbf{51.2} & 66.8 & \textbf{44.7} & 52.9 & 55.1 & 51.7 & 58.9 & 64.8 & 56.2 \\
$-11$ & 49.0 & 64.0 & 42.9 & 53.2 & \underline{56.8} & 51.4 & 59.0 & \underline{65.0} & 55.9 \\
$-12$ & 50.6 & 66.7 & \underline{43.9} & \textbf{53.5} & 56.1 & \underline{52.2} & 58.3 & \textbf{65.1} & 54.7 \\
\bottomrule
\end{tabular}
\end{adjustbox}
\endgroup
\vspace{-1.5em}
\end{table}

% \begin{table}[t]
% \centering
% \small
% \caption{\textbf{Runtime of DP-BOA.} Mean and max latency (ms / sample) over the full OCD stream on a single GPU. Numbers in parentheses denote the total number of categories (\#Cls).}
% \vspace{-0.5em}
% \label{tab:runtime}
% \setlength{\tabcolsep}{4pt}
% \resizebox{0.9\linewidth}{!}{
% \begin{tabular}{lccc}
% \toprule
% & Animalia (77) & CUB (200) & OxfordPets (38) \\
% \midrule
% Mean / Max latency (ms) & 26.6 / 46.1 & 58.7 / 81.7 & 17.4 / 37.0 \\
% \bottomrule
% \end{tabular}}
% \vspace{-1.5em}
% \end{table}

\begin{figure}[t]
    \centering

    \begin{minipage}[t]{0.52\linewidth}
        \centering
        \vspace{0pt}
        \includegraphics[width=\linewidth]{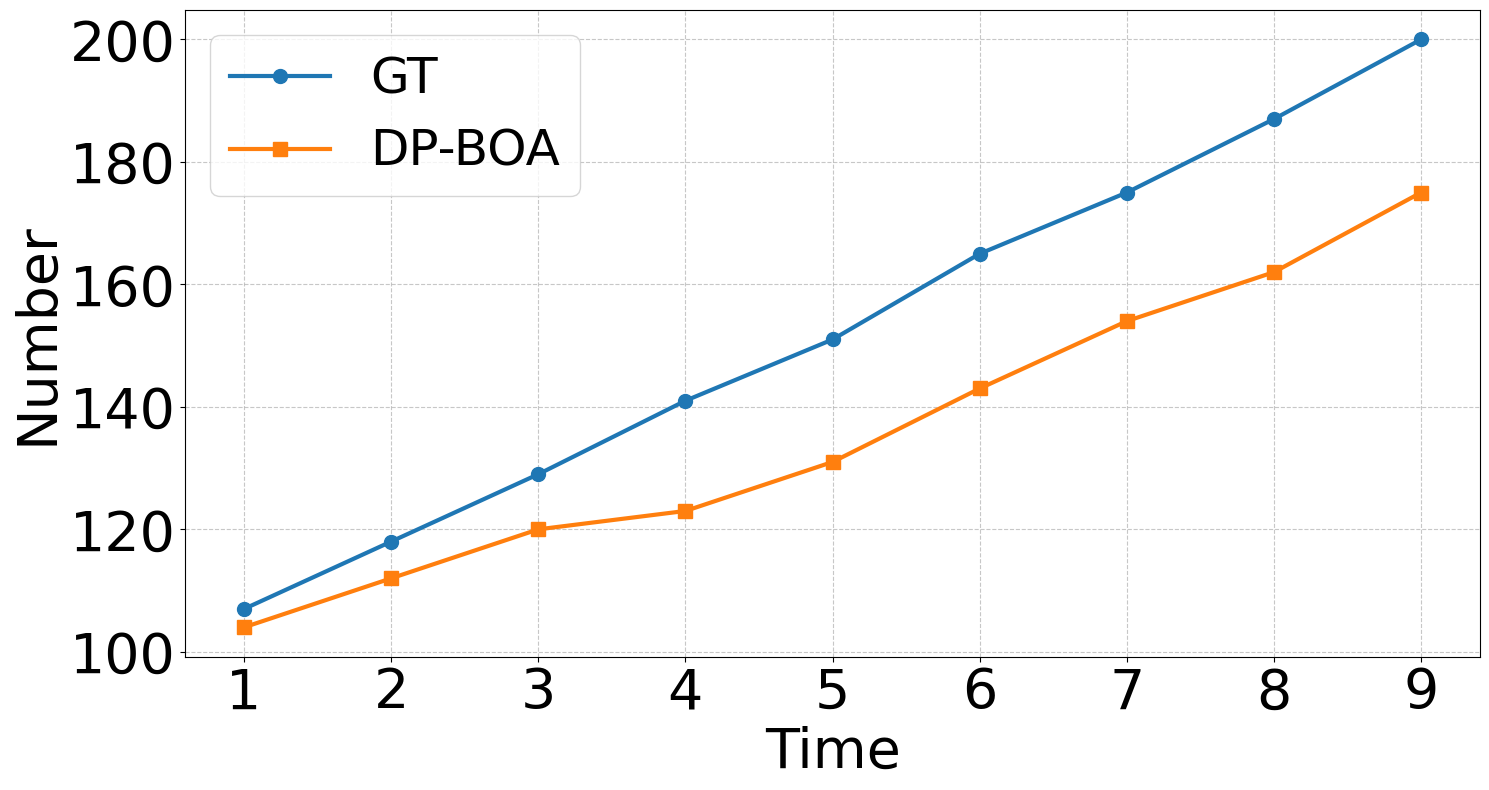}
        \vspace{-2em}
        \caption{Cluster growth on CUB.}
        \label{fig:cluster_number}
    \end{minipage}\hfill
    \begin{minipage}[t]{0.45\linewidth}
        \centering
        \vspace{0pt}
        \captionsetup{type=table,skip=1pt}
        \caption{Final cluster count.}
        \vspace{0.5em}
        \label{tab:kfinal_vs_gt}
    
        \setlength{\tabcolsep}{6pt}
        \renewcommand{\arraystretch}{1.1}
    
        \resizebox{\linewidth}{!}{%
        \begin{tabular}{lcc}
            \toprule
            Method & CUB & StanfordCars \\
            \midrule
            SMILE-16bit & 924 & 896 \\
            PHE-16bit   & 318 & 709 \\
            DiffGRE     & 116 & 171 \\
            Sync-AL     & 255 & 447 \\
            DP-BOA      & 175 & 177 \\
            Ground Truth & 200 & 196 \\
            \bottomrule
        \end{tabular}%
        }
    \end{minipage}

    \vspace{-0.5em}
\end{figure}

\vspace{-0.5em}
\paragraph{Cluster Growth and Estimated Number of Classes.}
\label{sec:clusternum}
We analyze the evolution of the number of discovered clusters along the OCD stream. \cref{fig:cluster_number} plots $K(t)$ on CUB, showing that DP-BOA exhibits smooth, controlled growth that broadly tracks the ground-truth trend while remaining slightly conservative. Consistently, \cref{tab:kfinal_vs_gt} reports $K_\mathrm{final}{=}175/177$ on CUB/StanfordCars versus $200/196$ ground-truth classes, substantially closer than other baselines, indicating a more calibrated birth rate without severe stream fragmentation.

\begin{table}[t]
\centering
\footnotesize
\setlength{\tabcolsep}{3pt}
\caption{\textbf{Runtime and accuracy--efficiency trade-off.} Accuracy, maximum per-sample latency (Lat.) in milliseconds, and head memory (Mem.) in MB. Numbers in parentheses denote the number of categories.}
\vspace{-0.5em}
\label{tab:dpboa_l}
\resizebox{1.0\textwidth}{!}{%
\begin{tabular}{l ccc cc ccc cc ccc cc}
\toprule
& \multicolumn{5}{c}{Animalia (77)} & \multicolumn{5}{c}{CUB (200)} & \multicolumn{5}{c}{StanfordCars (196)}\\
\cmidrule(lr){2-6}\cmidrule(lr){7-11}\cmidrule(lr){12-16}
Method & All & Known & Novel & Lat. & Mem.
       & All & Known & Novel & Lat. & Mem.
       & All & Known & Novel & Lat. & Mem. \\
\midrule
DP-BOA & \textbf{50.7} & \textbf{67.6} & \textbf{43.7} & 46.1 & 299.2
         & \textbf{53.4} & \textbf{57.2} & \textbf{51.6} & 81.7 & 619.2 
         & \textbf{32.4} & \textbf{58.8} & 19.7 & 82.9 & 626.7 \\
DP-BOA-L & 49.2 & 64.2 & 42.9 & \textbf{34.0} & \textbf{40.5}
         & 52.2 & 54.6 & 50.9 & \textbf{50.3} & \textbf{51.0} 
         & 31.5 & 54.7 & \textbf{20.2} & \textbf{44.2} & \textbf{49.7} \\
\bottomrule
\end{tabular}%
}
\vspace{-1em}
\end{table}

\paragraph{Efficiency.}
\label{sec:runtime_tradeoff}
To complement the theoretical complexity in \cref{sec:complexity}, we report the wall-clock cost and head memory footprint of DP-BOA and its low-rank variant in \cref{tab:dpboa_l}. Although full DP-BOA uses full-covariance posterior predictive and therefore incurs a quadratic memory cost in the feature dimension, its absolute overhead remains moderate in the OCD setting: the maximum per-sample latency is below 83 ms on the reported benchmarks, and the head memory remains below 0.7 GB. Overall, DP-BOA trades additional computation and memory for accuracy, but the resulting overhead remains moderate and practically acceptable in the OCD setting.

To further mitigate the cost of full-covariance modeling, we evaluate DP-BOA-L. As shown in \cref{tab:dpboa_l}, DP-BOA-L preserves most of the accuracy of full DP-BOA while substantially reducing both latency and head memory. Its All accuracy drops by only 1.5, 1.2, and 0.9 points on Animalia, CUB, and StanfordCars, respectively, while head memory is reduced from 299.2/619.2/626.7 MB to 40.5/51.0/49.7 MB and maximum latency decreases from 46.1/81.7/82.9 ms to 34.0/50.3/44.2 ms. The reduction is especially clear on CUB and StanfordCars, where the larger number of categories makes the full-covariance head more expensive. 
%Since DP-BOA-L stores only low-rank per-category covariance states plus shared overhead, its memory scales approximately linearly with the number of categories; the measured head memory remains around 50 MB even at nearly 200 categories. 
Overall, DP-BOA-L offers a favorable accuracy-efficiency trade-off and provides a practical approximation for larger-K or resource-constrained settings.

\paragraph{Robustness to stream order.}

% The standard OCD protocol follows prior work\cite{zheng2024prototypical} and evaluates all methods on the same query stream, enabling controlled comparison. To test whether DP-BOA relies on this particular ordering, we further construct bursty-correlated streams with local class bursts of size (B=10). Concretely, we first shuffle samples within each ground-truth class, split them into chunks of up to (B) samples, and then randomly shuffle these class-specific chunks to form the final query stream. All others are the same. This setting induces stronger local temporal correlation, but the model still processes samples sequentially without ground-truth labels. As shown in \cref{tab:stream_order}, DP-BOA outperforms PHE in \emph{All} and \emph{Novel} accuracy on all three datasets. The gains are particularly clear on CUB, where DP-BOA improves \emph{All}/\emph{Novel} accuracy by (16.7/19.5) points. These results suggest that DP-BOA's gains do not rely on a favorable random stream order; the posterior-predictive birth-or-assign rule remains effective under stronger local temporal correlation.

We use the standard OCD protocol following prior work~\cite{zheng2024prototypical} and evaluate all methods on the same query stream for controlled comparison. To test whether DP-BOA relies on this ordering, we construct bursty-correlated streams by shuffling samples within each ground-truth class, grouping them into chunks of up to $B=10$ samples, and then randomly shuffling these chunks, while keeping the same support/query split, query set, and evaluation metric. This induces stronger local temporal correlation than the standard protocol. As shown in \cref{tab:stream_order}, DP-BOA outperforms SMILE and PHE in \emph{All} and \emph{Novel} accuracy on all three datasets, with a particularly large gain on CUB $(+16.7/+19.5)$ points in \emph{All}/\emph{Novel} accuracy over PHE, suggesting that its gains do not rely on a favorable random stream order.

\begin{table}[t]
    \centering
    \caption{\textbf{Robustness to bursty-correlated stream orders.}
    We use local class bursts of size $B=10$ to induce strong temporal correlation. Best results are in \textbf{bold}.}
    \vspace{-1em}
    \label{tab:stream_order}
    \small
    \setlength{\tabcolsep}{5pt}
    \renewcommand{\arraystretch}{1.05}
    \begin{tabular}{lccccccccc}
        \toprule
        & \multicolumn{3}{c}{Pets}
        & \multicolumn{3}{c}{StanfordCars}
        & \multicolumn{3}{c}{CUB} \\
        \cmidrule(lr){2-4}
        \cmidrule(lr){5-7}
        \cmidrule(lr){8-10}
        Method
        & All & Known & Novel
        & All & Known & Novel
        & All & Known & Novel \\
        \midrule
        SMILE
        & 40.0 & 46.2 & 36.9 
        & 25.1 & 42.0 & 16.9
        & 30.9 & 50.0 & 21.2 \\
        PHE
        & 47.0 & 45.5 & 47.9
        & 30.5 & \textbf{62.0} & 15.3
        & 33.8 & 45.9 & 27.8 \\
        DP\mbox{-}BOA
        & \textbf{58.3} & \textbf{57.7} & \textbf{58.5}
        & \textbf{35.0} & 57.3 & \textbf{24.2}
        & \textbf{50.5} & \textbf{56.8} & \textbf{47.3} \\
        \bottomrule
    \end{tabular}
    \vspace{-1em}
\end{table}

% \vspace{-0.5em}
% \paragraph{Additional Analyses.}
% \label{sec:additional}
% Extended studies on efficiency (including memory and runtime), robustness, the dynamic evolution of discovered categories, and additional hyperparameter sensitivity results are provided in the Appendix.

\vspace{-2mm}
\section{Conclusion}
\label{sec:Conclusion}
\vspace{-1mm}

We revisited on-the-fly category discovery from the perspective of probabilistic decision-making.
Instead of relying on heuristic matching rules, DP-BOA formulates the assign-versus-birth choice as a Bayesian evidence comparison, combining a Dirichlet-process prior for open-ended category growth with NIW posterior predictives for geometry-aware and evidence-adaptive online decisions.
By leveraging labeled known classes to initialize both the global prior and known-category posteriors, DP-BOA provides a new framework for streaming category discovery and achieves strong performance across standard OCD benchmarks.

\vspace{-0.5em}
\paragraph{Limitations and future work.}
\label{sec:limitation}
DP-BOA is designed for the conventional OCD setting, where online decisions must be made efficiently without revisiting past samples.
First, given the limited auxiliary information in this protocol, we use a small set of hyperparameters and simple fixed defaults, a common practice in Bayesian nonparametric mixture models~\cite{escobar1995bayesian,gershman2012tutorial}; these defaults already yield strong empirical performance.
In richer scenarios, extra signals (e.g., prior knowledge about class granularity or semantic side information) could potentially enable more adaptive hyperparameter calibration.
Second, for efficiency, each category in our method is modeled by a single elliptical component, which provides a strong accuracy--efficiency trade-off on diverse benchmarks.
When streams exhibit more complex intra-class structure, such as multi-modality or non-elliptical geometry induced by domain shift, more complex and expressive category models (e.g., multi-component or non-Gaussian likelihoods) may further improve robustness~\cite{bishop2006pattern,rasmussen2000infinite}.
However, such expressiveness typically increases latency and memory due to additional state maintenance, and may also introduce extra parameters.
A future direction is to incorporate richer category models into our probabilistic framework while reducing these overheads.

\section*{Acknowledgements}
This work was supported by the MoE Key Lab of Intelligent Perception and Human-Machine
Collaboration (ShanghaiTech University), and the Shanghai Key Technology R\&D Program No. 25JC3200500.

\bibliographystyle{splncs04}
\bibliography{main}

\newpage
\appendix
\section*{Contents of the Appendix}
% \begin{itemize}[leftmargin=*, itemsep=0.25em]
\begin{itemize}

  % \item \hyperref[app:dataset]{\textbf{Appendix A: Implementation Details}}\\
  % Per-dataset statistics, known/novel splits, and support/query construction for all benchmarks, baseline fairness.
  \item \hyperref[app:dataset]{\textbf{Appendix A: Implementation Details}}\\ 
    Per-dataset statistics, known/novel splits, support/query construction, and baseline fairness.

    \item \hyperref[app:std_results]{\textbf{Appendix B: Standard Deviations}}\\ 
    Standard deviations over three runs for all benchmarks and evaluation metrics.

  \item \hyperref[app:appendix_niw]{\textbf{Appendix C: NIW Update Equations}}\\
    Definition of the Normal--Inverse--Wishart prior and derivation of the closed-form posterior updates.

  \item \hyperref[app:appendix_student_t]{\textbf{Appendix D: Student-\texorpdfstring{$t$}{t} Predictive Densities}}\\
    Derivation of the multivariate Student-\(t\) predictives for existing and new categories, plus a 1D illustration of their behavior as evidence grows.

  \item \hyperref[app:appendix_dp_rule]{\textbf{Appendix E: Dirichlet--Process Predictive Rule}}\\
    Derivation of the DP class-prior over category indices and the corresponding birth probability used in DP-BOA.

  \item \hyperref[app:appendix_eb_niw]{\textbf{Appendix F: Estimation of NIW Hyperparameters}}\\
  % Empirical–Bayes estimates of \(\mu_0,\Psi_0,\kappa_0\) from labeled support data.
    Empirical-Bayes-inspired moment matching for estimating \(\mu_0\), \(\Psi_0\), and \(\kappa_0\) from labeled support data.

  \item \hyperref[app:appendix_online_stats]{\textbf{Appendix G: Online Update of Sufficient Statistics}}\\
    Welford-style streaming updates for per-category counts, means, and scatters \((n_k,\bar{\mathbf{z}}_k,\mathbf{S}_k)\).

  \item \hyperref[app:stronger_encoder]{\textbf{Appendix H: Stronger Encoder Analysis}}\\
    Additional stronger-encoder evaluation with supervised contrastive learning while keeping the DP\mbox{-}BOA online head unchanged.

  \item \hyperref[app:appendix_qualitative]{\textbf{Appendix I: Category Decision Boundary Visualization}}\\
    2D PCA visualizations of Student-\(t\) decision ellipses for known and novel DP-BOA categories.

  \item \hyperref[app:appendix_pseudo]{\textbf{Appendix J: Pseudocode for DP-BOA}}\\
    Pseudocode for the offline initialization and online birth-or-assign loop.

  % \item \hyperref[app:appendix_runtime]{\textbf{Appendix I: Complexity Analysis}}\\
  % Time/memory complexity of the DP-BOA head, memory footprint study, and a low-rank variant DP-BOA-L for improved scalability.
  \item \hyperref[app:appendix_runtime]{\textbf{Appendix K: Complexity Analysis}}\\ 
    Time and memory complexity of the full DP-BOA head, memory footprint analysis, and the low-rank DP-BOA-L variant for improved scalability.

  \item \hyperref[app:appendix_hparams]{\textbf{Appendix L: More Hyperparameters Analysis}}\\
    Ablations and sensitivity analysis for the NIW mean-strength \(\kappa_0\) and the degrees-of-freedom cap \(n_{\mathrm{cap}}\).

  \item \hyperref[app:appendix_geometry]{\textbf{Appendix M: Feature Geometry Analysis}}\\
    % Statistics of category geometry, empirical support for Gaussian modeling, and discussion of extensions beyond single-Gaussian assumptions.
    Empirical category-geometry statistics, analysis of the Gaussian approximation, and support--novel mismatch stress tests.

  % \item \hyperref[app:appendix_robust]{\textbf{Appendix L: Comparison with a Fixed-Threshold Variant}}\\
  %   Comparison between DP-BOA and a fixed-threshold variant.
  \item \hyperref[app:appendix_robust]{\textbf{Appendix N: Comparison with Simpler Probabilistic Heads}}\\ 
    Comparison between DP-BOA and simpler heads under identical frozen features, including posterior-threshold and Mahalanobis-threshold variants.
  
  % \item \hyperref[app:appendix_future]{\textbf{Appendix M: Future Work}}\\
  % Discussion of future directions.
  % \item \hyperref[app:stream_order]{\textbf{Appendix M: Robustness to Stream Order}}\\ 
  %   Additional evaluation under bursty-correlated streams to test robustness beyond the standard OCD stream order.

  \item \hyperref[app:temporal_diagnostics]{\textbf{Appendix O: Temporal Diagnostics}}\\ 
    Stream-level diagnostics of DP-BOA, including false-birth rates, cluster-mean drift, and evidence margin.

  % \item \hyperref[app:std_results]{\textbf{Appendix N: Standard Deviations}}\\ 
  %   Standard deviations over three runs for all benchmarks and evaluation metrics.

\end{itemize}
\newpage

\section{Implementation Details}
\label{app:dataset}

\subsection{Dataset Details}
\label{app:dataset_details}

We evaluate DP-BOA on ten OCD benchmarks. For CIFAR100, ImageNet100, and Herbarium19, we adopt the official splits from On-the-Fly Category Discovery~\cite{du2023fly}. For the remaining seven fine-grained benchmarks---CUB, Stanford Cars, Oxford-IIIT Pet, and four iNaturalist 2017~\cite{van2018inaturalist} super-categories, namely Fungi, Arachnida, Animalia, and Mollusca---we follow the class splits used by PHE~\cite{zheng2024prototypical}. 

Following the standard OCD protocol~\cite{du2023fly,zheng2024prototypical}, for each known class \(c \in \mathcal{Y}_S\), we use \(50\%\) of its images as the labeled support set \(\mathcal{D}_S\), and place the remaining \(50\%\) into the query stream \(\mathcal{D}_Q\). All images from novel classes \(\mathcal{Y}_Q \setminus \mathcal{Y}_S\) are used only in \(\mathcal{D}_Q\). \cref{tab:dataset_stats} summarizes the dataset statistics.

\begin{table*}[t]
    \centering
    \caption{Dataset statistics. \(|\mathcal{Y}_S|\) / \(|\mathcal{Y}_Q|\) denote the numbers of known / all classes, and \(|\mathcal{D}_S|\) / \(|\mathcal{D}_Q|\) denote the sizes of the labeled support set / unlabeled query stream.}
    \vspace{-0.5em}
    \label{tab:dataset_stats}
    \resizebox{\textwidth}{!}{
        \begin{tabular}{lcccccccccc}
            \toprule
            & CIFAR100 & ImageNet100 & CUB & Cars & Herb19 &
            Pets & Fungi & Arachnida & Animalia & Mollusca \\
            \midrule
            \(|\mathcal{Y}_S|\) &
            80 & 50 & 100 & 98 & 341 &
            19 & 61 & 28 & 39 & 47 \\
            \(|\mathcal{Y}_Q|\) &
            100 & 100 & 200 & 196 & 683 &
            38 & 121 & 56 & 77 & 93 \\
            \midrule
            \(|\mathcal{D}_S|\) &
            20.0K & 31.9K & 1.5K & 2.0K & 8.9K &
            0.9K & 1.8K & 1.7K & 1.5K & 2.4K \\
            \(|\mathcal{D}_Q|\) &
            30.0K & 95.3K & 4.5K & 6.1K & 25.4K &
            2.7K & 5.8K & 4.3K & 5.1K & 7.0K \\
            \bottomrule
        \end{tabular}
    }
    \vspace{-1em}
\end{table*}

\vspace{-1em}
\subsection{Baseline Fairness}
\label{app:baseline_fairness}

For a controlled comparison, we follow the standard evaluation setting used in prior OCD methods. All compared methods are evaluated with the same DINO backbone family, except Sync~\cite{banerjeelanguage}, which uses CLIP features and language information. Our encoder is fine-tuned only with standard supervised cross-entropy on the labeled support set, whereas several baselines employ additional method-specific representation-learning objectives, such as hashing regularization in SMILE~\cite{du2023fly}. 

We also compare against prior methods in their standard forms. Many existing OCD heads are based on hash codes, prototype matching, or thresholded similarity rules, and are not designed to maintain full-covariance category statistics online. Retrofitting these methods with full-covariance predictive modeling would require non-trivial algorithmic changes and would no longer correspond to their original implementations. We therefore report their standard results and isolate the contribution of our probabilistic head through additional same-feature comparisons in \cref{app:appendix_robust}.

\begin{table}[t]
    \centering
    \caption{\textbf{Standard deviations over three runs.}
    We report the standard deviations of DP\mbox{-}BOA for \emph{All}, \emph{Known}, and \emph{Novel} accuracy on all benchmarks. Values are measured in percentage points.}
    \label{tab:std_results}
    \small
    \setlength{\tabcolsep}{6pt}
    \renewcommand{\arraystretch}{0.98}
    \begin{tabular}{lccc}
        \toprule
        Dataset & All & Known & Novel \\
        \midrule
        Animalia     & 1.41 & 0.24 & 2.04 \\
        Arachnida    & 0.72 & 5.21 & 2.13 \\
        CIFAR100     & 0.19 & 1.24 & 2.11 \\
        CUB          & 1.40 & 1.16 & 2.63 \\
        Fungi        & 1.09 & 2.00 & 1.57 \\
        Herbarium19  & 0.38 & 0.99 & 0.22 \\
        ImageNet100  & 0.89 & 0.57 & 1.05 \\
        Mollusca     & 0.60 & 0.73 & 0.54 \\
        OxfordPets   & 2.17 & 2.60 & 3.28 \\
        StanfordCars & 0.54 & 1.66 & 1.17 \\
        \bottomrule
    \end{tabular}
    \vspace{0.5em}
\end{table}

\section{Standard Deviations}
\label{app:std_results}

In the main paper, we report the mean accuracy over three runs with different random seeds. For completeness, \cref{tab:std_results} reports the corresponding standard deviations of DP\mbox{-}BOA on all benchmarks. All values are measured in percentage points. The standard deviations are generally small for \emph{All} accuracy, indicating that the overall performance is stable across runs. 
%Some \emph{Known} or \emph{Novel} metrics show larger variation, which is expected in OCD because stream order, online birth decisions, and Hungarian matching can affect the known/novel split differently.

\section{NIW Update Equations}
\label{app:appendix_niw}
In this section, we first introduce the Normal--Inverse--Wishart (NIW)
distribution and its parameters.
We then derive step by step that the NIW prior is conjugate to a
multivariate Gaussian with unknown mean and covariance, and obtain the
closed-form update equations used in DP-BOA.

\vspace{-1em}
\subsection{The Normal--Inverse--Wishart Distribution}

The Normal--Inverse--Wishart (NIW) distribution is a four-parameter family
that serves as the conjugate prior for a multivariate Gaussian
$\mathcal{N}(\mu,\Sigma)$ with unknown mean $\mu$ and covariance
$\Sigma$.
It is defined as the product of a Normal distribution over the mean
and an Inverse-Wishart distribution over the covariance:
\begin{equation}
\label{eq:niw_def}
% \resizebox{1.0\linewidth}{!}{$
    \displaystyle
    \mathrm{NIW}(\mu,\Sigma \mid \mu_0,\kappa_0,\Psi_0,\nu_0)
    \triangleq
    \mathcal{N}\!\left(\mu \mid \mu_0, \tfrac{1}{\kappa_0}\Sigma\right)
    \times
    \mathcal{W}^{-1}(\Sigma \mid \Psi_0,\nu_0)
% $}
\end{equation}
where we use $\mathcal{W}^{-1}(\cdot \mid \Psi_0,\nu_0)$ to denote the
Inverse-Wishart distribution with scale matrix $\Psi_0$ and degrees of
freedom $\nu_0$.
Its density is
\begin{equation}
\label{eq:iw_def}
% \resizebox{1.0\linewidth}{!}{$
    \displaystyle
    \mathcal{W}^{-1}(\Sigma \mid \Psi_0,\nu_0)
    =
    \frac{|\Psi_0|^{\frac{\nu_0}{2}}}
         {2^{\frac{\nu_0 d}{2}} \Gamma_d(\frac{\nu_0}{2})}
    |\Sigma|^{-\frac{\nu_0 + d + 1}{2}}
    \exp\!\left(
        -\frac{1}{2} \mathrm{tr}(\Psi_0 \Sigma^{-1})
    \right)
% $}
\end{equation}
where $d$ is the feature dimension and $\Gamma_d(\cdot)$ is the multivariate
gamma function.

Combining \cref{eq:niw_def} and \cref{eq:iw_def}, and dropping
normalization constants that do not depend on $(\mu,\Sigma)$, the prior
density $p(\mu,\Sigma)$ can be written as
\begin{equation}\resizebox{1.0\linewidth}{!}{$
    p(\mu,\Sigma)
    \propto
    |\Sigma|^{-\frac{\nu_0 + d + 2}{2}}
    \exp\!\left(
        -\frac{1}{2} \mathrm{tr}(\Psi_0 \Sigma^{-1})
    \right)
    \exp\!\left(
        -\frac{\kappa_0}{2}
        (\mu - \mu_0)^\top
        \Sigma^{-1}
        (\mu - \mu_0)
    \right)$}
\end{equation}

\noindent
The four hyperparameters have intuitive interpretations:
\begin{itemize}
    \item $\mu_0 \in \mathbb{R}^d$: prior mean (the expected center);
    \item $\kappa_0 > 0$: prior strength (effective pseudo-count);
    \item $\Psi_0 \in \mathbb{R}^{d \times d}$: prior scale matrix;
    \item $\nu_0 > d - 1$: degrees of freedom of the Inverse-Wishart.
\end{itemize}

\vspace{-1em}
\subsection{Estimation detail of Posterior Updates}

Consider a category $k$ with $n_k$ observed feature vectors
$\mathcal{D}_k = \{z_1,\dots,z_{n_k}\}$.
As described in \cref{sec:GMMNIW}, we assume
$z_i \sim \mathcal{N}(\mu,\Sigma)$.
The likelihood of the data (up to a constant) is
\begin{equation}
% \resizebox{1.0\linewidth}{!}{$
    \displaystyle
    p(\mathcal{D}_k \mid \mu,\Sigma)
    \propto
    |\Sigma|^{-\frac{n_k}{2}}
    \exp\!\left(
        -\frac{1}{2}
        \sum_{i=1}^{n_k}
        (z_i - \mu)^\top \Sigma^{-1} (z_i - \mu)
    \right)
% $}
\end{equation}

To facilitate combination with the prior, we rewrite the quadratic form
using the sample mean
$\bar{z}_k = \frac{1}{n_k} \sum_i z_i$
and the scatter matrix
$S_k = \sum_i (z_i - \bar{z}_k)(z_i - \bar{z}_k)^\top$:
\vspace{-1em}
\begin{equation}
\label{eq:likelihood_decomp}
% \resizebox{1.0\linewidth}{!}{$
    \displaystyle
    \sum_{i=1}^{n_k}
    (z_i - \mu)^\top \Sigma^{-1} (z_i - \mu)
    =
    \mathrm{tr}(S_k \Sigma^{-1})
    +
    n_k (\bar{z}_k - \mu)^\top \Sigma^{-1} (\bar{z}_k - \mu)
% $}
\end{equation}

By Bayes' rule, the posterior (again up to a constant) is
\begin{equation}
% \resizebox{0.9\linewidth}{!}{$
    \displaystyle
    p(\mu,\Sigma \mid \mathcal{D}_k)
    =
    \frac{p(\mathcal{D}_k \mid \mu,\Sigma)\, p(\mu,\Sigma)}
         {p(\mathcal{D}_k)}
    \propto
    p(\mathcal{D}_k \mid \mu,\Sigma)\, p(\mu,\Sigma)
% $}
\end{equation}
Multiplying the likelihood and prior, and using
\cref{eq:likelihood_decomp}, we obtain
\begin{equation}
% \resizebox{1.0\linewidth}{!}{$
    \displaystyle
    \begin{aligned}
    p(\mu,\Sigma \mid \mathcal{D}_k)
    &\propto
    |\Sigma|^{-\frac{(\nu_0 + n_k) + d + 2}{2}}
    \exp\!\left(
        -\frac{1}{2}
        \mathrm{tr}\bigl((\Psi_0 + S_k)\Sigma^{-1}\bigr)
    \right)
    \\
    &\quad \times
    \exp\!\left(
        -\frac{1}{2}
        \Big[
            \kappa_0 \|\mu - \mu_0\|^2_{\Sigma^{-1}}
            +
            n_k \|\mu - \bar{z}_k\|^2_{\Sigma^{-1}}
        \Big]
    \right)
    \end{aligned}
% $}
\end{equation}
where we use the shorthand
$\|x\|^2_{\Sigma^{-1}} = x^\top \Sigma^{-1} x$.

The quadratic term in $\mu$ can be simplified by completing the square:
\begin{equation}
\label{eq:square_completion}
% \resizebox{1.0\linewidth}{!}{$
    \displaystyle
    \begin{aligned}
    &\kappa_0 \|\mu - \mu_0\|^2_{\Sigma^{-1}}
    + n_k \|\mu - \bar{z}_k\|^2_{\Sigma^{-1}}
    =\\&
    (\kappa_0 + n_k)
    \left\|
        \mu - \frac{\kappa_0 \mu_0 + n_k \bar{z}_k}{\kappa_0 + n_k}
    \right\|^2_{\Sigma^{-1}}
    \quad +
    \frac{\kappa_0 n_k}{\kappa_0 + n_k}
    \|\bar{z}_k - \mu_0\|^2_{\Sigma^{-1}}
    \end{aligned}
% $}
\end{equation}

Substituting \cref{eq:square_completion} back into the posterior and
collecting terms, we can recognize the form of another NIW
distribution with updated parameters.

\vspace{-1em}
\subsection{Final Update Equations}

Matching the posterior to the NIW form, we obtain the update rules
used in the main paper.
For a category $k$ with sufficient statistics
$\{n_k, \bar{z}_k, S_k\}$, the posterior
$
    (\mu,\Sigma) \mid \mathcal{D}_k
    \sim
    \mathrm{NIW}(\mu_k,\kappa_k,\Psi_k,\nu_k)
$
has hyperparameters
\begin{align}
    \kappa_k &= \kappa_0 + n_k, \\
    \nu_k    &= \nu_0 + n_k, \\
    \mu_k    &= \frac{\kappa_0 \mu_0 + n_k \bar{z}_k}{\kappa_0 + n_k}, \\
    \Psi_k   &= \Psi_0
    + S_k
    + \frac{\kappa_0 n_k}{\kappa_0 + n_k}
      (\bar{z}_k - \mu_0)(\bar{z}_k - \mu_0)^\top.
\end{align}
These closed-form updates allow DP-BOA to maintain the exact NIW
posterior for each category using only $(n_k, \bar{z}_k, S_k)$.

\section{Student-\texorpdfstring{$t$}{t} Predictive Densities}
\label{app:appendix_student_t}
In this section, we derive the closed-form predictive distributions
obtained by marginalizing the Gaussian parameters $(\mu,\Sigma)$ under
the Normal--Inverse--Wishart (NIW) prior/posterior.
As shown in Appendix~\ref{app:appendix_niw}, conditioning on a category
$k$ with sufficient statistics $\{n_k,\bar{z}_k,S_k\}$ yields the NIW
posterior
\begin{equation}
    (\mu,\Sigma) \mid \mathcal{D}_k
    \sim
    \mathrm{NIW}(\mu_k,\kappa_k,\Psi_k,\nu_k),
\end{equation}
where the updated hyperparameters
$(\mu_k,\kappa_k,\Psi_k,\nu_k)$ are given in
Appendix~\ref{app:appendix_niw}.
We show that integrating out $(\mu,\Sigma)$ leads to multivariate
Student-$t$ predictive densities, which correspond to the
category-wise likelihood terms used in \cref{sec:GMMNIW}.

\vspace{-1em}
\subsection{Multivariate Student-\texorpdfstring{$t$}{t} Distribution}

We briefly recall the formulation of the multivariate Student-$t$
distribution.
For a $d$-dimensional random vector $z \in \mathbb{R}^d$, the
Student-$t$ density with location $\mu \in \mathbb{R}^d$,
positive-definite scale matrix $\Lambda \in \mathbb{R}^{d\times d}$,
and degrees of freedom $\nu > 0$ is
\begin{equation}
\label{eq:student_t_def}
% \resizebox{1.0\linewidth}{!}{$
    \displaystyle
    t_d(z \mid \mu,\Lambda,\nu)
    =
    \frac{
        \Gamma\!\left(\tfrac{\nu + d}{2}\right)
    }{
        \Gamma\!\left(\tfrac{\nu}{2}\right)
        (\nu\pi)^{d/2}
        \,|\Lambda|^{1/2}
    }
    \left[
        1
        +
        \frac{1}{\nu}
        (z - \mu)^\top
        \Lambda^{-1}
        (z - \mu)
    \right]^{-\frac{\nu + d}{2}},
% $}
\end{equation}
where $\Gamma(\cdot)$ is the gamma function.

\vspace{-1em}
\subsection{Posterior Predictive for an Existing Category}

Consider a category $k$ with NIW posterior
$(\mu,\Sigma) \mid \mathcal{D}_k
\sim \mathrm{NIW}(\mu_k,\kappa_k,\Psi_k,\nu_k)$.
We are interested in the predictive density of a new feature
$z \in \mathbb{R}^d$ given $\mathcal{D}_k$:
\begin{equation}
    p(z \mid \mathcal{D}_k)
    =
    \iint
    p(z \mid \mu,\Sigma)\,
    p(\mu,\Sigma \mid \mathcal{D}_k)
    \,\mathrm{d}\mu\,\mathrm{d}\Sigma,
\end{equation}
where $p(z \mid \mu,\Sigma) = \mathcal{N}(z \mid \mu,\Sigma)$.
Since $z$ is conditionally independent of $\mathcal{D}_k$ given
$(\mu,\Sigma)$, we write $p(z \mid \mu,\Sigma)$ instead of
$p(z \mid \mu,\Sigma,\mathcal{D}_k)$, and keep the dependence on
$\mathcal{D}_k$ only in $p(\mu,\Sigma \mid \mathcal{D}_k)$.

\paragraph{Step 1: Integrating out \texorpdfstring{$\mu$}{mu}.}
Using the hierarchical form of the NIW posterior,
\begin{align}
    \Sigma \mid \mathcal{D}_k
    &\sim \mathcal{W}^{-1}(\Psi_k,\nu_k), \\
    \mu \mid \Sigma,\mathcal{D}_k
    &\sim \mathcal{N}\!\left(\mu_k,\tfrac{1}{\kappa_k}\Sigma\right),
\end{align}
the conditional predictive density given $\Sigma$ is
\begin{equation}
    p(z \mid \Sigma,\mathcal{D}_k)
    =
    \int
    \mathcal{N}(z \mid \mu,\Sigma)\,
    \mathcal{N}\!\left(\mu \mid \mu_k,\tfrac{1}{\kappa_k}\Sigma\right)
    \,\mathrm{d}\mu.
\end{equation}
This is a convolution of two Gaussians with shared covariance structure,
which yields another Gaussian:
\begin{equation}
\label{eq:cond_pred_gaussian}
    p(z \mid \Sigma,\mathcal{D}_k)
    =
    \mathcal{N}\!\left(
        z \mid \mu_k,
        \left(1 + \tfrac{1}{\kappa_k}\right)\Sigma
    \right).
\end{equation}

\paragraph{Step 2: Integrating out \texorpdfstring{$\Sigma$}{Sigma}.}
We now integrate out $\Sigma \sim \mathcal{W}^{-1}(\Psi_k,\nu_k)$:
\begin{equation}
    % \resizebox{1.0\linewidth}{!}{$
    p(z \mid \mathcal{D}_k)
    =
    \int
    \mathcal{N}\!\left(
        z \mid \mu_k,
        \left(1 + \tfrac{1}{\kappa_k}\right)\Sigma
    \right)
    \mathcal{W}^{-1}(\Sigma \mid \Psi_k,\nu_k)
    \,\mathrm{d}\Sigma.
% $}
\end{equation}
It is helpful to recall the general relationship between
Gaussian--Inverse-Wishart mixtures and the multivariate Student-$t$
distribution.
Let
\begin{equation}
\label{eq:giw_general}
\begin{aligned}
\Sigma &\sim \mathcal{W}^{-1}(\Psi,\nu), \\
z \mid \Sigma &\sim \mathcal{N}(z \mid \mu, c\,\Sigma),
\end{aligned}
\end{equation}
for some scalar $c>0$.
A standard calculation (see, e.g.,
\cite{murphy2012machine,bishop2006pattern}) shows that the marginal
distribution of $z$ is multivariate Student-$t$:
\begin{equation}
\label{eq:giw_to_t}
    z \sim
    t_d\!\left(
        \mu,\;
        \frac{c}{\nu - d + 1}\,\Psi,\;
        \nu - d + 1
    \right).
\end{equation}
Intuitively, the Inverse-Wishart prior over $\Sigma$ plays the role of a
scale mixture over Gaussian covariances, and integrating out $\Sigma$
yields heavy-tailed Student-$t$ marginals.

Comparing \cref{eq:cond_pred_gaussian} with the general form
\cref{eq:giw_general}, we identify
\[
    c = 1 + \tfrac{1}{\kappa_k}, \quad
    \mu = \mu_k, \quad
    \Psi = \Psi_k, \quad
    \nu = \nu_k.
\]
Substituting these into \cref{eq:giw_to_t}, and rewriting the scale
matrix in the form used in \cref{eq:student_t_def}, we obtain
\begin{equation}
\label{eq:post_pred_t}
    p(z \mid \mathcal{D}_k)
    =
    t_d\!\left(
        z \,\middle|\,
        \mu_k,\;
        \frac{\kappa_k + 1}{\kappa_k(\nu_k - d + 1)}\Psi_k,\;
        \nu_k - d + 1
    \right).
\end{equation}
That is, the predictive distribution for category $k$ is a
Student-$t$ with location $\mu_k$, degrees of freedom
$\nu_k - d + 1$, and scale matrix
$\frac{\kappa_k + 1}{\kappa_k(\nu_k - d + 1)}\Psi_k$.

\vspace{-1em}
\subsection{Prior Predictive for a New Category}

For the ``birth'' hypothesis, we require a predictive density for a new
category whose parameters have not yet been updated by any data.
In DP-GMM, such a category is represented only by the
global NIW hyperparameters $(\mu_0,\kappa_0,\Psi_0,\nu_0)$, which encode
our prior belief about a generic category before seeing any of its
samples.
Equivalently, a hypothetical new category with zero observations
($n_k = 0$) has posterior equal to this prior:
\[
(\mu,\Sigma) \mid \text{new}
\;\sim\;
\mathrm{NIW}(\mu_0,\kappa_0,\Psi_0,\nu_0).
\]

The predictive density for a feature $z$ under this new category is
therefore the NIW \emph{prior predictive}, obtained by marginalizing out
$(\mu,\Sigma)$ under this global prior:
\begin{equation}
\label{eq:new_prior_pred_def}
% \resizebox{1.0\linewidth}{!}{$
    p(z \mid \text{new})
    =
    \iint
    p(z \mid \mu,\Sigma)\,
    p(\mu,\Sigma \mid \mu_0,\kappa_0,\Psi_0,\nu_0)
    \,\mathrm{d}\mu\,\mathrm{d}\Sigma,
% $}
\end{equation}
where $p(z \mid \mu,\Sigma) = \mathcal{N}(z \mid \mu,\Sigma)$ and
$(\mu,\Sigma) \sim \mathrm{NIW}(\mu_0,\kappa_0,\Psi_0,\nu_0).$
In other words, the ``new'' category reuses exactly the same
Gaussian–NIW hierarchy as in the posterior case, but with the
hyperparameters $(\mu_k,\kappa_k,\Psi_k,\nu_k)$ replaced by the global
prior $(\mu_0,\kappa_0,\Psi_0,\nu_0)$.

Repeating the same two-step marginalization as above but now with the
prior parameters, we obtain the prior predictive density
\begin{equation}
\label{eq:prior_pred_t}
    p(z \mid \text{new})
    =
    t_d\!\left(
        z \,\middle|\,
        \mu_0,\;
        \frac{\kappa_0 + 1}{\kappa_0(\nu_0 - d + 1)}\Psi_0,\;
        \nu_0 - d + 1
    \right).
\end{equation}
In other words, the ``birth'' likelihood is also a multivariate
Student-$t$ with location $\mu_0$, degrees of freedom
$\nu_0 - d + 1$, and scale matrix
$\frac{\kappa_0 + 1}{\kappa_0(\nu_0 - d + 1)}\Psi_0$.

\vspace{-1em}
\subsection{Qualitative Behavior of Student-\texorpdfstring{$t$}{t} Predictives}

The multivariate Student-\(t\) distribution
$
t_d(z \mid \mu,\Lambda,\nu)
$
is controlled by three parameters: the location \(\mu\), the positive-definite
scale matrix \(\Lambda\), and the degrees of freedom \(\nu\).
In this appendix we focus on how \(\nu\) alone shapes the density when
\(\mu\) and \(\Lambda\) are fixed.
For small \(\nu\), the distribution has heavy, power-law tails and a
relatively low central peak; as \(\nu\) increases, probability mass moves
towards the center, the tails decay faster, and the Student-\(t\) gradually
approaches the corresponding Gaussian \(\mathcal{N}(\mu,\Lambda)\).
In the limit \(\nu \to \infty\) the two coincide.

\begin{figure}[t]
    \centering
    \includegraphics[width=0.8\linewidth]{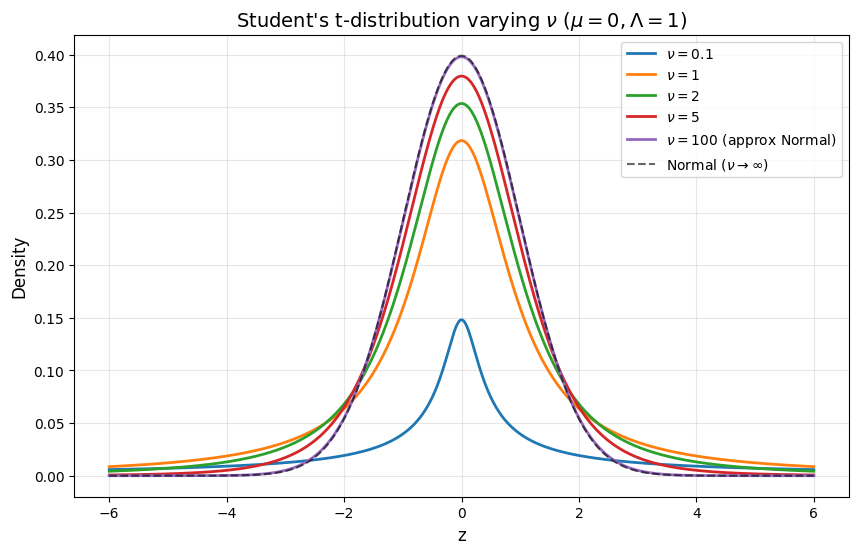}
    \caption{One-dimensional Student-\protect\(t\) densities
    \(t_1(z \mid \mu,\Lambda,\nu)\) with \(\mu{=}0\) and \(\Lambda{=}1\)
    for different degrees of freedom
    \(\nu \in \{0.1,1,2,5,100\}\), together with the standard normal
    \(\mathcal{N}(0,1)\) (dashed) corresponding to the limit
    \(\nu \to \infty\).
    As \(\nu\) increases, the distribution becomes more concentrated around
    \(\mu\) and its tails become lighter, eventually matching the Gaussian
    shape.}
    \label{fig:student_t_slices}
    \vspace{-1em}
\end{figure}

To visualize this effect, we consider the one-dimensional case \(d{=}1\)
and fix \(\mu{=}0\) and \(\Lambda{=}1\).
\cref{fig:student_t_slices} plots \(t_1(z \mid 0,1,\nu)\) for
\(\nu \in \{0.1,1,2,5,100\}\), together with the standard normal
\(\mathcal{N}(0,1)\) as the Gaussian limit.
When \(\nu\) is very small (e.g., \(\nu{=}0.1\)), the density has a flatter, less peaked center and extremely heavy tails.
As \(\nu\) increases to \(1\), \(2\), and \(5\), the peak at the center becomes higher and more concentrated while the tails become lighter.
By \(\nu{=}100\), the Student-\(t\) curve is almost indistinguishable from
\(\mathcal{N}(0,1)\), illustrating how increasing the degrees of freedom
removes heavy tails and yields a more Gaussian-like predictive.

Importantly, the predictive densities used in DP-BOA have exactly this
Student-\(t\): for category \(k\) we have
a predictive with degrees of freedom
\(\nu_k' = \nu_k - d + 1\), where \(\nu_k = \nu_0 + n_k\) grows with the
cluster size \(n_k\).
Thus, as more samples are assigned to a category, \(\nu_k'\) increases,
its predictive becomes more concentrated around \(\mu_k\) with lighter
tails, and the corresponding decision region of DP-BOA becomes sharper
and more confident.

\section{Dirichlet--Process Predictive Rule}
\label{app:appendix_dp_rule}

In this appendix, we derive the Dirichlet--process (DP) predictive rule
that gives the class--prior terms
$P(c_t \mid \mathcal{D}_{t-1})$ used in the main text.
We start from the definition of a DP and then derive the
predictive distribution step by step.

\vspace{-1em}
\subsection{Dirichlet Process: Definition}

Let $\Theta$ be the parameter space for component parameters
(e.g., Gaussian/NIW parameters in our mixture model).
A Dirichlet process $\mathrm{DP}(\alpha,G_0)$ is a distribution over
random probability measures $G$ on $\Theta$ such that for any finite
measurable partition
$(A_1,\dots,A_M)$ of $\Theta$ we have
\begin{equation}
\label{eq:dp_def}
    \bigl(G(A_1),\dots,G(A_M)\bigr)
    \sim
    \mathrm{Dir}\bigl(
        \alpha G_0(A_1),\dots,\alpha G_0(A_M)
    \bigr),
\end{equation}
where $\alpha > 0$ is the concentration parameter and $G_0$ is a fixed
\emph{base measure} on $\Theta$.
The Dirichlet distribution has density
\begin{equation}
    \mathrm{Dir}(x_1,\dots,x_M \mid \beta_1,\dots,\beta_M)
    \propto
    \prod_{m=1}^M x_m^{\beta_m - 1},
    \qquad x_m \ge 0,\ \sum_m x_m = 1
\end{equation}
with parameters $\beta_m > 0$.

A useful fact is that if
\[
(X_1,\dots,X_M)
\sim \mathrm{Dir}(\beta_1,\dots,\beta_M),
\]
then the expectation is
\begin{equation}
\label{eq:dirichlet_expectation}
    \mathbb{E}[X_m]
    =
    \frac{\beta_m}{\sum_{j=1}^M \beta_j}.
\end{equation}

\vspace{-1em}
\subsection{DP Mixture and Posterior over $G$}

In a DP mixture model, we draw a random measure $G$ and latent
component parameters $\theta_t$ for each sample:
\begin{align}
    G &\sim \mathrm{DP}(\alpha,G_0), \\
    \theta_t \mid G &\sim G, \\
    z_t \mid \theta_t &\sim p(z \mid \theta_t),
\end{align}
where $z_t$ is the observed feature and $\theta_t \in \Theta$ is the
parameter of the component (category) that generated $z_t$.

Suppose we have observed $t-1$ latent parameters
$\theta_{1:t-1} = (\theta_1,\dots,\theta_{t-1})$.
We are interested in the posterior over $G$ and the predictive
distribution for $\theta_t$.
To make the DP conjugacy explicit, consider an arbitrary finite
partition $(A_1,\dots,A_M)$ of $\Theta$.
Define the counts
\begin{equation}
    n_m
    =
    \sum_{i=1}^{t-1}
    \mathbf{1}\{\theta_i \in A_m\},
    \qquad
    N_{t-1} = \sum_{m=1}^M n_m = t-1.
\end{equation}
By the definition in \cref{eq:dp_def}, under the prior we have
\[
(G(A_1),\dots,G(A_M))
\sim
\mathrm{Dir}\bigl(
    \alpha G_0(A_1),\dots,\alpha G_0(A_M)
\bigr).
\]
Conditioning on $\theta_{1:t-1}$, the likelihood only depends on $G$
through the vector $(G(A_1),\dots,G(A_M))$.
Each $\theta_i$ that falls in $A_m$ contributes a factor $G(A_m)$ to
the likelihood, so the posterior over
$\bigl(G(A_1),\dots,G(A_M)\bigr)$ given $\theta_{1:t-1}$ is
\begin{equation}
\label{eq:dirichlet_posterior_partition}
\begin{aligned}
    &\bigl(G(A_1),\dots,G(A_M)\bigr)
    \mid \theta_{1:t-1} \\
    &\sim
    \mathrm{Dir}\bigl(
        \alpha G_0(A_1) + n_1,\dots,
        \alpha G_0(A_M) + n_M
    \bigr).
\end{aligned}
\end{equation}
by Dirichlet--multinomial conjugacy.

Comparing \cref{eq:dirichlet_posterior_partition} with the defining
property in \cref{eq:dp_def}, we see that the posterior over $G$ is still
a Dirichlet process:
\begin{equation}
\label{eq:dp_posterior}
    G \mid \theta_{1:t-1}
    \sim
    \mathrm{DP}\!\left(
        \alpha + N_{t-1},\;
        G_0'
    \right),
\end{equation}
where the updated base measure $G_0'$ satisfies
\begin{equation}
\label{eq:dp_posterior_base}
    G_0'(A_m)
    =
    \frac{\alpha G_0(A_m) + n_m}{\alpha + N_{t-1}}
    \quad\text{for all }m.
\end{equation}
\cref{eq:dp_posterior_base} extends uniquely from the
partition to all measurable sets $A$, and can be written compactly as
\begin{equation}
\label{eq:dp_posterior_base_compact}
    G_0'
    =
    \frac{\alpha}{\alpha + N_{t-1}} G_0
    +
    \frac{1}{\alpha + N_{t-1}}
    \sum_{i=1}^{t-1} \delta_{\theta_i},
\end{equation}
where $\delta_{\theta}$ is a point mass at $\theta$.

\vspace{-1em}
\subsection{Predictive Distribution for $\theta_t$}

We now derive the predictive distribution for the next component
parameter $\theta_t$ given $\theta_{1:t-1}$.
By the generative model,
\begin{equation}
    p(\theta_t \in A \mid \theta_{1:t-1})
    =
    \int G(A)\, p(G \mid \theta_{1:t-1})\,\mathrm{d}G.
\end{equation}
The right-hand side is the posterior expectation of $G(A)$, which we
can compute using the finite-dimensional Dirichlet representation.
For the partition $(A, A^c)$, the prior is
\[
\bigl(G(A), G(A^c)\bigr)
\sim
\mathrm{Dir}\bigl(\alpha G_0(A),\ \alpha G_0(A^c)\bigr),
\]
and the posterior over $(G(A), G(A^c))$ is
\begin{equation}
\begin{aligned}
&\bigl(G(A), G(A^c)\bigr)
\mid \theta_{1:t-1} \\
&\sim 
\mathrm{Dir}\bigl(
    \alpha G_0(A) + n_A,\;
    \alpha G_0(A^c) + N_{t-1} - n_A
\bigr).
\end{aligned}
\end{equation}
where $n_A = \sum_{i=1}^{t-1} \mathbf{1}\{\theta_i \in A\}$.
By the expectation formula \cref{eq:dirichlet_expectation},
\begin{equation}
\label{eq:predictive_theta_set}
    p(\theta_t \in A \mid \theta_{1:t-1})
    =
    \mathbb{E}[G(A) \mid \theta_{1:t-1}]
    =
    \frac{\alpha G_0(A) + n_A}{\alpha + N_{t-1}}.
\end{equation}
This is the Dirichlet--process predictive rule:
the next draw $\theta_t$ lands in $A$ with probability proportional to
a weighted sum of the prior mass $\alpha G_0(A)$ and the number of
previous draws that fell in $A$.

\vspace{-1em}
\subsection{Predictive Rule for Category Indices}

To connect \cref{eq:predictive_theta_set} with the category index
$c_t$ used in the main text, let
$\{\phi_1,\dots,\phi_{K_{t-1}}\}$ be the $K_{t-1}$ distinct values
among $\{\theta_1,\dots,\theta_{t-1}\}$, corresponding to the existing
categories at time $t-1$.
Let $n_k$ be the number of times $\theta_i$ takes the value $\phi_k$:
\begin{equation}
    n_k
    =
    \sum_{i=1}^{t-1}
    \mathbf{1}\{\theta_i = \phi_k\},
    \qquad
    N_{t-1}
    =
    \sum_{k=1}^{K_{t-1}} n_k.
\end{equation}
Consider the measurable sets
$A_k = \{\phi_k\}$ for $k=1,\dots,K_{t-1}$.
For each such singleton, \cref{eq:predictive_theta_set} gives
\begin{equation}
\label{eq:predictive_existing_atom}
    p(\theta_t = \phi_k \mid \theta_{1:t-1})
    =
    \frac{\alpha G_0(\{\phi_k\}) + n_k}{\alpha + N_{t-1}}.
\end{equation}
In our DP mixture, the base measure $G_0$ is a continuous prior over
component parameters (it puts no point mass on any specific value), so
$G_0(\{\phi_k\}) = 0$ for every previously seen atom $\phi_k$.
In this case, \cref{eq:predictive_existing_atom} simplifies to
\begin{equation}
\label{eq:predictive_existing_continuous}
    p(\theta_t = \phi_k \mid \theta_{1:t-1})
    =
    \frac{n_k}{\alpha + N_{t-1}}.
\end{equation}

We now define the category index $c_t$ by
\[
c_t =
\begin{cases}
k, & \text{if } \theta_t = \phi_k \text{ for some }k,\\[2pt]
\text{new}, & \text{if } \theta_t \text{ is a new draw from } G_0.
\end{cases}
\]
The probability of assigning $z_t$ to an \emph{existing} category $k$
is exactly the probability that $\theta_t$ equals the corresponding
atom $\phi_k$:
\begin{equation}
\label{eq:dp_assign_prob}
    P(c_t = k \mid \theta_{1:t-1})
    =
    p(\theta_t = \phi_k \mid \theta_{1:t-1})
    =
    \frac{n_k}{\alpha + N_{t-1}}.
\end{equation}
The remaining probability mass corresponds to drawing a \emph{new}
atom from $G_0$; this is the ``birth'' event:
\vspace{-1em}
\begin{equation}
\label{eq:dp_birth_prob}
\begin{aligned}
    P(c_t = \text{new} \mid \theta_{1:t-1})
    &= 1 - \sum_{k=1}^{K_{t-1}} P(c_t = k \mid \theta_{1:t-1}) \\
    &= 1 - \sum_{k=1}^{K_{t-1}} \frac{n_k}{\alpha + N_{t-1}} \\
    &= \frac{\alpha}{\alpha + N_{t-1}},
\end{aligned}
\end{equation}
where we used $\sum_k n_k = N_{t-1}$.
Since the history $\mathcal{D}_{t-1}$ determines the category counts
$\{n_k\}$ (and hence the empirical partition of $\theta_{1:t-1}$), the
same predictive rule holds when conditioning on $\mathcal{D}_{t-1}$:
\begin{align}
    P(c_t = k \mid \mathcal{D}_{t-1})
    &= \frac{n_k}{\alpha + N_{t-1}}, \\[2pt]
    P(c_t = \text{new} \mid \mathcal{D}_{t-1})
    &= \frac{\alpha}{\alpha + N_{t-1}}.
\end{align}

\section{Estimation of NIW Hyperparameters}
\label{app:appendix_eb_niw}
% In this appendix, we derive the empirical--Bayes estimates of the
% Normal--Inverse--Wishart (NIW) hyperparameters
% $(\mu_0,\kappa_0,\Psi_0,\nu_0)$ used in \cref{sec:initialization}, corresponding to
% \cref{eq:estimate_nu0,eq:estimate_psi0,eq:kappa_trace_match} in the main text.
% The key idea is to treat the labeled known classes in $\mathcal{D}_S$
% as draws from the NIW--Gaussian hierarchy and to match theoretical
% moments of this hierarchy to empirical moments computed from
% $\mathcal{D}_S$.
In this appendix, we provide the moment-matching estimation details for the Normal--Inverse--Wishart (NIW) hyperparameters \((\mu_0,\kappa_0,\Psi_0,\nu_0)\) used in Sec.~\ref{sec:initialization}. We refer to this procedure as empirical-Bayes-inspired because it uses labeled support-set statistics to calibrate the global prior, while also making practical approximations for stability and simplicity. In particular, the NIW--Gaussian hierarchy below should be viewed as a working calibration model rather than a claim that all known and future novel categories are exactly generated from this hierarchy.

The main idea is to match empirical moments computed from the labeled known classes in \(\mathcal{D}_S\) to the corresponding moments implied by the working NIW--Gaussian model. This yields data-adaptive estimates for \(\mu_0\), \(\Psi_0\), and \(\kappa_0\), which provide a practical initialization for the DP\mbox{-}BOA birth prior and known-category posteriors.

\vspace{-1em}
\subsection{Setup and Empirical Statistics}

Let $\mathcal{Y}_S$ denote the set of $K_S$ known classes, and let
$\mathcal{D}_k \subset \mathcal{D}_S$ be the support-set features for
class $k$.
We define the per-class and pooled statistics (repeating the notation
from Sec.~\ref{sec:initialization} for completeness):
\vspace{-1em}
\begin{equation*}
\resizebox{1.0\linewidth}{!}{
\begin{minipage}{\linewidth}
\begin{align*}
    \quad&\bar{\mathbf{z}}_k = \frac{1}{n_k}\sum_{\mathbf{z} \in \mathcal{D}_k} \mathbf{z}  \quad \text{(class mean)} \\
    \quad&\mathbf{S}_k = \sum_{\mathbf{z} \in \mathcal{D}_k} (\mathbf{z} - \bar{\mathbf{z}}_k)(\mathbf{z} - \bar{\mathbf{z}}_k)^\top  \quad \text{(class scatter)} \\
    \quad&\bar{\mathbf{z}} = \frac{1}{M}\sum_{k=1}^{K_S} n_k \bar{\mathbf{z}}_k  \quad \text{(global mean)} \\
    \quad&\mathbf{\Sigma}_{\text{within}} = \frac{1}{M-K_S}\sum_{k=1}^{K_S} \mathbf{S}_k  \quad \text{(pooled within-class covariance)} \\
    \quad&\mathbf{\Sigma}_{\text{means}} = \frac{1}{K_S-1}\sum_{k=1}^{K_S} (\bar{\mathbf{z}}_k - \bar{\mathbf{z}})(\bar{\mathbf{z}}_k - \bar{\mathbf{z}})^\top  \quad \text{(covariance of class means)} \\
    \quad&\overline{n^{-1}} = \frac{1}{K_S}\sum_{k=1}^{K_S} \frac{1}{n_k} \quad \text{(average inverse class size)}.
\end{align*}
\end{minipage}
}
\end{equation*}
Here $M = |\mathcal{D}_S|$ is the total number of labeled support
samples and $d$ is the feature dimension.

\vspace{-1em}
% \subsection{NIW Hierarchy and Basic Moments}
\subsection{NIW Working Model and Basic Moments}
We use the following NIW--Gaussian hierarchy as a working model for prior calibration. For each known class \(k\), we associate a class-level Gaussian with parameters \((\mu_k,\Sigma_k)\), and place a shared NIW prior over these parameters:
\begin{align}
    (\mu_k,\Sigma_k) &\sim \mathrm{NIW}(\mu_0,\kappa_0,\Psi_0,\nu_0), \label{eq:eb_niw_prior}\\
    z \mid \mu_k,\Sigma_k &\sim \mathcal{N}(z \mid \mu_k,\Sigma_k).
\end{align}
The support features in \(\mathcal{D}_k\) are then treated as samples from this class-level Gaussian for the purpose of estimating the global prior statistics. This modeling view is used only to obtain a stable support-set calibration; during online inference, category posteriors are updated sequentially from the stream evidence.

% We recall the NIW prior and the Gaussian likelihood from
% Appendix \ref{app:appendix_niw}.
% For each class $k$, we assume
% \begin{align}
%     (\mu_k,\Sigma_k)
%     &\sim \mathrm{NIW}(\mu_0,\kappa_0,\Psi_0,\nu_0),
%     \label{eq:eb_niw_prior}\\
%     z \mid \mu_k,\Sigma_k
%     &\sim \mathcal{N}(z \mid \mu_k,\Sigma_k),
% \end{align}
% and the $n_k$ support features in $\mathcal{D}_k$ are i.i.d. samples
% from this Gaussian.

The NIW hierarchy can be written in the standard conditional form
(Appendix \ref{app:appendix_niw}):
\begin{align}
    \Sigma_k &\sim \mathcal{W}^{-1}(\Psi_0,\nu_0), \\
    \mu_k \mid \Sigma_k
    &\sim \mathcal{N}\!\left(\mu_0,\tfrac{1}{\kappa_0}\Sigma_k\right).
\end{align}
From this we obtain the basic prior moments:
\begin{align}
    \mathbb{E}[\mu_k] &= \mu_0, \label{eq:eb_E_mu}\\
    \mathrm{Cov}(\mu_k)
    &= \mathbb{E}\!\bigl[\mathrm{Cov}(\mu_k \mid \Sigma_k)\bigr]
     + \mathrm{Cov}\!\bigl(\mathbb{E}[\mu_k \mid \Sigma_k]\bigr) \\
    &= \mathbb{E}\!\left[\tfrac{1}{\kappa_0}\Sigma_k\right] + 0
    = \frac{1}{\kappa_0}\,\mathbb{E}[\Sigma_k].
    \label{eq:eb_cov_mu}
\end{align}
For the Inverse--Wishart part, a standard property (see, e.g.,
\cite{murphy2012machine,bishop2006pattern}) is that if
$\Sigma \sim \mathcal{W}^{-1}(\Psi_0,\nu_0)$ with $\nu_0 > d + 1$,
then
\begin{equation}
\label{eq:eb_E_Sigma}
    \mathbb{E}[\Sigma]
    =
    \frac{\Psi_0}{\nu_0 - d - 1}.
\end{equation}
We will use \cref{eq:eb_E_mu,eq:eb_cov_mu,eq:eb_E_Sigma} to construct
empirical--Bayes estimates.

\vspace{-1em}
\subsection{Prior Mean \texorpdfstring{$\mu_0$}{mu0}}
We justify the choice of $\mu_0$ as the empirical global mean
$\bar{\mathbf{z}}$ by computing the unconditional mean of a randomly drawn
support feature under the NIW--Gaussian hierarchy.

Consider drawing a labeled feature $z$ according to the generative
process in Appendix \ref{app:appendix_niw} and \cref{eq:eb_niw_prior}.
For each sample we can write
\[
    z \sim p(z)
    =
    \sum_{k \in \mathcal{Y}_S}
    P(y{=}k)\,
    p(z \mid y{=}k),
\]
where
$(\mu_k,\Sigma_k) \sim \mathrm{NIW}(\mu_0,\kappa_0,\Psi_0,\nu_0)$
and $z \mid y{=}k,\mu_k,\Sigma_k \sim \mathcal{N}(z \mid \mu_k,\Sigma_k)$.
Using the law of total expectation,
\begin{align}
    \mathbb{E}[z]
    &= \mathbb{E}_y
       \mathbb{E}_{\mu_k,\Sigma_k}
       \bigl[
           \mathbb{E}[z \mid y{=}k,\mu_k,\Sigma_k]
       \bigr] \\
    &= \mathbb{E}_y
       \mathbb{E}_{\mu_k,\Sigma_k}[\mu_k]
     = \mathbb{E}_y[\mu_0]
     = \mu_0,
\end{align}
where we used $\mathbb{E}[\mu_k]=\mu_0$ from \cref{eq:eb_E_mu}.
Thus $\mu_0$ is the population mean of the marginal feature
distribution obtained by integrating out both the class label and the
class parameters.

Given the labeled support set
$\mathcal{D}_S = \{z_i\}_{i=1}^M$, the empirical mean
\begin{equation}
    \bar{\mathbf{z}}
    =
    \frac{1}{M}\sum_{i=1}^M z_i
    =
    \frac{1}{M}
    \sum_{k=1}^{K_S} n_k \bar{\mathbf{z}}_k
\end{equation}
is a consistent, approximately unbiased estimator of
\(\mathbb{E}[z] = \mu_0\).
In the empirical--Bayes spirit, we therefore set
\begin{equation}
\label{eq:eb_mu0}
    \mu_0 = \bar{\mathbf{z}},
\end{equation}
so that the global mean implied by the NIW prior matches the empirical
mean of the labeled features.

\vspace{-1em}
\subsection{Covariance Scale \texorpdfstring{$\Psi_0$}{Psi0}}
We now derive the initialization of the covariance scale $\Psi_0$ under
the NIW hierarchy.

\paragraph{Prior mean of class covariances.}
From Appendix \ref{app:appendix_niw}, the covariance of class $k$ obeys
\begin{equation}
    \Sigma_k \sim \mathcal{W}^{-1}(\Psi_0,\nu_0),
\end{equation}
and \cref{eq:eb_E_Sigma} gives its prior mean
\begin{equation}
\label{eq:eb_E_Sigma_iid}
    \mathbb{E}[\Sigma_k]
    =
    \frac{\Psi_0}{\nu_0 - d - 1}.
\end{equation}
Since all classes share the same NIW hyperparameters,
$\mathbb{E}[\Sigma_k]$ is identical across $k$.

\paragraph{Pooled within-class covariance as an estimator of $\mathbb{E}[\Sigma_k]$.}
For class $k$ with $n_k$ support features
$\{z_i\}_{i=1}^{n_k} \subset \mathcal{D}_k$ drawn i.i.d. from
$\mathcal{N}(\mu_k,\Sigma_k)$, define the sample covariance as
\begin{align}
    C_k
    &= \frac{\mathbf{S}_k}{n_k - 1}.
\end{align}
Standard Gaussian results imply
\begin{equation}
\label{eq:eb_unbiased_Ck}
    \mathbb{E}[C_k \mid \mu_k,\Sigma_k]
    =
    \Sigma_k.
\end{equation}
Taking expectation over the NIW prior gives
\begin{equation}
\label{eq:eb_unbiased_Ck_marginal}
    \mathbb{E}[C_k]
    =
    \mathbb{E}[\Sigma_k].
\end{equation}

The pooled within-class covariance is
\begin{equation}
    \mathbf{\Sigma}_{\mathrm{within}}
    =
    \frac{1}{M - K_S}
    \sum_{k=1}^{K_S} \mathbf{S}_k
    =
    \sum_{k=1}^{K_S}
    w_k\,C_k,
\end{equation}
where
\(
    w_k = \frac{n_k - 1}{M - K_S}
\)
and $\sum_{k} w_k = 1$.
Using linearity of expectation and \cref{eq:eb_unbiased_Ck_marginal},
\begin{align}
\label{eq:eb_E_within_equals_E_Sigma}
    \mathbb{E}[\mathbf{\Sigma}_{\mathrm{within}}]
    &=
    \mathbb{E}\Bigl[
        \sum_{k=1}^{K_S} w_k\,C_k
    \Bigr]
    =
    \sum_{k=1}^{K_S} w_k\,\mathbb{E}[C_k] \nonumber\\
    &=
    \sum_{k=1}^{K_S} w_k\,\mathbb{E}[\Sigma_k]
    =
    \mathbb{E}[\Sigma_k].
\end{align}
Thus $\mathbf{\Sigma}_{\mathrm{within}}$ is an unbiased, sample-size–weighted
estimator of the prior mean $\mathbb{E}[\Sigma_k]$.

\paragraph{Moment matching for \texorpdfstring{$\Psi_0$}{Psi0}.}
In the empirical--Bayes framework, we match the NIW prior mean of the
class covariances to the pooled within-class covariance:
\begin{equation}
    \label{eq:eb_sigma_match}
    \mathbb{E}[\Sigma_k]
    \;\approx\;
    \mathbf{\Sigma}_{\mathrm{within}}.
\end{equation}
Combining \cref{eq:eb_E_Sigma_iid,eq:eb_E_within_equals_E_Sigma} yields
\begin{equation}
    \mathbf{\Sigma}_{\mathrm{within}}
    \approx
    \frac{\Psi_0}{\nu_0 - d - 1}
    \quad\Longrightarrow\quad
    \Psi_0
    =
    (\nu_0 - d - 1)\,\mathbf{\Sigma}_{\mathrm{within}}.
\label{eq:eb_Psi0}
\end{equation}
This gives the covariance-scale initialization used in the main text.

\vspace{-1em}
\subsection{Mean Strength \texorpdfstring{$\kappa_0$}{kappa0}}
We now derive the empirical--Bayes estimate for the mean-strength
parameter $\kappa_0$ by matching the covariance of class means.

\paragraph{Class-mean covariance under the NIW hierarchy.}
Let $\bar{Z}_k$ denote the class mean corresponding to class
$k$, obtained from $n_k$ i.i.d.\ samples
$z_i \sim \mathcal{N}(\mu_k,\Sigma_k)$ under the NIW prior
\cref{eq:eb_niw_prior}.
Conditionally on $(\mu_k,\Sigma_k)$,
\[
    \bar{Z}_k \mid \mu_k,\Sigma_k
    \sim
    \mathcal{N}\!\left(
        \mu_k,\;
        \frac{1}{n_k}\Sigma_k
    \right),
\]
so
\[
    \mathrm{Cov}(\bar{Z}_k \mid \mu_k,\Sigma_k)
    =
    \frac{1}{n_k}\Sigma_k,
    \qquad
    \mathbb{E}[\bar{Z}_k \mid \mu_k,\Sigma_k]
    =
    \mu_k.
\]
Applying the law of total covariance,
\begin{equation}
\label{eq:eb_cov_Zbar_exact_short}
\begin{aligned}
    \mathrm{Cov}(\bar{Z}_k)
    &= \mathbb{E}\!\bigl[\mathrm{Cov}(\bar{Z}_k \mid \mu_k,\Sigma_k)\bigr]
     + \mathrm{Cov}\!\bigl(\mathbb{E}[\bar{Z}_k \mid \mu_k,\Sigma_k]\bigr) \\
    &= \frac{1}{n_k}\,\mathbb{E}[\Sigma_k]
     + \mathrm{Cov}(\mu_k). 
\end{aligned}
\end{equation}
Using \cref{eq:eb_cov_mu}, this becomes
\begin{equation}
\label{eq:eb_cov_Zbar_final}
    \mathrm{Cov}(\bar{Z}_k)
    =
    \left(
        \frac{1}{n_k}
        +
        \frac{1}{\kappa_0}
    \right)
    \mathbb{E}[\Sigma_k].
\end{equation}

\paragraph{Matching to empirical class-mean covariance.}
On the empirical side, the observed class means
$\{\bar{\mathbf{z}}_k\}_{k=1}^{K_S}$ with global mean
$\bar{\mathbf{z}}$ define the sample covariance of class means
$\mathbf{\Sigma}_{\mathrm{means}}$ introduced in the setup above.
This is a natural estimator of the model-side covariance
$\mathrm{Cov}(\bar{Z}_k)$ in \cref{eq:eb_cov_Zbar_final}, i.e.,
\[
    \mathrm{Cov}(\bar{Z}_k)
    \;\approx\;
    \mathbf{\Sigma}_{\mathrm{means}}.
\]
By \cref{eq:eb_sigma_match}, the pooled within-class
covariance $\mathbf{\Sigma}_{\mathrm{within}}$ provides an empirical
estimate of the prior mean of the class covariances,
\[
    \mathbb{E}[\Sigma_k]
    \;\approx\;
    \mathbf{\Sigma}_{\mathrm{within}}.
\]
Substituting these plug-in estimates into
\cref{eq:eb_cov_Zbar_final} and averaging over classes (so that
$\mathbb{E}[1/n_k]$ is approximated by
$\overline{n^{-1}}$) yields the matrix-valued moment-matching relation
\begin{equation}
\label{eq:eb_Sigma_means_matrix_short}
    \mathbf{\Sigma}_{\mathrm{means}}
    \;\approx\;
    \left(
        \overline{n^{-1}}
        +
        \frac{1}{\kappa_0}
    \right)
    \mathbf{\Sigma}_{\mathrm{within}},
\end{equation}
which we use to estimate $\kappa_0$.

\paragraph{Trace-based scalar approximation.}
The matrix relation in Eq.~\eqref{eq:eb_Sigma_means_matrix_short} cannot in general be matched exactly with a single scalar \(\kappa_0\), especially when the empirical covariance of class means and the pooled within-class covariance have different eigenspaces. We therefore use a trace-based moment-matching approximation, which matches the average variance across dimensions:
\begin{equation}
    \mathrm{tr}(\mathbf{\Sigma}_{\mathrm{means}})
    \;\approx\;
    \left(
        \overline{n^{-1}} + \frac{1}{\kappa_0}
    \right)
    \mathrm{tr}(\mathbf{\Sigma}_{\mathrm{within}}).
\end{equation}
Solving for \(1/\kappa_0\) gives
\begin{equation}
    \frac{1}{\kappa_0}
    \;\approx\;
    \frac{\mathrm{tr}(\mathbf{\Sigma}_{\mathrm{means}})}
         {\mathrm{tr}(\mathbf{\Sigma}_{\mathrm{within}})}
    -
    \overline{n^{-1}},
    \label{eq:eb_kappa0_scalar}
\end{equation}
and hence
\begin{equation}
    \kappa_0
    \;\approx\;
    \left(
    \frac{\mathrm{tr}(\mathbf{\Sigma}_{\mathrm{means}})}
         {\mathrm{tr}(\mathbf{\Sigma}_{\mathrm{within}})}
    -
    \overline{n^{-1}}
    \right)^{-1}.
\end{equation}
In practice, we use this trace-based approximation as a stable scalar calibration of the NIW mean strength, while retaining the full covariance structure in \(\mathbf{\Sigma}_{\mathrm{within}}\) and the subsequent NIW posterior updates.

% \paragraph{Trace-based scalar approximation.}
% To obtain a scalar equation for $\kappa_0$, we match average variance
% across dimensions by taking traces on both sides of
% \cref{eq:eb_Sigma_means_matrix_short}:
% \begin{equation}
%     \mathrm{tr}(\mathbf{\Sigma}_{\mathrm{means}})
%     \;\approx\;
%     \left(
%         \overline{n^{-1}}
%         +
%         \frac{1}{\kappa_0}
%     \right)
%     \mathrm{tr}(\mathbf{\Sigma}_{\mathrm{within}}).
% \end{equation}
% Solving for $1/\kappa_0$ gives
% \begin{equation}
% \label{eq:eb_kappa0_scalar}
%     \frac{1}{\kappa_0}
%     \;\approx\;
%     \frac{\mathrm{tr}(\mathbf{\Sigma}_{\mathrm{means}})}
%          {\mathrm{tr}(\mathbf{\Sigma}_{\mathrm{within}})}
%     - \overline{n^{-1}},
% \end{equation}
% and hence
% \begin{equation}
%     \kappa_0
%     \;\approx\;
%     \left(
%         \frac{\mathrm{tr}(\mathbf{\Sigma}_{\mathrm{means}})}
%              {\mathrm{tr}(\mathbf{\Sigma}_{\mathrm{within}})}
%         - \overline{n^{-1}}
%     \right)^{-1}.
% \end{equation}
% In practice, we use this trace-based approximation to obtain a stable,
% data-adaptive value of $\kappa_0$ while keeping the full covariance
% structure of $\mathbf{\Sigma}_{\mathrm{within}}$ and
% $\mathbf{\Sigma}_{\mathrm{means}}$.

\section{Online Update of Sufficient Statistics}
\label{app:appendix_online_stats}

In this part, following \cite{welford1962note}, we derive the online update rules for the per-category
sufficient statistics $(n_k,\bar{\mathbf{z}}_k,\mathbf{S}_k)$ used by
DP-BOA. These rules allow us to update the statistics of the assigned
category for each incoming feature $\mathbf{z}_t$ in constant memory,
without storing past samples.

\vspace{-1em}
\subsection{Sufficient Statistics and Invariants}

For a given category $k$, let $\mathcal{D}_k$ denote the current set of
features assigned to this category, and let $n_k = |\mathcal{D}_k|$ be
its cardinality.
We maintain the following sufficient statistics (\cref{sec:online_inference} and
Appendix~\ref{app:appendix_eb_niw}):
\begin{align}
    \bar{\mathbf{z}}_k
    &= \frac{1}{n_k} \sum_{\mathbf{z} \in \mathcal{D}_k} \mathbf{z},
    \label{eq:online_mean_def} \\
    \mathbf{S}_k
    &= \sum_{\mathbf{z} \in \mathcal{D}_k}
       (\mathbf{z} - \bar{\mathbf{z}}_k)
       (\mathbf{z} - \bar{\mathbf{z}}_k)^\top.
    \label{eq:online_scatter_def}
\end{align}
Here $n_k$ is the category size, $\bar{\mathbf{z}}_k$ is the sample
mean, and $\mathbf{S}_k$ is the within-category scatter matrix.
Note that \cref{eq:online_scatter_def} is defined with respect to the
\emph{current} mean $\bar{\mathbf{z}}_k$, so both
$\bar{\mathbf{z}}_k$ and $\mathbf{S}_k$ must be updated when a new
sample is added.

In what follows we drop the index $k$ and the time superscript for
readability, and write $(n,\bar{\mathbf{z}},\mathbf{S})$ for the
pre-update statistics of the category receiving $\mathbf{z}_t$.

\vspace{-1em}
\subsection{Online Update of the Mean}
Suppose the current category has statistics $(n,\bar{\mathbf{z}})$, and
a new feature $\mathbf{z}_t$ is assigned to this category.
The updated count is
\begin{equation}
    n^{+} = n + 1.
\end{equation}
By the definition of the sample mean, the updated mean is
\begin{equation}
\label{eq:online_mean_update_explicit}
    \bar{\mathbf{z}}^{+}
    =
    \frac{1}{n^{+}}
    \left(
        \sum_{\mathbf{z} \in \mathcal{D}_k} \mathbf{z}
        + \mathbf{z}_t
    \right)
    =
    \frac{n \bar{\mathbf{z}} + \mathbf{z}_t}{n^{+}}.
\end{equation}
It is convenient to write this in incremental form.
Define the deviation
\begin{equation}
    \boldsymbol{\delta}
    =
    \mathbf{z}_t - \bar{\mathbf{z}},
\end{equation}
then \cref{eq:online_mean_update_explicit} becomes
\begin{equation}
\label{eq:online_mean_update_incremental}
    \bar{\mathbf{z}}^{+}
    =
    \bar{\mathbf{z}}
    +
    \frac{1}{n^{+}} \boldsymbol{\delta}.
\end{equation}

\vspace{-1em}
\subsection{Online Update of the Scatter}
We now derive the corresponding online update for the scatter matrix
$\mathbf{S}$ in \cref{eq:online_scatter_def}.
Let $\mathcal{D}_k^{+} = \mathcal{D}_k \cup \{\mathbf{z}_t\}$ denote
the updated sample set, and let
$(n^{+},\bar{\mathbf{z}}^{+},\mathbf{S}^{+})$ be the updated
statistics.
By definition,
\begin{equation}
\label{eq:online_S_plus_def}
    \mathbf{S}^{+}
    =
    \sum_{\mathbf{z} \in \mathcal{D}_k^{+}}
    (\mathbf{z} - \bar{\mathbf{z}}^{+})
    (\mathbf{z} - \bar{\mathbf{z}}^{+})^\top.
\end{equation}
We decompose the sum in \cref{eq:online_S_plus_def} into the
contribution from the old samples and the new sample:
\begin{equation}
\label{eq:online_S_plus_split}
\begin{aligned}
    \mathbf{S}^{+}
    &=
    \sum_{\mathbf{z} \in \mathcal{D}_k}
    (\mathbf{z} - \bar{\mathbf{z}}^{+})
    (\mathbf{z} - \bar{\mathbf{z}}^{+})^\top \\
    &\quad+
    (\mathbf{z}_t - \bar{\mathbf{z}}^{+})
    (\mathbf{z}_t - \bar{\mathbf{z}}^{+})^\top.
\end{aligned}
\end{equation}
For the first term, write
\(
\mathbf{z} - \bar{\mathbf{z}}^{+}
 = (\mathbf{z} - \bar{\mathbf{z}}) + (\bar{\mathbf{z}} - \bar{\mathbf{z}}^{+})
\).
Using $\sum_{\mathbf{z} \in \mathcal{D}_k} (\mathbf{z} - \bar{\mathbf{z}}) = \mathbf{0}$,
we obtain
\begin{equation}
\label{eq:online_S_old_part}
\begin{aligned}
    &\sum_{\mathbf{z} \in \mathcal{D}_k}
    (\mathbf{z} - \bar{\mathbf{z}}^{+})
    (\mathbf{z} - \bar{\mathbf{z}}^{+})^\top \\
    &=
    \sum_{\mathbf{z} \in \mathcal{D}_k}
    \Bigl[
        (\mathbf{z} - \bar{\mathbf{z}})
        (\mathbf{z} - \bar{\mathbf{z}})^\top
        +
        (\bar{\mathbf{z}} - \bar{\mathbf{z}}^{+})
        (\bar{\mathbf{z}} - \bar{\mathbf{z}}^{+})^\top
    \Bigr] \\
    &=
    \mathbf{S}
    +
    n\,
    (\bar{\mathbf{z}} - \bar{\mathbf{z}}^{+})
    (\bar{\mathbf{z}} - \bar{\mathbf{z}}^{+})^\top.
\end{aligned}
\end{equation}
Substituting \cref{eq:online_S_old_part} into
\cref{eq:online_S_plus_split} gives
\begin{equation}
\label{eq:online_S_plus_intermediate}
\begin{aligned}
    \mathbf{S}^{+}
    &=
    \mathbf{S}
    +
    n\,
    (\bar{\mathbf{z}} - \bar{\mathbf{z}}^{+})
    (\bar{\mathbf{z}} - \bar{\mathbf{z}}^{+})^\top \\
    &\quad+
    (\mathbf{z}_t - \bar{\mathbf{z}}^{+})
    (\mathbf{z}_t - \bar{\mathbf{z}}^{+})^\top.
\end{aligned}
\end{equation}

Next we express all terms in \cref{eq:online_S_plus_intermediate} in
terms of the deviation $\boldsymbol{\delta} = \mathbf{z}_t - \bar{\mathbf{z}}$
and the updated mean \cref{eq:online_mean_update_incremental}.
From \cref{eq:online_mean_update_incremental} we have
\begin{align}
    \bar{\mathbf{z}}^{+} - \bar{\mathbf{z}}
    &= \frac{1}{n^{+}}\boldsymbol{\delta}, \\
    \bar{\mathbf{z}} - \bar{\mathbf{z}}^{+}
    &= -\frac{1}{n^{+}}\boldsymbol{\delta}, \\
    \mathbf{z}_t - \bar{\mathbf{z}}^{+}
    &= \boldsymbol{\delta}
       - (\bar{\mathbf{z}}^{+} - \bar{\mathbf{z}})
     = \boldsymbol{\delta}
       - \frac{1}{n^{+}}\boldsymbol{\delta}
     = \frac{n}{n^{+}}\boldsymbol{\delta}.
\end{align}
Substituting into \cref{eq:online_S_plus_intermediate} yields
\begin{equation}
\begin{aligned}
    \mathbf{S}^{+}
    &=
    \mathbf{S}
    +
    n
    \left(
        -\frac{1}{n^{+}}\boldsymbol{\delta}
    \right)
    \left(
        -\frac{1}{n^{+}}\boldsymbol{\delta}
    \right)^\top \\
    &\quad+
    \left(
        \frac{n}{n^{+}}\boldsymbol{\delta}
    \right)
    \left(
        \frac{n}{n^{+}}\boldsymbol{\delta}
    \right)^\top \\
    &=
    \mathbf{S}
    +
    \frac{n}{(n^{+})^2}\boldsymbol{\delta}\boldsymbol{\delta}^\top
    +
    \frac{n^2}{(n^{+})^2}\boldsymbol{\delta}\boldsymbol{\delta}^\top \\
    &=
    \mathbf{S}
    +
    \frac{n(n+1)}{(n+1)^2}\boldsymbol{\delta}\boldsymbol{\delta}^\top \\
    &=
    \mathbf{S}
    +
    \frac{n}{n^{+}}\,
    \boldsymbol{\delta}\boldsymbol{\delta}^\top.
\end{aligned}
\label{eq:online_S_update_simplified}
\end{equation}
Therefore, the scatter update can be written as
\begin{equation}
\label{eq:online_S_update_final}
    \mathbf{S}^{+}
    =
    \mathbf{S}
    +
    \frac{n}{n+1}
    (\mathbf{z}_t - \bar{\mathbf{z}})
    (\mathbf{z}_t - \bar{\mathbf{z}})^\top.
\end{equation}

In practice, a numerically stable and commonly used equivalent form is
\begin{equation}
\label{eq:online_S_update_welford}
    \mathbf{S}^{+}
    =
    \mathbf{S}
    +
    (\mathbf{z}_t - \bar{\mathbf{z}})
    (\mathbf{z}_t - \bar{\mathbf{z}}^{+})^\top,
\end{equation}
where $\bar{\mathbf{z}}^{+}$ is given by
\cref{eq:online_mean_update_incremental}.
\cref{eq:online_S_update_final,eq:online_S_update_welford} are
algebraically equivalent: substituting
$\mathbf{z}_t - \bar{\mathbf{z}}^{+}
 = \frac{n}{n^{+}}(\mathbf{z}_t - \bar{\mathbf{z}})$
into \cref{eq:online_S_update_welford} recovers
\cref{eq:online_S_update_final}.

\vspace{-1em}
\subsection{Initialization for a New Category}
When a new category is born at time $t$ (i.e., it has no previously
assigned samples), we initialize its statistics using the current
feature $\mathbf{z}_t$:
\begin{align}
    n^{+} &= 1, \\
    \bar{\mathbf{z}}^{+} &= \mathbf{z}_t, \\
    \mathbf{S}^{+} &= \mathbf{0}.
\end{align}
Subsequently, the category statistics
$(n,\bar{\mathbf{z}},\mathbf{S})$ are updated online using
\cref{eq:online_mean_update_incremental,eq:online_S_update_welford}.
Combined with the NIW posterior updates in
Appendix~\ref{app:appendix_niw}, these online updates ensure that
DP-BOA only needs to maintain $(n_k,\bar{\mathbf{z}}_k,\mathbf{S}_k)$
per category, without storing any past samples.

\section{Stronger Encoder Analysis}
\label{app:stronger_encoder}

The main results use our default encoder training protocol: a DINO-pretrained
ViT-B/16 backbone is fine-tuned on the labeled support set with standard
supervised cross-entropy (CE), and the encoder is then frozen during online inference.
To examine whether DP\mbox{-}BOA can benefit from stronger representations,
we additionally evaluate an encoder fine-tuned with a supervised contrastive
auxiliary loss (SupCon). This experiment changes only the offline encoder fine-tuning objective; the DP\mbox{-}BOA online head, empirical prior initialization, and
test-time update rules are kept unchanged. 

Specifically, we optimize
\begin{equation}
    \mathcal{L}_{\mathrm{enc}}
    =
    \mathcal{L}_{\mathrm{CE}}
    +
    \lambda \mathcal{L}_{\mathrm{supcon}},
    \qquad \lambda = 0.1,
\end{equation}
where \(\mathcal{L}_{\mathrm{CE}}\) is the standard cross-entropy loss on the
labeled support set. 
% We use the standard supervised contrastive loss
% \begin{equation}
%     \mathcal{L}_{\mathrm{supcon}}
%     =
%     \sum_{i \in \mathcal{I}}
%     \frac{-1}{|\mathcal{P}(i)|}
%     \sum_{p \in \mathcal{P}(i)}
%     \log
%     \frac{
%         \exp\!\left(\mathrm{sim}(\mathbf{h}_i,\mathbf{h}_p)/\tau\right)
%     }{
%         \sum_{a = 1, a \neq i}^{|B|}
%         \exp\!\left(\mathrm{sim}(\mathbf{h}_i,\mathbf{h}_a)/\tau\right)
%     },
%     \label{eq:supcon_loss}
% \end{equation}
% where \(\mathbf{h}_i\) denotes the normalized feature of sample \(i\),
% \(\mathrm{sim}(\cdot,\cdot)\) is cosine similarity, \(\mathcal{P}(i)\) is the set
% of samples in the mini-batch $B$ that share the same label as \(i\), and
% \(\tau\) is the temperature parameter.
We use the standard supervised contrastive loss
\begin{equation}
    \mathcal{L}_{\mathrm{supcon}}
    =
    \frac{1}{|\mathcal{B}|}
    \sum_{i \in \mathcal{B}}
    \frac{-1}{|\mathcal{P}(i)|}
    \sum_{p \in \mathcal{P}(i)}
    \log
    \frac{
        \exp\!\left(\mathrm{sim}(\mathbf{h}_i,\mathbf{h}_p)/\tau\right)
    }{
        \sum_{a \in \mathcal{B} \setminus \{i\}}
        \exp\!\left(\mathrm{sim}(\mathbf{h}_i,\mathbf{h}_a)/\tau\right)
    },
    \label{eq:supcon_loss}
\end{equation}
where $\mathbf{h}_i$ denotes the normalized feature of sample $i$,
$\mathrm{sim}(\cdot,\cdot)$ is cosine similarity,
$\mathcal{P}(i)=\{p \in \mathcal{B} \setminus \{i\}: y_p=y_i\}$ is the set of positive
samples in the mini-batch $\mathcal{B}$, and $\tau$ is the temperature parameter.

\cref{tab:stronger_encoder} compares the default CE encoder with the stronger
CE+SupCon encoder. DP\mbox{-}BOA consistently benefits from the stronger
representation, with especially clear gains on ImageNet100. This indicates that
better feature geometry is complementary to our probabilistic decision
framework: representation learning can improve the frozen feature space, while
DP\mbox{-}BOA remains responsible for online category birth and assignment. 

\begin{table}[t]
    \centering
    \caption{\textbf{Stronger encoder analysis.}
    CE denotes the default encoder trained with cross-entropy. CE+SupCon adds
    a supervised contrastive auxiliary loss with \(\lambda=0.1\).}
    \vspace{-0.5em}
    \label{tab:stronger_encoder}
    \small
    \setlength{\tabcolsep}{5pt}
    \resizebox{\linewidth}{!}{
    \begin{tabular}{lccccccccc}
        \toprule
        & \multicolumn{3}{c}{CUB}
        & \multicolumn{3}{c}{ImageNet100}
        & \multicolumn{3}{c}{Animalia} \\
        \cmidrule(lr){2-4}
        \cmidrule(lr){5-7}
        \cmidrule(lr){8-10}
        Loss
        & All & Known & Novel
        & All & Known & Novel
        & All & Known & Novel \\
        \midrule
        CE
        & 53.4 & 57.2 & 51.6
        & 33.8 & 75.8 & 12.7
        & 50.7 & 67.6 & 43.7 \\
        CE+SupCon
        & \textbf{55.8} & \textbf{62.4} & \textbf{52.5}
        & \textbf{47.1} & \textbf{84.7} & \textbf{28.2}
        & \textbf{51.5} & \textbf{68.6} & \textbf{44.5} \\
        \bottomrule
    \end{tabular}}
    \vspace{-1em}
\end{table}

\section{Category Decision Boundary Visualization}

\label{app:appendix_qualitative}
To qualitatively illustrate (i) the anisotropic geometry modeled by
DP\mbox{-}BOA and (ii) how its posterior–predictive high-density regions
(and hence the induced decision regions) adapt as evidence accumulates,
we visualize individual \emph{DP\mbox{-}BOA categories} in a fixed 2D
feature subspace.
We run OCD on the Oxford-IIIT Pet dataset with a frozen encoder.
Throughout this part, a ``category'' refers to a cluster indexed by
DP\mbox{-}BOA.
For known categories, each such cluster is initialized from a single
labeled class; for novel categories, the clusters are discovered online
from unlabeled query samples.

\begin{figure*}[t]
    \centering
    \includegraphics[width=\linewidth]{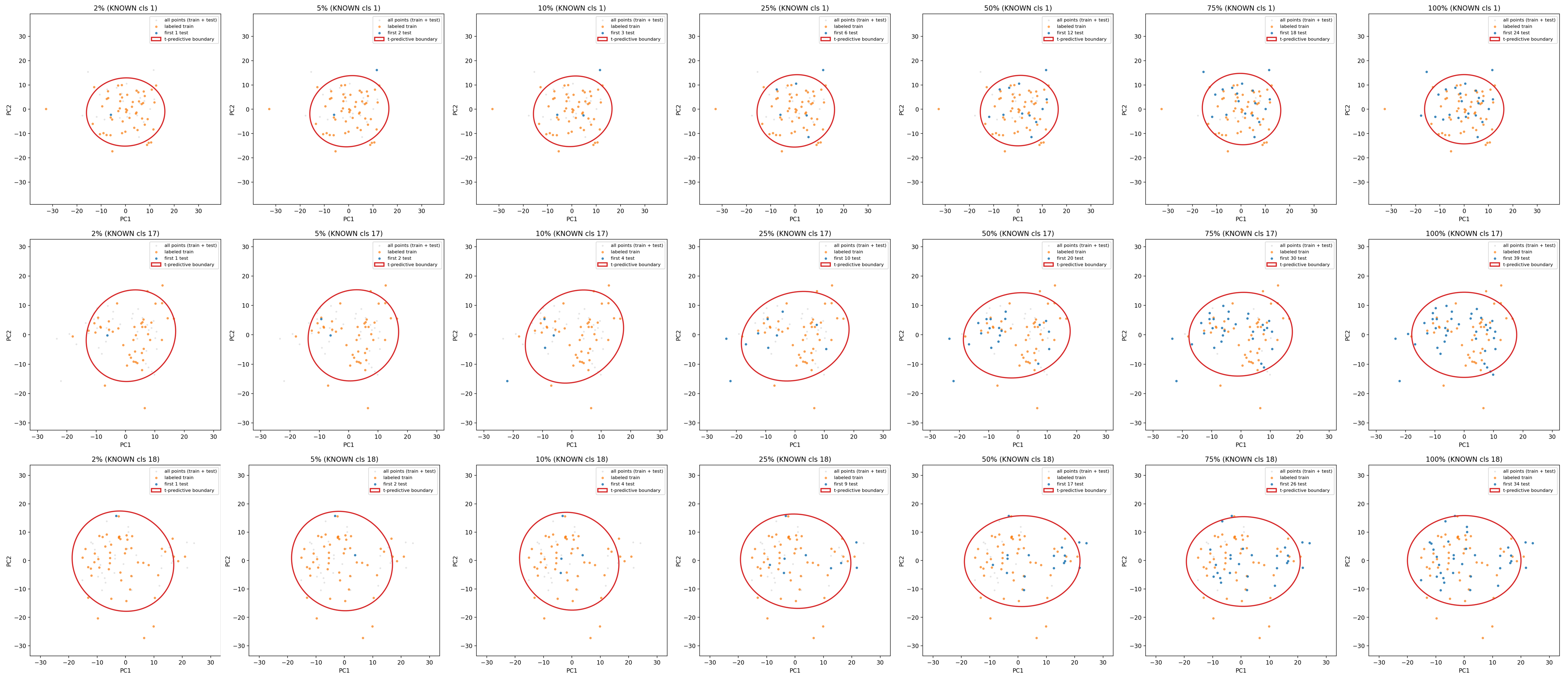}
    \vspace{-1.5em}
    \caption{Evidence-adaptive decision boundaries for three
    \emph{known} DP\mbox{-}BOA categories on the Oxford-IIIT Pet
    dataset.
    Each row corresponds to one category that is initialized from a
    labeled class, and columns show checkpoints where 2\%, 5\%,
    10\%, 25\%, 50\%, 75\%, and 100\% of the category’s assigned query
    samples have been observed (left to right).
    Light gray points indicate all PCA-projected features from the
    dataset, colored points highlight the labeled support and the query
    samples currently assigned to this category, and the red ellipse is
    a fixed Mahalanobis-radius level set of the 2D Student-$t$
    posterior–predictive distribution.
    The ellipses are anisotropic and largely stable over time,
    reflecting well-estimated covariances and high confidence for
    known categories.}
    \label{fig:qual_known}
    \vspace{-1em}
\end{figure*}

\begin{figure*}[t]
    \centering
    \includegraphics[width=\linewidth]{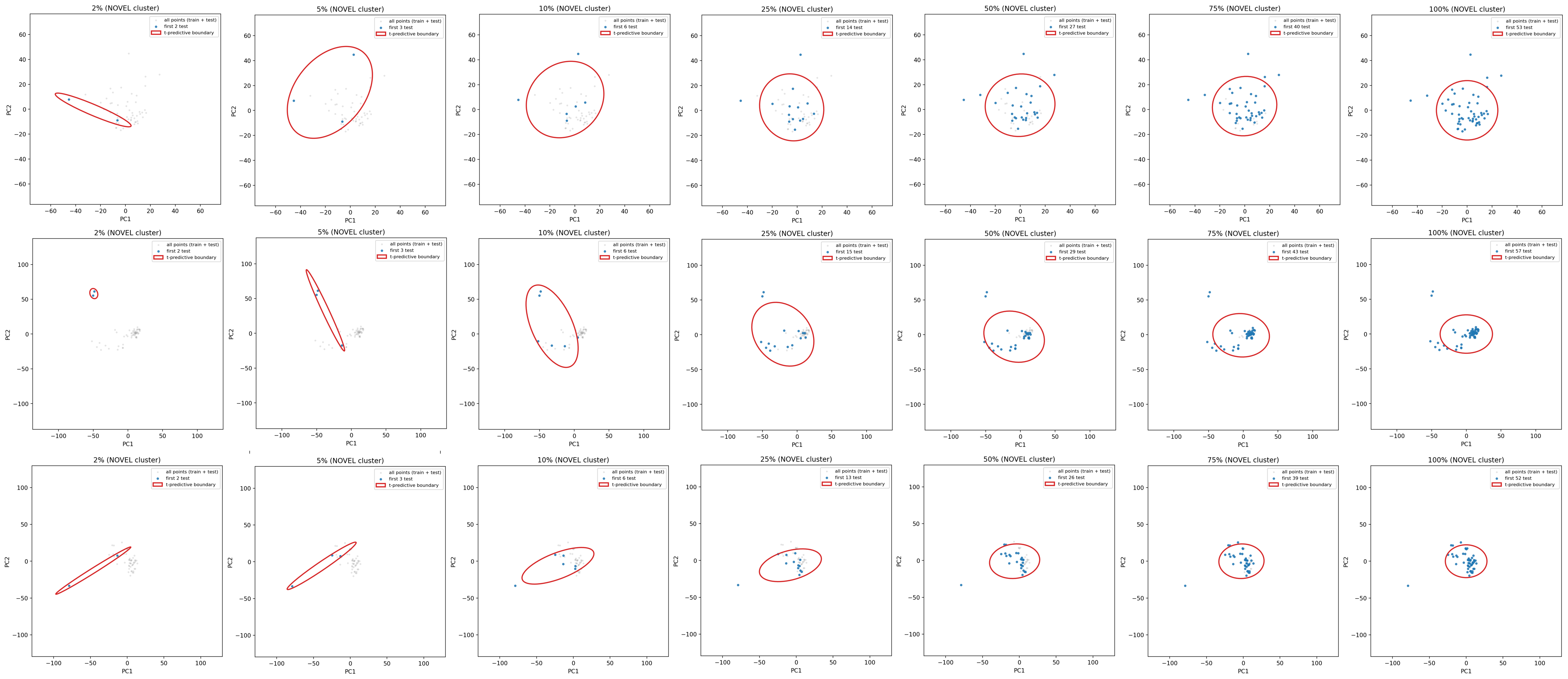}
    \vspace{-1.5em}
    \caption{Evidence-adaptive decision boundaries for three
    \emph{novel} DP\mbox{-}BOA categories on the Oxford-IIIT Pet
    dataset.
    These clusters contain only unlabeled query samples and have no
    ground-truth labels.
    Early contours are influenced by the global NIW prior and a small,
    noisy empirical scatter, so they can expand or rotate as $S_k$
    rapidly reshapes the posterior covariance $\Psi_k$.
    As more points are assigned, $S_k$ stabilizes, the NIW parameters
    $(\kappa_k,\nu_k)$ grow, and the Student-$t$ predictive contours
    become smaller and more stable, yielding tight, anisotropic
    decision regions around each discovered novel category.}
    \label{fig:qual_novel}
    \vspace{-1em}
\end{figure*}

For a chosen DP\mbox{-}BOA category $k$, we collect all features that
were ever assigned to it: for known categories this includes both
labeled support examples and their subsequent query assignments, while
for novel categories it consists only of the query samples assigned
after the category is born.
Let $\{\mathbf{x}_i\}_{i=1}^{N_k} \subset \mathbb{R}^d$ denote these
assigned features. We construct a \emph{category-specific} 2D projection by PCA on
$\{\mathbf{x}_i\}$.
Writing $\bar{\mathbf{x}}_k$ for the empirical mean and
$\mathbf{C}_k$ for the empirical covariance, we take the top two
eigenvectors $\mathbf{v}_{k,1},\mathbf{v}_{k,2} \in \mathbb{R}^d$ and
form
$
\mathbf{P}_k
=
[\mathbf{v}_{k,1},\mathbf{v}_{k,2}] \in \mathbb{R}^{d \times 2}.
$
Each feature is mapped to
$
\mathbf{z}_i = \mathbf{P}_k^\top (\mathbf{x}_i - \bar{\mathbf{x}}_k)
\in \mathbb{R}^2,
$
and we treat this 2D PCA subspace as the working space for
visualization.
For consistency with the full-dimensional NIW prior, we project the
pooled within-class covariance into this subspace to obtain a 2D NIW
prior, and then run the same online NIW/Student-$t$ updates as in the
main algorithm, now on the 2D points $\mathbf{z}_i$ (so the Student-$t$
formula in Appendix~\ref{app:appendix_student_t} is used with $d{=}2$).

% Along each category’s query stream we record several checkpoints
% (2\%, 5\%, 10\%, 25\%, 50\%, 75\%, and 100\% of the category’s assigned
% query samples).
% At each checkpoint we take the current NIW posterior
% $(\mu_k,\kappa_k,\Psi_k,\nu_k)$, convert it to the corresponding
% Student-$t$ predictive as in Appendix~\ref{app:appendix_student_t}, and
% plot an elliptical level set of constant Mahalanobis radius (a fixed
% ``$n_{\text{std}}\sigma$'' contour with respect to the predictive
% scale matrix).
% For a fixed category k, we keep the PCA projection $\mathbf{P}_k$ fixed
% across all checkpoints, so changes in the ellipse directly reflect the
% evolution of the posterior–predictive geometry rather than rescaling of
% the axes.
Along each category’s query stream, we record several representative checkpoints
at different stages of online evidence accumulation
(2\%, 5\%, 10\%, 25\%, 50\%, 75\%, and 100\%).
At each checkpoint, we take the current NIW posterior
$(\mu_k,\kappa_k,\Psi_k,\nu_k)$ estimated from the samples observed so far,
convert it to the corresponding Student-$t$ predictive distribution as in
Appendix~\ref{app:appendix_student_t}, and visualize an elliptical level set
with a constant Mahalanobis radius, namely a fixed
``$n_{\text{std}}\sigma$'' contour with respect to the predictive scale matrix.
This allows us to inspect how the posterior-predictive uncertainty
and category geometry evolve as more samples are assigned to the category.
For a fixed category $k$, we keep the PCA projection $\mathbf{P}_k$ fixed
across all checkpoints, so that changes in the ellipse directly reflect the
evolution of the posterior-predictive geometry, rather than being caused by
axis rescaling or changes in the visualization coordinate system.

\paragraph{Known categories.}
\cref{fig:qual_known} shows three \emph{known}
DP\mbox{-}BOA categories on Oxford-IIIT Pet, each initialized from a
distinct labeled class.
Each row corresponds to one such category, and columns show snapshots
at increasing fractions of assigned query samples (from 2\% to 100\%,
left to right).

The induced high-density regions are clearly anisotropic: different
known categories exhibit different orientations and aspect ratios that
are aligned with their empirical scatter in the PCA subspace.

In the 2D subspace, the predictive scale used to draw the ellipse has
the form
\vspace{-0.5em}
\begin{equation}
    \Sigma_{\text{pred},k}
    =
    \frac{\kappa_k + 1}{\kappa_k(\nu_k - d + 1)}\,\Psi_k,
\end{equation}
where $d{=}2$ here and $\Psi_k$ is the NIW scale matrix updated from
the scatter $S_k$ of the category.
For i.i.d.\ data with (finite) population covariance $\Sigma^\star$ in
this subspace, the scatter satisfies
$
    \mathbb{E}[S_k] = (n_k - 1)\,\Sigma^\star,
$
and by the law of large numbers $S_k / n_k \to \Sigma^\star$ as
$n_k \to \infty$.
Thus $S_k$ (and hence $\Psi_k$ through the NIW update) grows
approximately linearly with the number of assigned samples $n_k$ for
moderate to large $n_k$.
At the same time,
$
    \kappa_k = \kappa_0 + n_k,
    \quad
    \nu_k = \nu_0 + n_k,
$
so in the large-$n_k$ regime
\begin{equation}
    \frac{\kappa_k + 1}{\kappa_k(\nu_k - d + 1)}
    \;\approx\;
    \frac{1}{n_k}.
\end{equation}
As a result, the linear growth of $\Psi_k$ with $n_k$ and the roughly
$1/n_k$ decay of the scalar prefactor largely cancels, and
$\Sigma_{\text{pred},k}$ converges to a finite, anisotropic covariance
close to the underlying within-class covariance in this subspace.
For known classes, we already start with dozens of labeled examples,
so this large-sample regime is reached quickly: the empirical scatter
changes slowly, the eigenvectors and eigenvalue ratios of $\Psi_k$ are
stable, and the ellipses in \cref{fig:qual_known} preserve their
orientation and aspect ratio with only mild adjustments in overall size.

In addition, as $n_k$ grows, the degrees of freedom
$\nu_k - d + 1$ increase, so the Student-$t$ predictive becomes less
heavy-tailed and closer to a Gaussian.
Intuitively, this concentrates probability mass closer to the mean
$\mu_k$ and makes high-density regions slightly tighter.
Together, these effects produce the intended behavior: for
well-supported known categories, DP\mbox{-}BOA maintains a stable,
anisotropic posterior–predictive region (and thus a stable local
decision region) whose scale and tail behavior reflect high confidence
in the underlying category.
\vspace{-1.5em}
\paragraph{Novel categories.}
\cref{fig:qual_novel} shows three automatically discovered
\emph{novel} DP\mbox{-}BOA categories on Oxford-IIIT Pet, visualized
with the same protocol.
These clusters are formed purely from unlabeled query samples and do
not correspond to any ground-truth label.
When such a category is first born, its initial NIW posterior is
obtained by combining the global prior
$(\mu_0,\kappa_0,\Psi_0,\nu_0)$ with the first birth sample.
The prior covariance $\Psi_0$ encodes only an average ``typical''
scale across known classes; in the local 2D subspace of a specific
novel cluster, the true covariance can differ substantially.

In the very early snapshots, $n_k$ is small and the empirical scatter
$S_k$ is based on only a few points, so the NIW update
$
    \Psi_k
    =
    \Psi_0 + S_k + \text{(rank-one mean term)}
$
is still a compromise between the global prior $\Psi_0$ and a noisy
local estimate.
The resulting $\Psi_k$ may not yet be aligned with the principal
direction of the emerging cluster, and the degrees of freedom
$\nu_k - d + 1$ are still close to their prior value, so the
Student-$t$ predictive remains relatively heavy-tailed.
As more samples are assigned, $S_k$ quickly grows and dominates the
update of $\Psi_k$, causing the ellipse to expand or rotate to align
with the principal direction of the novel cluster (left to middle
columns of \cref{fig:qual_novel}).
Once a novel category has accumulated enough samples, its empirical
covariance $S_k / (n_k - 1)$ stabilizes and closely matches a local
population covariance $\Sigma^\star$ aligned with the discovered
cluster.
Beyond this point, the same compensation mechanism as for known
categories takes effect: $S_k$ grows roughly linearly with $n_k$, while
the scalar factor
$\frac{\kappa_k + 1}{\kappa_k(\nu_k - d + 1)}$ decays approximately
like $1/n_k$, so the predictive covariance $\Sigma_{\text{pred},k}$
approaches a stable anisotropic limit rather than shrinking
indefinitely.
Meanwhile, the increasing degrees of freedom
$\nu_k - d + 1$ make the Student-$t$ predictive less heavy-tailed,
further tightening the high-density region around $\mu_k$.
Accordingly, the later checkpoints (rightmost columns in
\cref{fig:qual_novel}) show ellipses that have settled into a
tight, stable shape around each novel cluster, with only minor further
changes as more evidence arrives.
Taken together, \cref{fig:qual_known} and \cref{fig:qual_novel}
demonstrate that DP\mbox{-}BOA (i) models category-specific anisotropic
geometry and (ii) adapts each posterior–predictive (and hence decision)
region from a prior-dominated shape to a data-driven, stable covariance
as evidence accumulates, both for known categories aligned with labeled classes and for purely discovered novel clusters.

%Finally, we want to emphasize that this compression is orthogonal to our main contribution---the principled Bayesian assign-vs-birth framework for OCD---serving as an optional engineering choice for deployment.

% \subsection{Cluster number growth}
% We report $K(t)$ over the stream in \cref{fig:cluster_number}. We report $K_{\text{final}}$ and compare it to the ground-truth class count and prior OCD methods in \cref{tab:kfinal_vs_gt}.

\section{Pseudocode for DP-BOA}
\label{app:appendix_pseudo}
We summarize DP-BOA as a single procedure containing an offline
initialization phase and an online birth-or-assign phase. Notation and
all closed-form NIW / Student-\texorpdfstring{$t$}{t} expressions follow
the main text.

\begin{algorithm}[H]
\caption{DP\mbox{-}BOA (Offline Initialization + Online Inference)}
\label{alg:dpboa}
\small   
\begin{algorithmic}[1]
  \State \textbf{Input:} support set $\mathcal{D}_S$, query stream $\mathcal{D}_Q$.
  \State \textbf{Output:} assignments $\{\hat{c}_t\}_{t=1}^N$

  \Statex
  \State \textbf{Offline: supervised encoder + NIW initialization}
  \State Train encoder $f_\theta$ and a linear classifier on $\mathcal{D}_S$ with cross-entropy
  \State Discard the classifier and freeze $f_\theta$
  \State Extract features $z_i=f_\theta(x_i)$ for $(x_i,y_i)\in\mathcal{D}_S$ and group by label
  \State For each class $k$, compute $n_k$, class mean $\bar{z}_k$, and scatter $S_k$
  \State From $\{(n_k,\bar{z}_k,S_k)\}$, estimate NIW prior $(\mu_0,\Psi_0,\kappa_0)$
  \Statex \hspace{\algorithmicindent} by the Empirical--Bayes procedure and set $\nu_0$ via the heuristic in \cref{eq:n0_rule}
  \State For each known class $k$, apply the NIW posterior update to obtain
  \Statex \hspace{\algorithmicindent} $\theta_k=(\mu_k,\Psi_k,\nu_k,\kappa_k)$ and initialize a cluster $C_k$

  \Statex
  \State \textbf{Online: DP\mbox{-}BOA birth-or-assign loop}
  \State Let $\mathcal{C}$ be the list of current clusters (initialized by known classes)
  \For{each query sample $x_t$ in $\mathcal{D}_Q$}
    \State Compute feature $z_t=f_\theta(x_t)$
    \State Compute ``new''-cluster Student-\(t\) predictive $t_d(z_t\mid\theta_0)$
    \Statex \hspace{\algorithmicindent} using \cref{eq:prior_pred}
    \State Birth log-score:
           $\ell_{\text{new}} \gets \log\alpha + \log t_d(z_t\mid\theta_0)$

    \For{each existing cluster $C_k \in \mathcal{C}$}
      \State From $\theta_k$, compute Student-\(t\) predictive $t_d(z_t\mid\theta_k)$
      \Statex \hspace{\algorithmicindent} using \cref{eq:post_pred}
      \State Assign log-score:
             $\ell_k \gets \log n_k + \log t_d(z_t\mid\theta_k)$
    \EndFor

    \State \textbf{Birth-or-assign decision:}
    \If{$\ell_{\text{new}} > \max_k \ell_k$}
      \State Spawn a new cluster $C_{\text{new}}$ with $n=1$, mean $z_t$, scatter $S=\mathbf{0}$
      \State Compute its NIW posterior $\theta_{\text{new}}$ from $\theta_0$
      \State Append $C_{\text{new}}$ to $\mathcal{C}$, set $\hat{c}_t \gets |\mathcal{C}|$
    \Else
      \State Let $k^\star=\arg\max_k \ell_k$ and set $\hat{c}_t \gets k^\star$
      \State Update $(n_{k^\star},\bar{z}_{k^\star},S_{k^\star})$ with a Welford step
      \State Recompute $\theta_{k^\star}$ from $(n_{k^\star},\bar{z}_{k^\star},S_{k^\star})$
    \EndIf
  \EndFor
\end{algorithmic}
\end{algorithm}

\section{Complexity Analysis}
\label{app:appendix_runtime}

In the main text, we analyzed the complexity of the DP-BOA head and showed that
its additional cost comes from full-covariance posterior-predictive scoring and online
Bayesian updates.
Let $d$ be the feature dimension and $K$ the current number of active categories.
As summarized in Appendix~\ref{app:appendix_pseudo}, each online step for an incoming
feature $\mathbf z_t$ consists of three parts:
(i) computing posterior-predictive Student-$t$ scores for all existing categories together
with the ``new'' hypothesis;
(ii) making the birth-or-assign decision by comparing these scores; and
(iii) updating the statistics $(n_k,\bar{\mathbf z}_k,\mathbf S_k)$ and the NIW
posterior of the selected category.

With full covariance matrices, the dominant cost in step~(i) is evaluating one quadratic
form per category, which gives $O(Kd^2)$ time.
In step~(iii), updating the count, mean, and scatter matrix via the online Welford-style
rules costs $O(d^2)$, while recomputing the predictive covariance factors for the updated
category costs $O(d^3)$ through a Cholesky factorization.
In practice, we cache the inverse and log-determinant of each category's predictive
covariance, so only the selected category needs to be refactorized after an update.
Therefore, the per-sample complexity of the DP-BOA head is
$O(Kd^2 + d^3)$ time and $O(Kd^2)$ memory, consistent with the statement in the main text. \\

\begin{table}[t]
\centering
\caption{\textbf{Runtime of DP-BOA.} Mean and max latency (ms / sample) over the full OCD stream. Numbers in parentheses denote the total number of categories (\#Cls).}
\vspace{-0.5em}
\label{tab:runtime}
\begingroup
\footnotesize
\setlength{\tabcolsep}{4pt}
\begin{adjustbox}{max width=0.9\linewidth, scale=0.9}
\begin{tabular}{lccc}
\toprule
& Animalia (77) & CUB (200) & OxfordPets (38) \\
\midrule
Mean / Max latency (ms) & 26.6 / 46.1 & 58.7 / 81.7 & 17.4 / 37.0 \\
\bottomrule
\end{tabular}
\end{adjustbox}
\endgroup
\vspace{-0em}
\end{table}
 
\begin{table}[t]
\centering
\footnotesize
\setlength{\tabcolsep}{3pt}
\caption{DP-BOA memory (MB). ``E2E'' denotes end-to-end.}
\label{tab:mem_dpboa}
\begin{adjustbox}{max width=1.0\linewidth, scale=1.0}
\begin{tabular}{l ccc ccc ccc}
\toprule
& \multicolumn{3}{c}{Animalia} 
& \multicolumn{3}{c}{CUB} 
& \multicolumn{3}{c}{StanfordCars} \\
\cmidrule(lr){2-4}\cmidrule(lr){5-7}\cmidrule(lr){8-10}
Method & Head & E2E & \#Clusters & Head & E2E & \#Clusters & Head & E2E & \#Clusters\\
\midrule
DP-BOA & 299.2 & 779.8 & 81 & 619.2 & 1108.8 & 175 & 626.7 & 1120.4 & 177\\
\bottomrule
\end{tabular}%
\end{adjustbox}
\end{table}

% \subsection{Memory Footprint}

% We also report both head-only and end-to-end GPU memory in \cref{tab:mem_dpboa}.
% As expected from the $O(Kd^2)$ storage of full-covariance statistics, the head memory grows approximately linearly with the number of active categories:
% 299.2\,MB at 81 categories on Animalia, 619.2\,MB at 175 categories on CUB, and 626.7\,MB at 177 categories on StanfordCars.
% This corresponds to an empirical cost of roughly 3.6--3.8\,MB per category.
% Extrapolating this trend suggests that even at $K \approx 1000$, the DP-BOA head would require only about 4\,GB of memory.
% Therefore, although it is heavier than lightweight hashing- or radius-based OCD heads, it remains feasible on modern 8--16\,GB GPUs.

\subsection{Memory Footprint and Runtime of DP-BOA}

To complement the theoretical complexity in \cref{sec:complexity}, we report more detailed wall-clock latency of DP-BOA in \cref{tab:runtime}.
Across all benchmarks, the mean test-time cost is in the tens of milliseconds per sample on a single GPU (NVIDIA TITAN RTX), with worst-case latency still under $90$\,ms.
CUB, which has the largest number of categories, exhibits the highest latency (58.7/81.7\,ms), while Animalia and OxfordPets are noticeably faster, consistent with the $O(d^3 + K d^2)$ dependence of the head.
Overall, DP-BOA trades additional computation and memory for accuracy, but the resulting overhead remains moderate and practically acceptable in the OCD setting.

We also report both head-only and end-to-end GPU memory in \cref{tab:mem_dpboa}.
As expected from the $O(Kd^2)$ storage of full-covariance statistics, the head memory grows approximately linearly with the number of active categories:
299.2\,MB at 81 categories on Animalia, 619.2\,MB at 175 categories on CUB, and 626.7\,MB at 177 categories on StanfordCars.
This corresponds to an empirical cost of roughly 3.6--3.8\,MB per category.
Extrapolating this trend suggests that even at $K \approx 1000$, the DP-BOA head would require only about 4\,GB of memory.
Therefore, although it is heavier than lightweight hashing- or radius-based OCD heads, it remains feasible on modern 8--16\,GB GPUs.

% \begin{table}[t]
% \centering
% \footnotesize
% \setlength{\tabcolsep}{3pt}
% \caption{Accuracy, max latency (Lat.) in milliseconds, and head memory (Mem.) in MB. Numbers in parentheses denote the number of classes.}
% \vspace{-0.5em}
% \label{tab:dpboa_l}
% \resizebox{1.0\textwidth}{!}{%
% \begin{tabular}{l ccc cc ccc cc ccc cc}
% \toprule
% & \multicolumn{5}{c}{Animalia (77)} & \multicolumn{5}{c}{CUB (200)} & \multicolumn{5}{c}{StanfordCars (196)}\\
% \cmidrule(lr){2-6}\cmidrule(lr){7-11}\cmidrule(lr){12-16}
% Method & All & Known & Novel & Lat. & Mem.
%        & All & Known & Novel & Lat. & Mem.
%        & All & Known & Novel & Lat. & Mem. \\
% \midrule
% DP-BOA & 50.7 & 67.6 & 43.7 & 46.1 & 299.2
%          & 53.4 & 57.2 & 51.6 & 81.7 & 619.2 
%          & 32.4 & 58.8 & 19.7 & 82.9 & 626.7 \\
% DP-BOA-L & 49.2 & 64.2 & 42.9 & 34.0 & 40.5
%          & 52.2 & 54.6 & 50.9 & 50.3 & 51.0 
%          & 31.5 & 54.7 & 20.2 & 44.2 & 49.7 \\
% \bottomrule
% \end{tabular}%
% }
% \vspace{-1em}
% \end{table}

\subsection{A Low-Rank Variant for Large-$K$ Settings}

To further improve scalability in higher-dimensional or larger-$K$ regimes, we also evaluate a lightweight variant, DP-BOA-L.
Importantly, DP-BOA-L keeps the same Bayesian birth-or-assign framework as DP-BOA: it still compares prior-weighted posterior-predictive evidence for assigning a sample to an existing category versus spawning a new one.
The only change is how each category covariance is represented and updated.

Instead of storing a full $d \times d$ predictive covariance for every category, DP-BOA-L approximates it by a low-rank anisotropic term plus an isotropic residual,
\begin{equation}
    \mathbf{\Sigma}_k \approx \sigma_k^2 \mathbf I
    + \mathbf U_k \operatorname{diag}(\boldsymbol{\lambda}_k)\mathbf U_k^\top,
\end{equation}
where $\mathbf U_k \in \mathbb R^{d \times r}$ contains the top-$r$ directions, $\boldsymbol{\lambda}_k \in \mathbb R^r$ stores their corresponding spectral strengths, and $\sigma_k^2$ captures the average residual variance outside the low-rank subspace, with $r \ll d$.
This form preserves the dominant anisotropic geometry while reducing the per-category degrees of freedom from $O(d^2)$ to $O(dr)$.

In implementation, DP-BOA-L still maintains the count $n_k$ and mean $\boldsymbol{\mu}_k$ for each category, but it no longer stores a full second-order matrix explicitly.
Instead, each category keeps a fixed-size covariance sketch $\mathbf B_k \in \mathbb R^{\ell \times d}$, with $\ell = O(r)$, and updates it online using Frequent Directions~\cite{ghashami2016frequent}.
Each new centered residual is inserted into the sketch, and a small spectral shrinkage step is applied when needed.
The resulting sketch provides an approximate low-rank decomposition from which $\mathbf U_k$, $\boldsymbol{\lambda}_k$, and $\sigma_k^2$ can be recovered.
With this representation, evaluating a posterior-predictive score only requires projecting $(\mathbf z_t - \boldsymbol{\mu}_k)$ onto the rank-$r$ subspace, together with a few scalar operations for the residual term, so the per-category scoring cost becomes $O(dr)$ rather than $O(d^2)$.
Traversing all $K$ candidate categories therefore costs about $O(Kdr)$, and updating the sketch introduces an additional low-rank maintenance cost that is approximately $O(dr^2)$.
Overall, the head complexity is reduced from $O(Kd^2 + d^3)$ time and $O(Kd^2)$ memory to approximately $O(Kdr + dr^2)$ time and $O(Kdr)$ memory.
In our implementation, we use $r=\ell=32$.

\section{More Hyperparameters Analysis}
\label{app:appendix_hparams}

We briefly analyze two hyperparameters of DP\mbox{-}BOA:
(i) the Empirical--Bayes estimate of the NIW mean-strength $\kappa_0$
from Appendix~\ref{app:appendix_eb_niw}, and
(ii) the degrees-of-freedom cap $n_{\mathrm{cap}}$ used in the heuristic   
for $n_0$ (\cref{eq:n0_rule} in the main text).

\subsection{Effect of the Empirical--Bayes \texorpdfstring{$\kappa_0$}{kappa0}}

\cref{tab:kappa_ablation} compares several fixed
$\kappa_0 \in \{1,10^{-1},\dots,10^{-6}\}$ with our
Empirical--Bayes value $\kappa_0^{\text{EB}}$ on three fine-grained
benchmarks.

\begin{table}[t]
    \centering
    \caption{Ablation of the NIW mean-strength $\kappa_0$ on three
    fine-grained benchmarks.
    ``Ours'' uses the Empirical--Bayes estimate
    $\kappa_0^{\text{EB}}$ from Appendix~\ref{app:appendix_eb_niw}.
    The estimated values are
    $\kappa_0^{\text{EB}}\!\approx\!5.6\times10^{-3}$ (Animalia),
    $2.0\times10^{-2}$ (CUB), and
    $4.4\times10^{-3}$ (Oxford-IIIT Pet).}
    \label{tab:kappa_ablation}
    \small
    % \resizebox{\linewidth}{!}{
    \begin{tabular}{lccccccccc}
    \toprule
    & \multicolumn{3}{c}{Animalia} &
      \multicolumn{3}{c}{CUB} &
      \multicolumn{3}{c}{Oxford-IIIT Pet} \\
    \cmidrule(lr){2-4}\cmidrule(lr){5-7}\cmidrule(lr){8-10}
    $\kappa_0$
    & All & Known & Novel
    & All & Known & Novel
    & All & Known & Novel \\
    \midrule
    $1$       & 45.2 & 59.1 & 39.4 & 52.2 & 51.7 & 52.4 & 57.5 & 60.1 & 56.2 \\
    $10^{-1}$ & 42.6 & 63.8 & 33.9 & 52.8 & 59.1 & 49.6 & \textbf{59.6} & 57.9 & \textbf{60.5} \\
    $10^{-2}$ & 48.0 & 66.6 & 40.3 & 49.3 & 49.0 & 49.5 & 58.8 & 61.5 & 57.3 \\
    $10^{-3}$ & 43.0 & 55.1 & 38.0 & 38.8 & 27.9 & 44.2 & 53.2 & 51.9 & 53.9 \\
    $10^{-4}$ & 37.9 & 45.2 & 34.9 & 38.8 & 26.1 & 45.1 & 47.8 & 33.8 & 55.1 \\
    $10^{-5}$ & 38.1 & 45.3 & 35.2 & 40.6 & 25.0 & 48.4 & 48.1 & 33.9 & 55.5 \\
    $10^{-6}$ & 37.2 & 46.2 & 33.4 & 40.6 & 24.9 & 48.4 & 47.9 & 28.2 & 58.2 \\
    \midrule
    Ours      & \textbf{50.7} & \textbf{67.6} & \textbf{43.7}
              & \textbf{53.4} & \textbf{57.2} & \textbf{51.6}
              & 59.0          & \textbf{63.6} & 56.6 \\
    \bottomrule
    \end{tabular}
    % }
    \vspace{1em}
\end{table}

Across all three datasets, $\kappa_0^{\text{EB}}$ lies in the same order of magnitude as the best grid value and yields the strongest or near-strongest ``All'' accuracy, while clearly outperforming commonly used baselines such as $\kappa_0{=}1$. Very large or very small $\kappa_0$ cause noticeable degradation, confirming that the data-driven estimate provides a good default without extra tuning.

\subsection{Sensitivity of \texorpdfstring{$n_0$}{n0} to \texorpdfstring{$n_{\mathrm{cap}}$}{nmin}}

We choose $n_0$ via
\begin{equation}
\label{eq:n0_rule_app}
    n_0 = \min\!\left(\tfrac{1}{2}\,\bar{n},\,n_{\mathrm{cap}}\right),
    \qquad
    \bar{n}
    =
    \frac{1}{K_S}\sum_{k=1}^{K_S} n_k,
\end{equation}
where $n_{\mathrm{cap}}$ caps the effective degrees of freedom.
This cap is mainly needed on large-scale datasets, where $\bar{n}$ can
be very large; without it, $n_0$ (and hence the Student-$t$ degrees of
freedom) would grow with dataset size, making the predictive almost
Gaussian and overly concentrated.

\cref{tab:nmin_ablation} studies $n_{\mathrm{cap}}$ on CIFAR-100 and
ImageNet-100. Very small $n_{\mathrm{cap}}$ (e.g., $10$) keep the predictive distributions
too heavy-tailed and yield poor overall accuracy, especially on novel
classes. As $n_{\mathrm{cap}}$ increases, performance improves
substantially and is strongest around $50$, with a fairly broad plateau:
on CIFAR-100, the best results lie in the range $50$--$65$, while on
ImageNet-100, $45$--$55$ already gives very similar performance.
This suggests that the method is not overly sensitive to the exact choice
of $n_{\mathrm{cap}}$ as long as it is set in a moderate range near $50$.
Larger values (e.g., $100$ or $\bar{n}/2$) further improve known-class
accuracy but sharply degrade novel-class accuracy, as overly large
degrees of freedom make the Student-$t$ predictive almost Gaussian and
discourage the birth of new categories.
We therefore fix $n_{\mathrm{cap}}{=}50$ in all experiments as a simple,
robust choice near the center of this stable high-performing region.

\begin{table}[t]
    \centering
    \caption{Ablation of the cap $n_{\mathrm{cap}}$ in
    \cref{eq:n0_rule_app} on CIFAR-100 and ImageNet-100.
    ``$\bar{n}/2$'' corresponds to using $n_0=\bar{n}/2$ without the
    cap. Here $\bar{n}/2\!\approx\!250$ (CIFAR-100) and
    $\bar{n}/2\!\approx\!638$ (ImageNet-100). The row marked ${}^*$ is our default.}
    \label{tab:nmin_ablation}
    \small
    % \resizebox{\linewidth}{!}{
    \begin{tabular}{lcccccc}
    \toprule
    & \multicolumn{3}{c}{CIFAR-100} &
      \multicolumn{3}{c}{ImageNet-100} \\
    \cmidrule(lr){2-4}\cmidrule(lr){5-7}
    $n_{\mathrm{cap}}$ &
    All & Known & Novel &
    All & Known & Novel \\
    \midrule
    10   & 48.8 & 65.0 & 16.2 & 29.6 & 57.3 & \textbf{15.7} \\
    25   & 58.2 & 74.3 & 26.0 & 28.9 & 61.4 & 12.6 \\
    35 & 59.8 & 74.0 & 31.6 & 31.6 & 70.1 & 12.4\\
    45 & 60.6 & 74.9 & 31.9 & 33.5 & 74.9 & \underline{12.8}\\
    $50^*$   & \underline{60.7} & \underline{75.0} & 32.0
         & \textbf{33.8} & 75.8 & 12.7 \\
    55 & \underline{60.7} & 74.8 & \underline{32.6} & \underline{33.7} & 77.0 & 12.1\\
    65 & \textbf{61.0} & 74.6 & \textbf{33.7} & 33.0 & 77.1 & 10.9\\
    100  & 59.5 & \textbf{75.8} & 26.8
         & 33.1 & \underline{83.7} &  7.6 \\
    $\bar{n}/2$ & 56.1 & 72.7 & 22.8
                & 33.1 & \textbf{92.8} &  3.2 \\
    \bottomrule
    \end{tabular}
    % }
\end{table}

\section{Feature Geometry Analysis}
\label{app:appendix_geometry}

\subsection{Category Geometry Is Heteroscedastic and Anisotropic}
\label{app:geometry_online}

We first examine the geometry of categories appearing in the unlabeled query stream, since these are the distributions that the online discovery head must model at test time. On CUB, we extract query features using the trained encoder and group unlabeled samples by their ground-truth categories \emph{only for post-hoc analysis}. For each category \(c\), we compute its empirical covariance \(\Sigma_c\) from the corresponding query features.

\begin{figure}[t]
  \centering
  \begin{subfigure}{0.48\linewidth}
    \includegraphics[width=\linewidth]{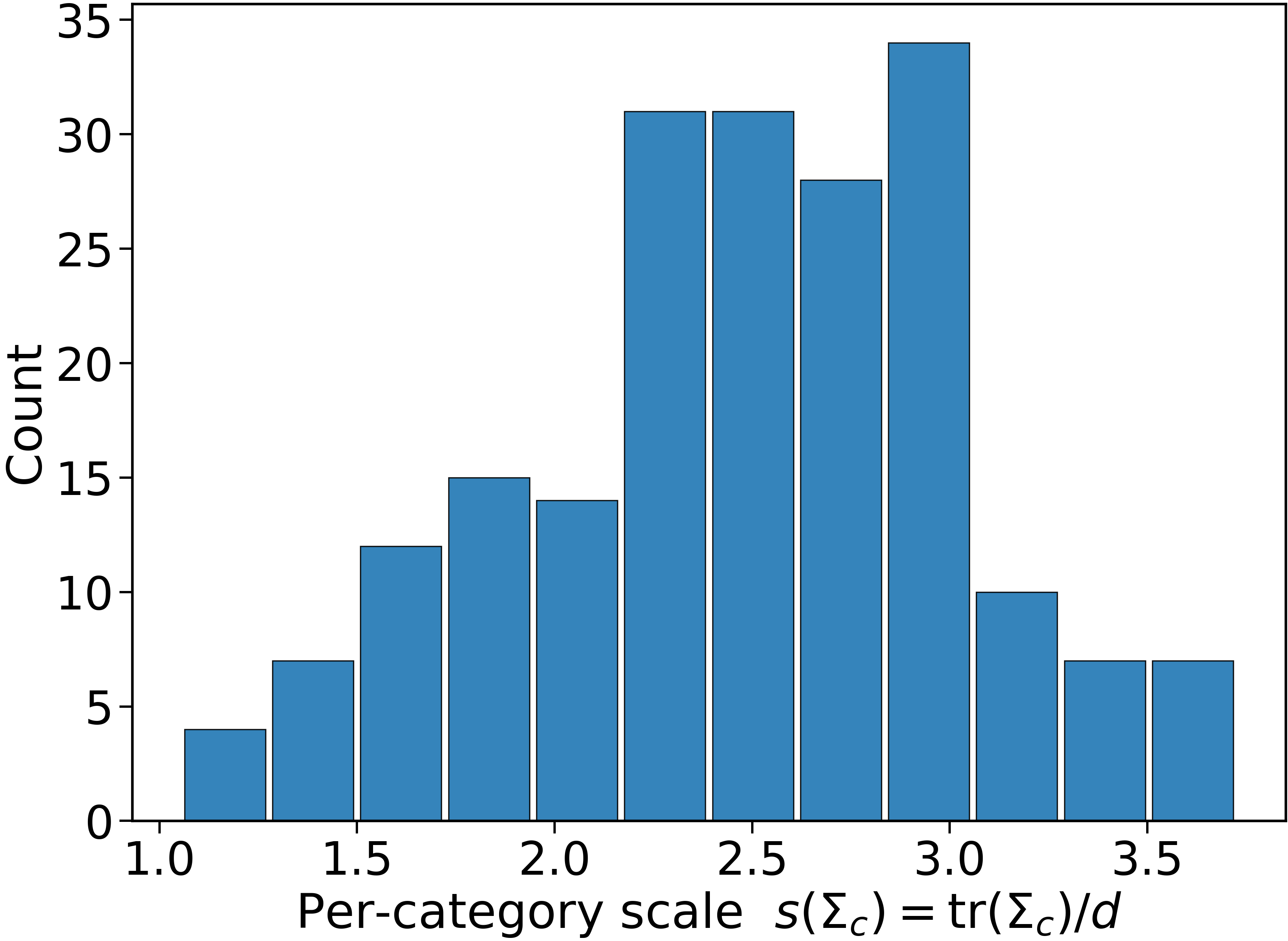}
    \caption{Per-category scale.}
    \label{fig:geom-scale}
  \end{subfigure}\hfill
  \begin{subfigure}{0.48\linewidth}
    \includegraphics[width=\linewidth]{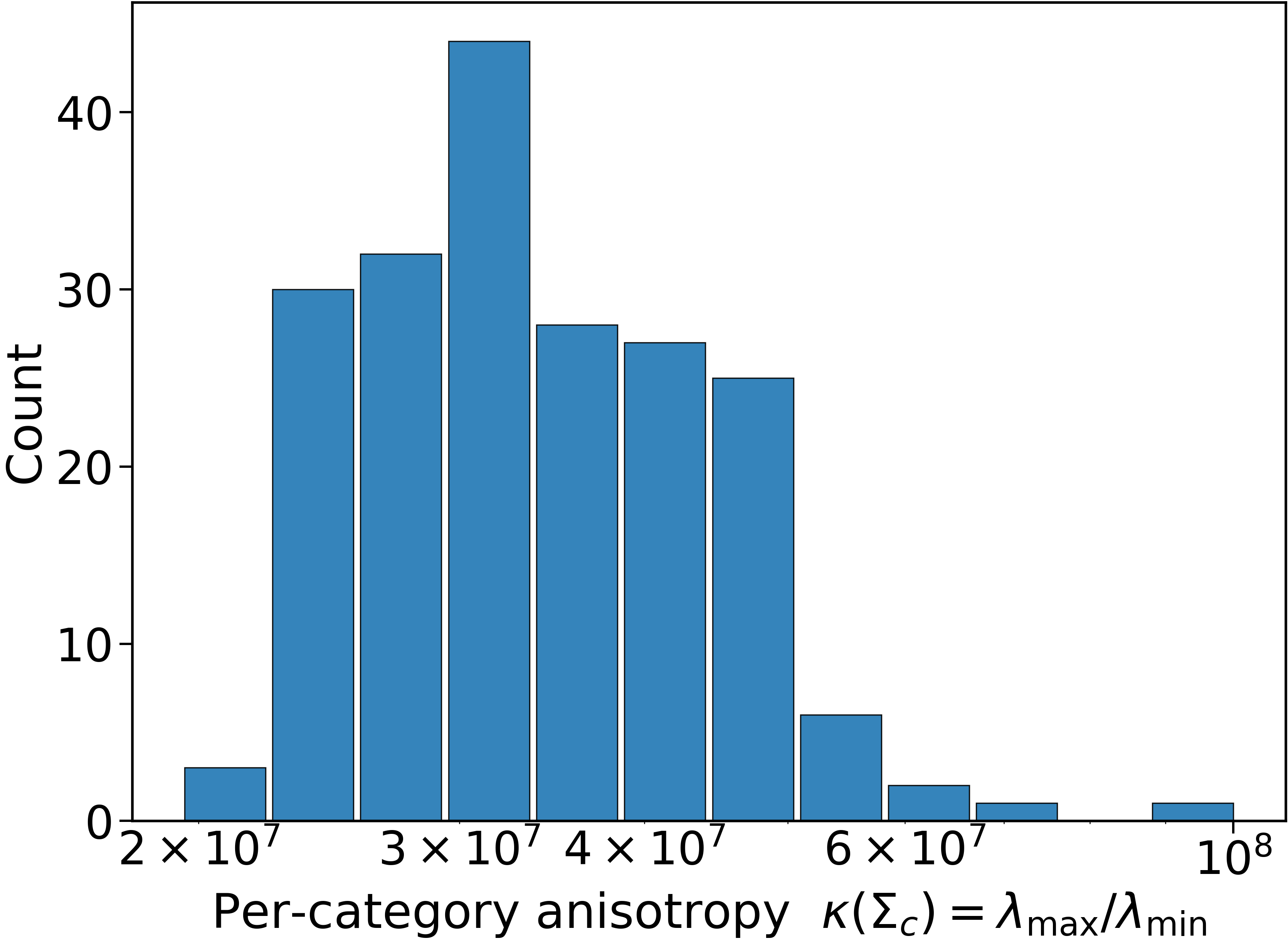}
    \caption{Per-category anisotropy.}
    \label{fig:geom-kappa}
  \end{subfigure}
  \vspace{-4pt}
  \caption{\textbf{Online category geometry on CUB.}
  We analyze the feature distributions of categories in the unlabeled query split.
  In (a), we report the per-category scale
  \(s(\Sigma_c)=\mathrm{tr}(\Sigma_c)/d\), where \(d\) is the feature dimension.
  In (b), we report the per-category anisotropy
  \(\kappa(\Sigma_c)=\lambda_{\max}/\lambda_{\min}\).
  The distributions show substantial variation in scale and clear deviations from
  sphericity.}
  \label{fig:geom-online-cub}
  \vspace{-1.0em}
\end{figure}

As shown in \cref{fig:geom-online-cub}(a), the per-category scale \(s(\Sigma_c)=\mathrm{tr}(\Sigma_c)/d\) varies substantially across query categories, indicating heteroscedasticity rather than a shared variance across categories. \cref{fig:geom-online-cub}(b) further shows a pronounced right tail in the anisotropy ratio \(\kappa(\Sigma_c)=\lambda_{\max}/\lambda_{\min}\), suggesting that many categories occupy elongated, direction-dependent regions rather than spherical neighborhoods. These observations motivate geometry-aware second-order modeling in the online head. In particular, they explain why fixed-shape regions such as Euclidean balls, Hamming balls, or angular caps may be insufficient to describe the category geometry encountered during online discovery.

\subsection{Gaussian Approximation to Category Geometry}
\label{app:geometry_gaussian}

We next examine whether category-conditioned features are reasonably compatible with a \emph{single full-covariance Gaussian approximation}. This analysis should not be interpreted as claiming that every category is exactly Gaussian. Rather, our goal is to assess whether a category-specific mean and covariance can capture the dominant location, scale, and anisotropy of the observed feature clouds.

\begin{figure}[t]
    \centering
    \includegraphics[width=\linewidth]{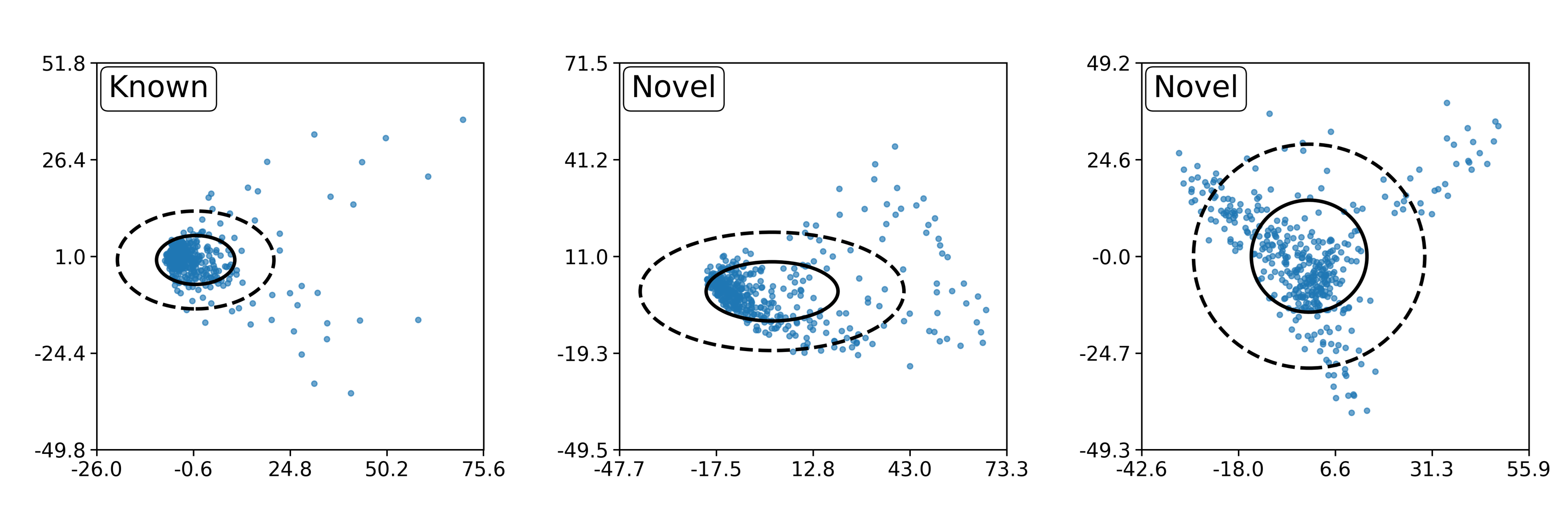}
    \vspace{-2em}
    \caption{\textbf{Category visualizations on CIFAR100.}
    Each panel shows one category in its own 2D PCA plane.
    Blue points are the category features, and the solid / dashed curves denote
    the fitted Gaussian \(1\sigma\) / \(2\sigma\) contours, respectively.}
    \label{fig:gaussian_vis_cifar100}
    \vspace{-1.0em}
\end{figure}

\begin{figure}[t]
    \centering
    \includegraphics[width=\linewidth]{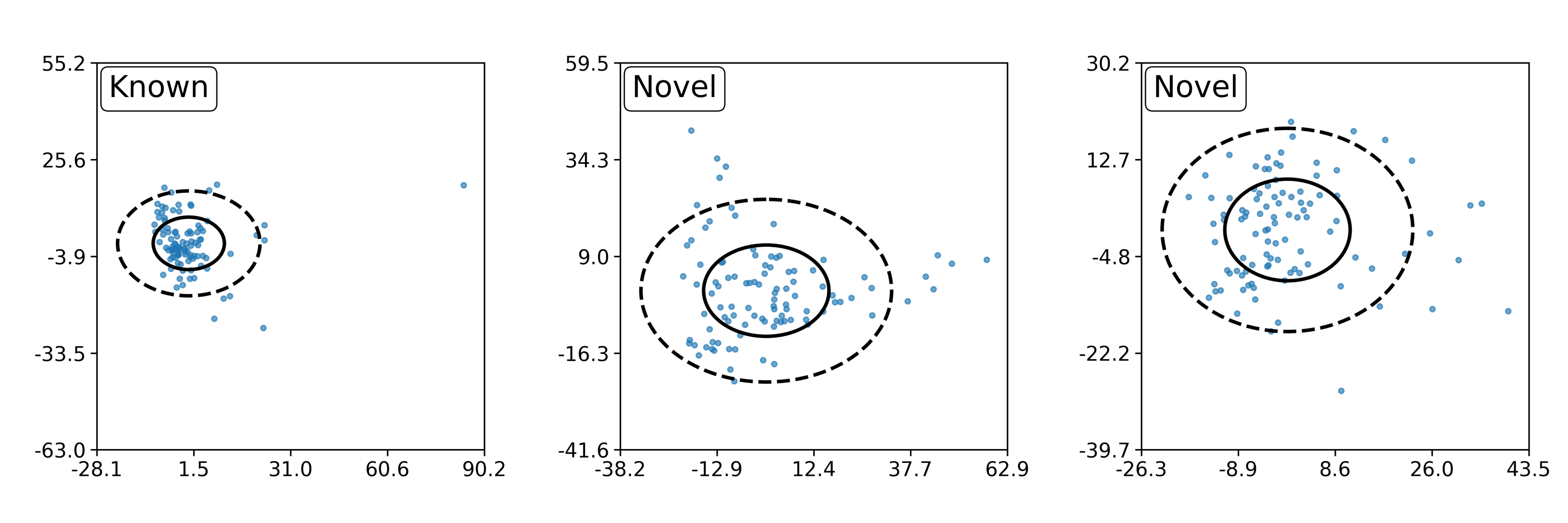}
    \vspace{-2em}
    \caption{\textbf{Category visualizations on Oxford-IIIT Pet.}
    Each panel shows one category in its own 2D PCA plane.
    Blue points are the category features, and the solid / dashed curves denote
    the fitted Gaussian \(1\sigma\) / \(2\sigma\) contours, respectively.}
    \label{fig:gaussian_vis_pets}
    \vspace{-1.0em}
\end{figure}

For visualization, we use all samples of each category and project each category to its own 2D PCA plane, following the protocol in Appendix~\ref{app:appendix_qualitative}. A Gaussian is then fitted to the projected features, and its \(1\sigma\) and \(2\sigma\) contours are overlaid on the scatter plot. %%%%%%

As shown in \cref{fig:gaussian_vis_cifar100,fig:gaussian_vis_pets}, many categories exhibit a dominant central region whose spread is direction-dependent but broadly elliptical. In several examples, the fitted \(1\sigma\) contour aligns with the densest part of the feature cloud, while the \(2\sigma\) contour covers much of the broader support. The fit is not perfect, and some categories show non-elliptical tails or local substructure. Nevertheless, the visualizations suggest that a unimodal full-covariance Gaussian often provides a useful first-order approximation to the dominant category geometry.

Together with the heteroscedasticity and anisotropy observed above, this supports the modeling choice in DP\mbox{-}BOA as a practical approximation: the model does not require categories to be spherical, but instead represents each category by an adaptive full-covariance posterior predictive distribution. At the same time, the single-Gaussian assumption remains an approximation, and we analyze its behavior under a more challenging mismatch setting below.

\subsection{Beyond Single-Gaussian Modeling}
\label{app:geometry_beyond_single_gaussian}

A single full-covariance Gaussian is a tractable and effective default for category-conditioned features in our standard OCD benchmarks, but it is not exact in all regimes. In particular, under domain shift, samples from the same semantic category may occupy different sub-regions of the frozen feature space. This can induce a non-elliptical structure that is less well matched by a single Gaussian component. We therefore use source-target domain shift as a stress test for the single-Gaussian approximation and for the support-set prior calibration.

To probe this setting, we conduct a stress test on DomainNet~\cite{peng2019moment}, following the known--novel split protocol introduced in HiLo~\cite{wang2024hilo}. The training data contains a single source domain, while the online query stream mixes source-domain and target-domain samples. This creates a mismatch between the support-set statistics used to initialize DP\mbox{-}BOA and the geometry of future query samples, especially in the target domain.

The results in \cref{tab:domainnet_shift} show that DP\mbox{-}BOA remains competitive under this mismatch. It achieves the best \emph{All} accuracy in three of the four domain slices and the best \emph{Novel} accuracy in all four. This suggests that the proposed posterior-predictive birth-or-assign rule remains effective even when category geometry is distorted by cross-domain mismatch. Importantly, DP\mbox{-}BOA does not require the support-set prior to perfectly represent all future novel classes. The support statistics provide an initial calibration, while the category posterior states are updated online after each decision and increasingly reflect accumulated stream evidence.

At the same time, the source-target gap remains clear, and PHE is still stronger on some target-domain \emph{Known} metrics. This indicates that the current frozen-feature formulation is reasonably robust compared with existing OCD methods, but severe target-domain shift remains challenging, especially for preserving known-category assignments in the target domain.

Two natural extensions may further improve performance in this regime. The first is to enrich the category model itself. Because DP\mbox{-}BOA makes decisions through a modular comparison of prior-weighted predictive evidence, the current single-Gaussian predictive density can, in principle, be replaced by a richer model, such as a multi-component or hierarchical nonparametric mixture, while preserving the same birth-or-assign framework~\cite{teh2006hierarchical}. Such models may better capture domain-induced sub-modes or non-elliptical structure within a category. However, they would also enlarge the latent state per category, introduce additional components and hyperparameters, and increase time and memory complexity.

The second direction is to adapt the feature space at test time so that samples from different domains become better aligned. This could reduce the source-target gap observed above. However, integrating test-time adaptation into OCD is non-trivial. The stream contains both known and novel categories, so objectives that simply encourage confident assignment may bias the model toward absorbing truly novel samples into existing categories, thereby suppressing category birth. Online backbone updates also introduce extra backpropagation cost, optimizer state, and the risk of drift or forgetting under non-stationary streams.

\begin{table}[t]
\centering
\scriptsize
\setlength{\tabcolsep}{2pt}
\caption{\textbf{Model performance on DomainNet under source-target domain shift.}
Best results in each column are in \textbf{bold}.}
\vspace{-0.5em}
\label{tab:domainnet_shift}
\resizebox{\textwidth}{!}{%
\begin{tabular}{l ccc ccc | ccc ccc}
\toprule
& \multicolumn{3}{c}{Real (Source)}
& \multicolumn{3}{c|}{Clipart (Target)}
& \multicolumn{3}{c}{Real (Source)}
& \multicolumn{3}{c}{Painting (Target)} \\
\cmidrule(lr){2-4}\cmidrule(lr){5-7}\cmidrule(lr){8-10}\cmidrule(lr){11-13}
Method
& All & Known & Novel
& All & Known & Novel
& All & Known & Novel
& All & Known & Novel \\
\midrule
SMILE
& 18.3 & 18.5 & 18.3
&  8.6 & 11.0 &  6.2
& 19.1 & 23.5 & 18.2
& 13.1 & 16.7 &  9.5 \\
PHE
& 27.6 & 37.9 & 25.7
&  \textbf{19.1} & \textbf{31.6} &  6.3
& 26.1 & \textbf{41.0} & 23.3
& 22.4 & 36.3 &  8.2 \\
DP\mbox{-}BOA
& \textbf{37.9} & \textbf{51.0} & \textbf{35.4}
& 18.9 & 23.7 & \textbf{14.1}
& \textbf{35.6} & 31.4 & \textbf{36.4}
& \textbf{24.9} & \textbf{38.5} & \textbf{11.0} \\
\bottomrule
\end{tabular}%
}
\vspace{-1em}
\end{table}

Overall, these results suggest that the current single-Gaussian DP\mbox{-}BOA is a strong and robust default under standard OCD benchmarks and remains competitive under domain shift. Future gains may come from richer category models or carefully controlled online feature adaptation, but both directions introduce additional computational and stability challenges.

\section{Comparison with Simpler Probabilistic Heads}
\label{app:appendix_robust}
% \label{app:fixed_threshold}

We further compare DP\mbox{-}BOA with simpler online probabilistic heads under identical frozen features. These variants use the same encoder and support/query split as DP\mbox{-}BOA, but replace the explicit DP-weighted birth hypothesis with a fixed rejection mechanism. This comparison helps isolate the contribution of the proposed birth-or-assign rule from representation quality or support-set calibration alone.

\paragraph{Posterior-threshold variant.}
The first variant uses the same posterior probability as DP\mbox{-}BOA for existing categories, but replaces the explicit birth hypothesis with a threshold rule.
Specifically, for an incoming sample $z_t$, the variant predicts
\begin{equation}
    k^* = \arg\max_k P(c_t = k \mid z_t, \mathcal D_{t-1}),
\end{equation}
and spawns a new category if
\begin{equation}
    \max_k P(c_t = k \mid z_t, \mathcal D_{t-1}) < p_0;
\end{equation}
otherwise, it assigns the sample to $k^*$. For numerical stability, the threshold $p_0$ is tuned in log-space on a small held-out validation split carved out from the training data. All other settings, including the encoder and online posterior updates, are kept unchanged.

\begin{table}[t]
    \centering
    \caption{\textbf{Comparison with simpler probabilistic heads.}
    Maha uses a support-calibrated Mahalanobis metric with a fixed rejection threshold.
    Post.-thr. uses the same posterior probability as DP\mbox{-}BOA for existing categories but replaces the explicit birth hypothesis with a fixed posterior threshold.
    DP\mbox{-}BOA keeps the full DP-weighted birth-or-assign comparison.
    All methods use identical frozen features. Best results in each column are in \textbf{bold}.}
    \vspace{-0.5em}
    \label{tab:fixed_threshold}
    \small
    \setlength{\tabcolsep}{4.5pt}
    \renewcommand{\arraystretch}{1.05}
    \begin{tabular}{lccccccccc}
        \toprule
        & \multicolumn{3}{c}{Pets}
        & \multicolumn{3}{c}{CUB}
        & \multicolumn{3}{c}{Animalia} \\
        \cmidrule(lr){2-4}
        \cmidrule(lr){5-7}
        \cmidrule(lr){8-10}
        Method
        & All & Known & Novel
        & All & Known & Novel
        & All & Known & Novel \\
        \midrule
        Maha
        & 45.2 & 44.3 & 45.6
        & 36.8 & 45.2 & 32.6
        & 44.4 & \textbf{73.0} & 32.5 \\
        Post.-thr.
        & 57.8 & \textbf{65.0} & 54.0
        & 50.8 & 54.9 & 48.8
        & 48.6 & 65.6 & 41.5 \\
        DP\mbox{-}BOA
        & \textbf{59.0} & 63.6 & \textbf{56.6}
        & \textbf{53.4} & \textbf{57.2} & \textbf{51.6}
        & \textbf{50.7} & 67.6 & \textbf{43.7} \\
        \bottomrule
    \end{tabular}
    \vspace{-1em}
\end{table}

\paragraph{Mahalanobis-threshold variant.}
The second variant uses a support-calibrated Mahalanobis metric with a fixed rejection threshold. Let $\Sigma_{\mathrm{within}}$ denote the pooled within-class covariance estimated from the labeled support set. For each existing category $k$, we compute
\begin{equation}
    d_k^{\mathrm{Maha}}(z_t)
    =
    (z_t-\bar z_k)^\top
    \Sigma_{\mathrm{within}}^{-1}
    (z_t-\bar z_k),
\end{equation}
and select the nearest category
\begin{equation}
    k^* = \arg\min_k d_k^{\mathrm{Maha}}(z_t).
\end{equation}
The sample is assigned to $k^*$ if $\min_k d_k^{\mathrm{Maha}}(z_t) \leq \tau_{\mathrm{Maha}}$, and otherwise spawns a new category. The threshold $\tau_{\mathrm{Maha}}$ is tuned on the same held-out validation split. This variant uses the same support-set statistics for calibration, but does not maintain category-specific posterior uncertainty or compare assignment with an explicit DP-induced birth score.

The results are reported in \cref{tab:fixed_threshold}. 
The Mahalanobis-threshold variant can be competitive on known-class accuracy, but its novel-class performance is substantially weaker, suggesting that a fixed global rejection radius is insufficient for reliable online discovery.
The posterior-threshold variant improves over the Mahalanobis head by using NIW posterior predictives, but still underperforms DP\mbox{-}BOA.
Overall, DP\mbox{-}BOA achieves the best \emph{All} and \emph{Novel} accuracy on all three datasets, showing that its advantage does not come merely from posterior scoring or support-set covariance calibration. Instead, explicitly comparing assignment evidence with DP-weighted birth evidence provides a more calibrated online decision rule.

\section{Temporal Diagnostics}
\label{app:temporal_diagnostics}

We further analyze the temporal behavior of DP\mbox{-}BOA along the query stream. The goal is to examine whether the online posterior state becomes more stable as evidence accumulates, and whether the proposed birth-or-assign rule reduces unnecessary category fragmentation. We partition the query stream into five equal temporal bins, corresponding to $0.2T$, $0.4T$, $0.6T$, $0.8T$, and $T$, where $T$ denotes the stream length.

\begin{figure*}[t]
    \centering
    \begin{subfigure}[t]{0.32\textwidth}
        \centering
        \includegraphics[width=\linewidth]{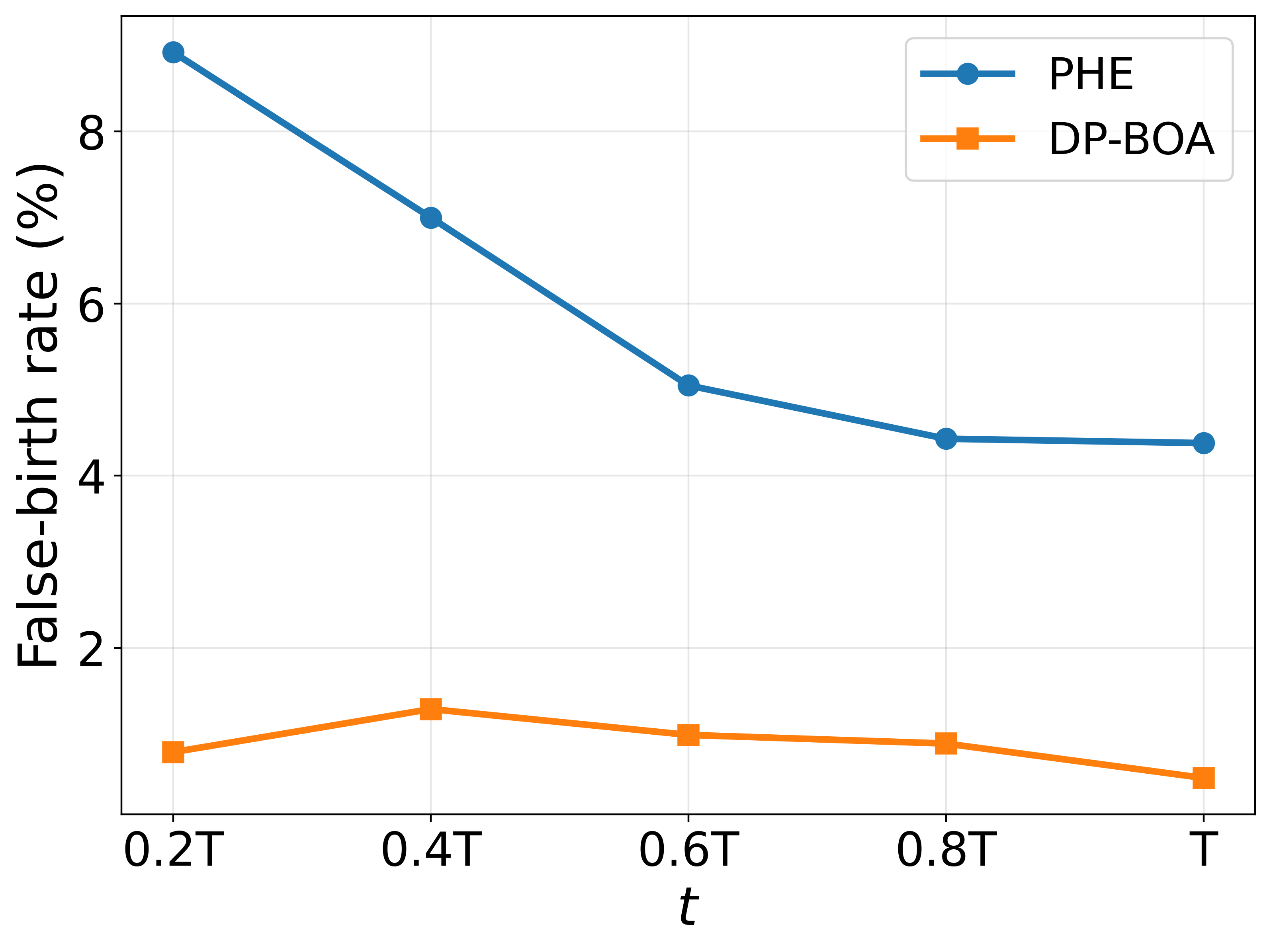}
        \caption{False-birth rate.}
        \label{fig:temporal_false_birth_rate}
    \end{subfigure}
    \hfill
    \begin{subfigure}[t]{0.32\textwidth}
        \centering
        \includegraphics[width=\linewidth]{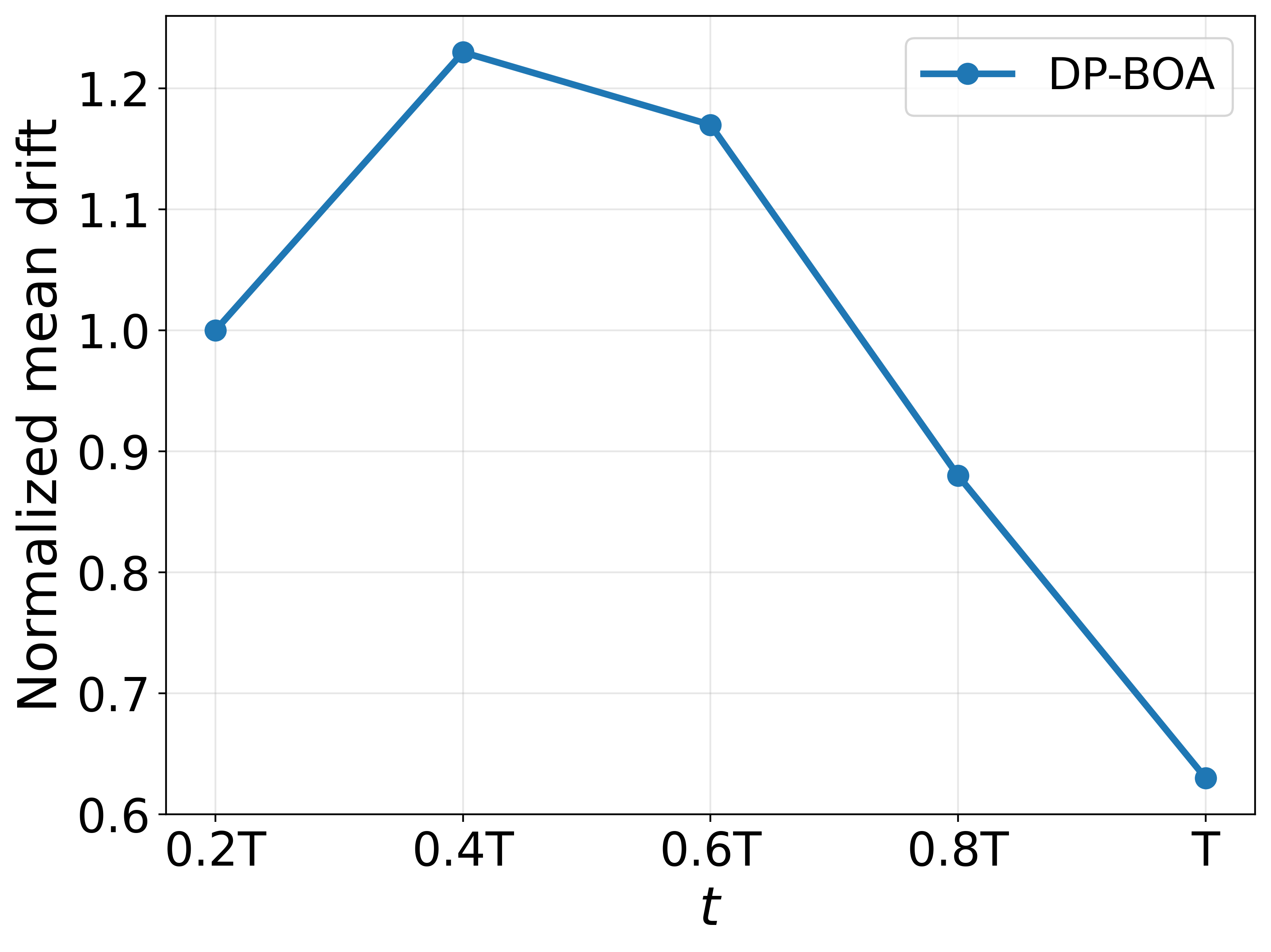}
        \caption{Cluster-mean drift.}
        \label{fig:temporal_mean_drift}
    \end{subfigure}
    \hfill
    \begin{subfigure}[t]{0.32\textwidth}
        \centering
        \includegraphics[width=\linewidth]{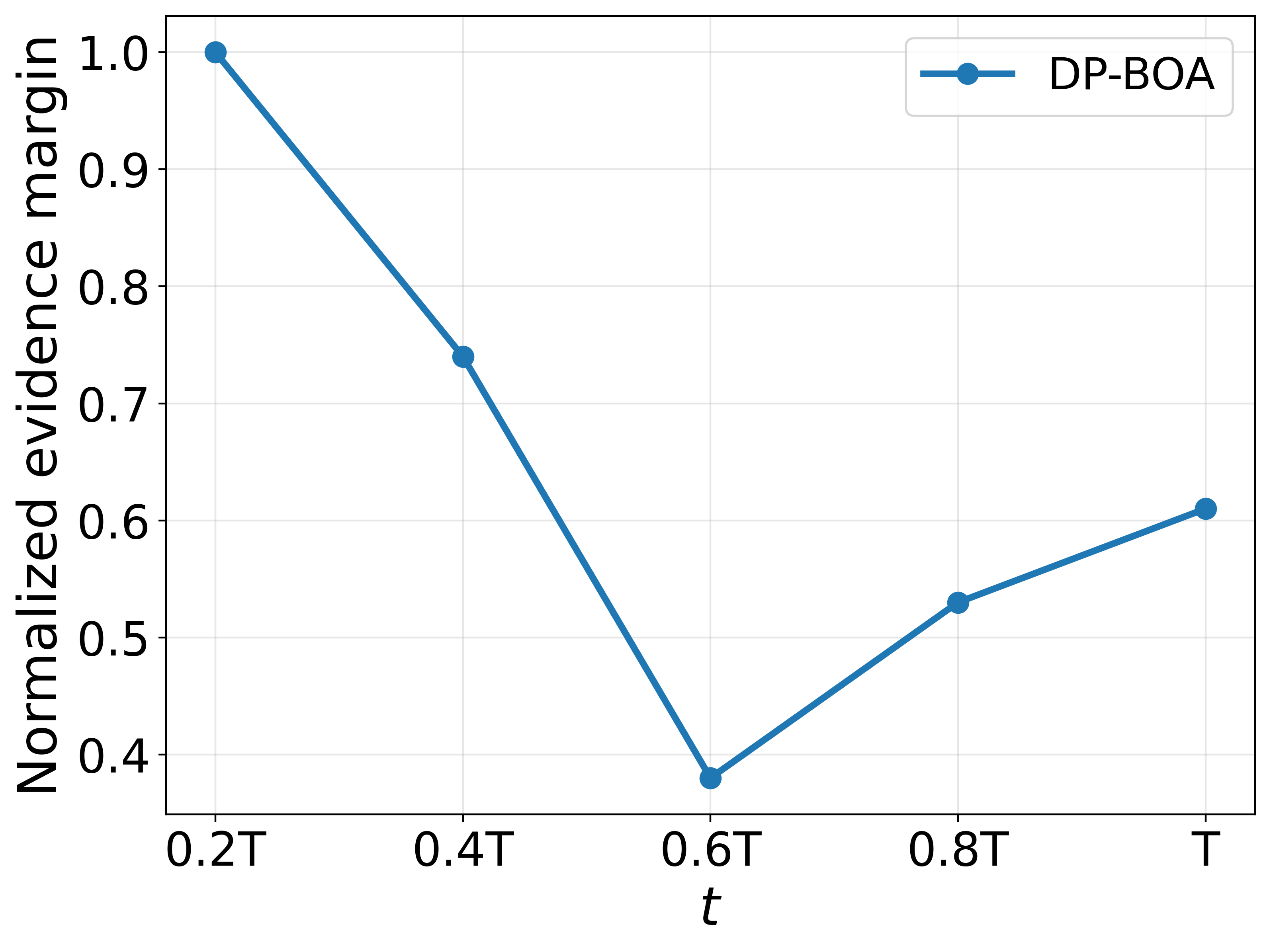}
        \caption{Evidence margin.}
        \label{fig:temporal_evidence_margin}
    \end{subfigure}
    \caption{\textbf{Temporal diagnostics on Animalia.}
    From left to right, we show the false-birth rate, the normalized posterior cluster-mean drift, and the normalized evidence margin over five temporal bins along the query stream.}
    \label{fig:temporal_diagnostics}
    \vspace{-1em}
\end{figure*}

\paragraph{False-birth rate.}
A birth decision is counted as false if the model spawns a new category for a sample whose ground-truth class has already been represented by an existing category in the current stream state. This includes known-class samples, whose categories are initialized from the labeled support set, and novel-class samples whose ground-truth class has appeared earlier in the query stream. For a temporal bin $\mathcal{B}$, we compute
\begin{equation}
    \mathrm{FBR}(\mathcal{B})
    =
    \frac{1}{|\mathcal{B}|}
    \sum_{t \in \mathcal{B}}
    \mathbf{1}\!\left[
        \hat c_t = \mathrm{new}
        \;\wedge\;
        y_t \text{ has been represented before time } t
    \right]
    \times 100\%.
    \label{eq:false_birth_rate}
\end{equation}
A lower false-birth rate indicates that the method is less likely to over-fragment existing categories as the stream progresses. As shown in \cref{fig:temporal_false_birth_rate}, DP\mbox{-}BOA maintains a much lower false-birth rate than PHE across all temporal bins on Animalia. Specifically, the false-birth rate of DP\mbox{-}BOA remains below $1.3\%$ throughout the stream and decreases to $0.5\%$ at the end, whereas PHE stays substantially higher, decreasing from $8.9\%$ to $4.4\%$. This suggests that the explicit birth hypothesis in DP\mbox{-}BOA does not lead to excessive category creation and is consistent with reduced category fragmentation.

\paragraph{Cluster-mean drift.}
To quantify how much the posterior state changes after online updates, we measure the drift of the selected category mean. Let $\mu_{\hat c_t}^{(t-1)}$ and $\mu_{\hat c_t}^{(t)}$ denote the posterior mean of the selected category before and after processing $z_t$, respectively. We define
\begin{equation}
    \Delta \mu_t
    =
    \left\|
        \mu_{\hat c_t}^{(t)}
        -
        \mu_{\hat c_t}^{(t-1)}
    \right\|_2 .
    \label{eq:mean_drift}
\end{equation}
We report the average $\Delta \mu_t$ in each temporal bin, normalized by the value in the first bin for readability. A decreasing trend indicates that category posteriors become more stable after early stream adaptation. As shown in \cref{fig:temporal_mean_drift}, the normalized posterior cluster-mean drift of DP\mbox{-}BOA first increases slightly from $1.0$ to $1.2$, remains close to this level at $0.6T$ with a value of $1.2$, and then decreases to $0.9$ and $0.6$ in the later bins. This pattern suggests that the posterior state undergoes an initial adaptation phase and then becomes progressively more stable as additional evidence is accumulated.

\paragraph{Evidence margin.}
To examine decision ambiguity along the stream, we measure the evidence margin for each incoming sample before the online update. Specifically, we compute the log evidence of all existing-category assignment hypotheses together with the explicit birth hypothesis, and define the margin as the gap between the largest and second-largest log evidence:
\begin{equation}
    m_t = \log p_t^{(1)} - \log p_t^{(2)} .
    \label{eq:evidence_margin}
\end{equation}
We report the average $m_t$ in each temporal bin, normalized by the value in the first bin for readability. A larger margin means that the selected birth-or-assign hypothesis is more clearly separated from its closest competitor, while a smaller margin indicates stronger ambiguity among competing hypotheses. As shown in \cref{fig:temporal_evidence_margin}, the normalized evidence margin of DP\mbox{-}BOA decreases from $1.0$ to $0.7$ and then to $0.4$ in the middle of the stream, before partially recovering to $0.5$ and $0.6$ in the later bins. This non-monotonic trend is reasonable in the online discovery setting, because two competing effects act simultaneously: as more samples are observed, posterior statistics become better adapted and can sharpen decisions; at the same time, newly created categories enlarge the hypothesis set and intensify competition among nearby categories.

% \clearpage\mbox{}Page \thepage\ of the manuscript.
% \clearpage\mbox{}Page \thepage\ of the manuscript.
% \clearpage\mbox{}Page \thepage\ of the manuscript.
% \clearpage\mbox{}Page \thepage\ of the manuscript.
% \clearpage\mbox{}Page \thepage\ of the manuscript. This is the last page.
% Now we have reached the maximum length of an ECCV \ECCVyear{} submission (excluding references and acknowledgements).
% References should start immediately after the main text, but can continue past p.\ 14 if needed. 
% \clearpage  % TODO FINAL: This \clearpage needs to be removed from both review and camera-ready versions.

% Please insert your acknowledgments here.

% \newpage

\end{document}